\RequirePackage{fix-cm}

\documentclass[smallcondensed]{svjour3}     % onecolumn (ditto)

\smartqed  % flush right qed marks, e.g. at end of proof

\usepackage[utf8]{inputenc} % allow utf-8 input
\usepackage[T1]{fontenc}    % use 8-bit T1 fonts
\usepackage{hyperref}       % hyperlinks
\usepackage{url}            % simple URL typesetting
\usepackage{booktabs}       % professional-quality tables
\usepackage{nicefrac}       % compact symbols for 1/2, etc.
\usepackage{microtype}      % microtypography
\usepackage{lipsum}
\usepackage{natbib}%[numbers]{natbib}

\usepackage{algorithm}
\usepackage{algorithmic}

\usepackage{amsthm,amsmath,amsfonts,amssymb,bm,bbm}

\def\*#1{\bm{#1}}
\allowdisplaybreaks

\newtheorem{theo}{Theorem}
\newtheorem{lem}{Lemma}

\newtheorem{asmn}{Assumption}
\newtheorem{rem}{Remark}

\usepackage{float}
\usepackage{adjustbox}
\usepackage{multicol}
\usepackage{multirow}
\usepackage{graphicx}
\usepackage{subfigure}
\usepackage{color}
\usepackage{epstopdf}
\usepackage{caption}
\usepackage[table,xcdraw]{xcolor}
\colorlet{changecolor}{black}
\colorlet{mr}{black}

\journalname{Machine Learning}

\begin{document}

\title{The Backbone Method for Ultra-High Dimensional Sparse Machine Learning}
\titlerunning{The Backbone Method}        % if too long for running head

\author{Dimitris Bertsimas
        \and
        Vassilis Digalakis Jr. 
}

\institute{Dimitris Bertsimas \at
              Operations Research Center and Sloan School of Management, Massachusetts Institute of Technology, Cambridge, MA 02139, USA\\
              dbertsim@mit.edu 
           \and
           Vassilis Digalakis Jr. \at
              Operations Research Center, Massachusetts Institute of Technology, Cambridge, MA 02139, USA\\
              vvdig@mit.edu 
}

\date{First submission: 06/2020. Revised and resubmitted to Machine Learning: 10/2021.}

\maketitle

%%%%%%%%%%%%%%%%%%%%%%%%%%%%%%%%%%%%%%%%%%%%%%%%%%%%%%%%%%%%%%%%%%%%%%%%%%%%%%%%%%%%%%%%%%%%%%%%%%%%%%%%%%%%%%
%%%%%%%%%%%%%%%%%%%%%%%%%%%%%%%%%%%%%%%%%%%%%%%%%%%%%%%%%%%%%%%%%%%%%%%%%%%%%%%%%%%%%%%%%%%%%%%%%%%%%%%%%%%%%%
%%%%%%%%%%%%%%%%%%%%%%%%%%%%%%%%%%%%%%%%%%%%%%%%%%%%%%%%%%%%%%%%%%%%%%%%%%%%%%%%%%%%%%%%%%%%%%%%%%%%%%%%%%%%%%

\begin{abstract}
We present the backbone method, a generic framework that enables sparse and interpretable supervised machine learning methods to scale to ultra-high dimensional problems. 
{\color{changecolor} We solve sparse regression problems with $10^7$ features in minutes and $10^8$ features in hours, as well as decision tree problems with $10^5$ features in minutes. }
The proposed method operates in two phases{\color{mr}:} 
we first determine the backbone set, {\color{mr}consisting} of potentially relevant features, by solving a number of tractable subproblems; 
then, we solve a reduced problem, considering only the backbone features. 
{\color{changecolor} For the sparse regression problem, {\color{mr}our theoretical analysis shows} that, under certain assumptions and with high probability, the backbone set consists of the {\color{mr}truly relevant} features.}
Numerical experiments {\color{changecolor}on both synthetic and real-world datasets demonstrate that our method outperforms or competes with state-of-the-art methods in ultra-high dimensional problems, and competes with optimal solutions in problems where exact methods scale,}
both in terms of recovering the {\color{mr}truly relevant} features and in its out-of-sample predictive performance.
\end{abstract}

%%%%%%%%%%%%%%%%%%%%%%%%%%%%%%%%%%%%%%%%%%%%%%%%%%%%%%%%%%%%%%%%%%%%%%%%%%%%%%%%%%%%%%%%%%%%%%%%%%%%%%%%%%%%%%
%%%%%%%%%%%%%%%%%%%%%%%%%%%%%%%%%%%%%%%%%%%%%%%%%%%%%%%%%%%%%%%%%%%%%%%%%%%%%%%%%%%%%%%%%%%%%%%%%%%%%%%%%%%%%%
%%%%%%%%%%%%%%%%%%%%%%%%%%%%%%%%%%%%%%%%%%%%%%%%%%%%%%%%%%%%%%%%%%%%%%%%%%%%%%%%%%%%%%%%%%%%%%%%%%%%%%%%%%%%%%

\section{Introduction} \label{sec:introduction}
In the {\color{mr}big data} era, the scalability of machine learning models is constantly challenged. \emph{Ultra-high dimensional} datasets are present in a variety of applications, ranging from physical sciences and engineering to social sciences and medicine. From a practical standpoint, we consider a dataset to be ultra-high dimensional when the number of features $p$ exceeds the number of data points $n$ and is at least two orders of magnitude greater than the known scalability limit of the learning task at hand; {\color{mr}as we discuss in the sequel, this limit is $p\sim10^5$ for exact sparse regression and $p\sim10^3$ for decision tree learning using state-of-the-art mixed-integer optimization (MIO) algorithms.} %this limit is $p\sim10^5$ for the sparse regression algorithm that we use, and $p\sim10^3$ for the decision tree induction algorithm. 
{\color{changecolor} From a theoretical standpoint, we assume that $\log p = O(n^\xi)$ for some $0<\xi<1.$}

A commonly held assumption to address high dimensional regimes is that the underlying model is \emph{sparse}, that is, among all input features, only $k<n\ll p$ (e.g., $k\leq100$) are relevant for the task at hand \citep{hastie2015statistical, bertsimas2019book}. For instance, in the context of regression, the sparsity assumption implies that only a few regressors are set to nonzero level \citep{beale1967discarding,hocking1967selection}. In the context of decision trees, it means that most features do not appear in any split node \citep{breiman1984classification}. The sparsity requirement can either be explicitly imposed via a cardinality constraint or penalty, as in sparse regression (also known as best subset selection), or can implicitly be imposed by the structure of the model, as in depth-constrained decision trees, where the number of features that we split on is upper bounded as function of the tree depth. In general, sparsity is a desirable property for machine learning models, as it makes them more interpretable \citep{rudin2019stop} and often improves their generalization ability \citep{ng1998feature}.

Learning a {\color{mr}\emph{sparse and interpretable machine learning model} can naturally be formulated as an MIO problem \citep{bertsimas2016best, carrizosa2016strongly, ustun2016supersparse}, e.g.,} by associating each feature with an auxiliary binary decision variable that is set to 1 if and only if the feature is identified as relevant \citep{bertsimas2016best}. Despite the -theoretical- computational hardness of solving MIO problems, the remarkable progress in the field has motivated their use in a variety of machine learning problems, ranging from sparse regression and decision trees {\color{mr}to principal component analysis, clustering, and matrix completion (see, e.g., the book by \cite{bertsimas2019book} and the survey by \cite{gambella2021optimization})}. The rich modeling framework of MIO enables us to directly model the problem at hand and often compute provably optimal or near-optimal solutions using either exact methods (e.g., branch and bound) or tailored heuristics (e.g., local search). Moreover, the computational efficiency and scalability of MIO-based methods is far beyond what one would imagine a decade ago. For example, {\color{mr}we are able to solve sparse regression problems with $p\sim10^5$ features \citep{bertsimas2020sparse} and induce decision trees for problems with $p\sim10^3$ features  \citep{bertsimas2017optimal} in minutes.} Nevertheless, MIO-based formulations are still \emph{challenged in ultra-high dimensional regimes} as the ones we examine.

The \emph{standard way of addressing ultra-high dimensional problems} is to perform either a screening (i.e., filtering) step \citep{fan2008sure}, that eliminates a large portion of the features, or more sophisticated dimensionality reduction methods: {\bf(a)} feature selection {\color{mr}methods, which} aim to select a subset of relevant features, or {\bf(b)} feature extraction {\color{mr}methods, which} project the original features to a new, lower-dimensional feature space; see \cite{guyon2003introduction,li2017feature} for detailed reviews as well as \cite{bertolazzi2016integer} for an MIO-based approach to feature selection. After the dimensionality reduction step, we can solve the MIO formulation for the reduced problem, provided that sufficiently many features have been discarded. However, commonly used screening (or, more generally, feature selection) approaches are heuristic and often lead to suboptimal solutions, whereas feature extraction approaches are uninterpretable in terms of the original problem's features.

In this paper, we introduce the \emph{backbone method for sparse supervised learning}, a two-phase framework that, as we empirically show, scales to ultra-high dimensions (as defined earlier) and provides significantly higher quality solutions than screening approaches, without sacrificing interpretability (as opposed to feature extraction methods). In the first phase of our proposed method, we aim to identify the ``backbone'' of the problem, which consists of features that are likely to be part of an optimal solution. We do so by collecting the solutions to a series of subproblems that are more tractable than the original problem and whose solutions are likely to contain relevant features. These subproblems also need to be ``selective,'' in the sense that only the fittest features survive after solving each of them. In the second phase, we solve the MIO formulation considering only the features that have been included in the backbone. Although our framework is generic and can be applied to a large variety of problems, we focus on two particular problems, namely sparse regression and decision trees, and illustrate how the backbone method can be applied to them. {\color{changecolor}We accurately solve sparse regression problems with $10^7$ features in minutes and $10^8$ features in hours, as well as decision tree problems with $10^5$ features in minutes,} that is, two orders of magnitude larger than the known scalability limit of {\color{mr}methods which} solve to optimality or near-optimality the actual MIO formulation for each problem.

Our proposed backbone method was inspired by a large-scale vehicle routing application; using a backbone algorithm, \cite{bertsimas2019online} solve, in seconds, problems with thousands of taxis serving tens of thousands of customers per hour. In this work, we develop the backbone method for sparse supervised learning in full generality. Moreover, we remark that the method can be applied to a variety of machine learning problems that exhibit sparse structure, even beyond supervised learning. For example, in the clustering problem, a big portion of pairs of data points will never be clustered together in any near optimal cluster assignment and hence we need not consider assigning them to the same cluster. Finally, we note that, within the sparse supervised learning framework that we examine in this paper, the backbone method can also be used as a feature selection technique in combination with \emph{any sparsity-imposing but not necessarily MIO-based method}.

\subsection{Advances in Sparse Regression} \label{subsec:advances-sr}
% {\color{red} (Removed about half of this section.)} 
In this section, we briefly review the landscape of the sparse regression problem, which is one of the two problems that we tackle using the backbone method. {\color{changecolor} Formally, a regression problem is considered \emph{high dimensional} when the rank of the design matrix $\*X$ is smaller than the number of features $p$, i.e., $\text{rank}(\*X)<p$; this is, for example, the case when $n$, the number of data points in the dataset, satisfies $n<p$. In such regimes, the regression model is challenged, as the underlying linear system is underdetermined, and further assumptions are required. The application of sparse or sparsity-inducing methods provides a way around this limitation and, in addition, hopefully leads to more interpretable models, where only a few features actually affect the prediction.}

\paragraph{Exact Sparse Regression Formulation.}
The sparse regression problem with an explicit sparsity constraint, also known as the best subset selection problem, is the most natural way of performing simultaneous feature selection and model fitting for regression. The sparsity constraint is imposed by requiring that the $\ell_0$ (pseudo)norm of the vector of regressors $\*w \in \mathbb{R}^p$ is less than a predetermined degree of sparsity $k \ll p$, namely, $ \| \*w \|_0 \leq k,$ where $\| \*w \|_0 = \sum_{j=1}^p \mathbf{1}_{( w_j \not= 0)}$ and $\mathbf{1}_{(\cdot)}$ denotes the indicator function. However, the $\ell_0$ constraint is very far from being convex and the resulting combinatorial optimization problem is NP-hard \citep{natarajan1995sparse}.

Motivated by the advances in MIO and despite the diagnosed hardness of the problem, a recent line of work solves the problem exactly to optimality or near-optimality (i.e., within a user-specified optimality gap that can be set to an arbitrarily small value up to machine precision). \cite{bertsimas2016best} cast best subset selection as a mixed integer quadratic program and solve it with a commercial solver for problems with $p\sim10^3$. \cite{bertsimas2020sparse} reformulate the sparse regression problem as a pure binary optimization problem and use a cutting planes-type algorithm that, based on the outer approximation method of \cite{duran1986outer}, iteratively tightens a piece-wise linear lower approximation of the objective function. By doing so, they solve to optimality sparse regression problems with $p\sim10^5$. {\color{changecolor}\cite{bertsimas2021slowly} consider a variant of the sparse regression problem, namely, slowly varying sparse regression, and develop a highly optimized version of the outer approximation method by utilizing a novel convex relaxation of the objective function; their technique can be directly applied to the standard sparse regression problem and scale it to larger instances.} \cite{hazimeh2020sparse} develop a specialized, nonlinear branch and bound framework that exploits the structure of the sparse regression problem (e.g., they use tailored heuristics to solve node relaxations) and are able to solve problems with $p\sim8\cdot10^6$. In our work, we use the solution method by \cite{bertsimas2020sparse}, but we remark that the method by \cite{hazimeh2020sparse} can also be used in combination with our proposed framework.

From a practical point of view, one argument commonly used against best subset selection is that, in real-world problems, the actual support size $k$ is not known and needs to be thoroughly cross-validated hence resulting in a dramatic increase in the required computational effort. Although, in many cases, $k$ is determined by the application, \cite{kenney2018efficient} address such concerns by proposing efficient cross validation strategies. In this paper, we assume that $k$ is known and given; the combination of efficient cross validation procedures with our proposed method is straightforward. 

% \begin{sloppypar}
% The success of MIO-based methods in sparse regression motivated the development of similar methods in a large variety of sparse machine learning problems, including classification \citep{bertsimas2017sparseclassif}, polynomial regression \citep{bertsimas2020sparsepoly}, matrix completion \citep{bertsimas2018interpretable}. In fact, the cutting planes method used in all the aforementioned approaches was eventually cast into a generic, unified framework that addresses a large variety of MIO problems \citep{bertsimas2019unified}.
% \end{sloppypar}

\paragraph{Heuristic Solution Methods and Surrogate Formulations.} 
Traditionally, the exact sparse regression formulation has been addressed via heuristic and greedy procedures (dating back to \cite{beale1967discarding,efroymson1966stepwise,hocking1967selection}). {\color{changecolor} Recently, numerous more sophisticated methods have been proposed, including the boolean relaxation of \cite{pilanci2015sparse}, the first-order method of \cite{bertsimas2016best}, the subgradient method of \cite{bertsimas2020sparse} and \cite{bertsimas2020sparseempirical}, the method of \cite{hazimeh2020fast} that combines coordinate descent with local search, and the approach of \citep{boyd2011distributed} that is based on the alternating direction method of multipliers.}

A lot of effort has been dedicated to developing and solving surrogate problems; the most studied of them all is the lasso formulation \citep{tibshirani1996regression}, whereby the non-convex $\ell_0$ norm is replaced by the convex $\ell_1$ norm. Namely, denoting by $\| \*w \|_1 = \sum_{j=1}^p | w_j |$, we require that $ \| \*w \|_1 \leq t$ or add a penalty term $ \lambda \| \*w \|_1 $ in the objective. Due to convexity, the constrained and the penalized versions can be shown to be equivalent for properly selected values of $t$ and $\lambda$; this is not the case if the $\ell_0$ norm is used. Due to the geometry of the unit $\ell_1$ ball, the resulting formulations indeed shrink the coefficients toward zero and produce sparse solutions by setting many coefficients to be exactly zero. There has been a substantial amount of algorithmic work on the lasso formulation \citep{efron2004least,beck2009fast,friedman2010regularization} and its many variants (e.g., the elastic net formulation of \cite{zou2005regularization}), on replacing the $\ell_1$ norm in the Lasso formulation by other sparsity inducing penalties (e.g., \cite{fan2001variable,zhang2010nearly}), and on developing scalable implementations \citep{friedman2009glmnet}.

\paragraph{Sparse Regression in Ultra-High Dimensions: Screening.} 
{\color{changecolor} Sure independence screening (SIS)} is a two-phase learning framework, introduced by \cite{fan2008sure,fan2014sure}. In the first phase, SIS ranks the features based on their marginal utilities; e.g., for linear regression, the proposed marginal utility is the correlation between each feature and the response. By keeping only the highest-ranked features, SIS guarantees that, under certain conditions, all relevant features are selected with high probability. The screening procedure can be implemented iteratively, conditioned on the estimated set of features from the previous step. In the second phase, SIS conducts learning and inference in the reduced feature space consisting only of the selected features, {\color{changecolor} using surrogate formulations such as lasso}. Among the various other extensions of SIS, we emphasize the works of \cite{fan2009ultrahigh} and \cite{fan2010sure} that extend SIS to generalized linear models, as well as \cite{fan2011nonparametric} and \cite{ni2016entropy} that extend SIS beyond the linear model. {\color{changecolor} \cite{atamturk20safe} propose safe screening rules for the exact sparse regression formulation, derived from a convex relaxation solution; such rules eliminate features based on guarantees that a feature may or may not be selected in an optimal solution. In this paper, we aim to address substantially bigger problems, so our proposed framework does not require solving a convex relaxation of the entire problem.}

\paragraph{Sparse Regression in Ultra-High Dimensions: Distributed Approaches.}  
\begin{sloppypar}
An alternative path to address large scale sparse regression problems is using {\color{changecolor} distributed methods}. The ADMM framework \citep{boyd2011distributed} can be used to fit, in a distributed fashion, a regression model on vertically partitioned data, provided that the regularizer is separable at the level of the blocks of features; this is, e.g., the case in lasso. The DECO framework of \cite{wang2016decorrelated}, after arbitrarily partitioning the features, performs a decorrelation step via the singular value decomposition of the design matrix,
fits a lasso model in each feature subset, and combines the {\color{mr}estimated coefficients centrally}. Several hybrid {\color{mr}methods which} combine screening with ideas from the distributed literature have also been developed \citep{yang2016feature,zhou2014parallel,song2015split}.

%\citep{bertsekas1989parallel}. The majority of the distributed statistical estimation literature focuses on subsampling data points (horizontally partitioning the data), i.e., assigning a subset of data points to each computing node and combining the results. Examples of such approaches include lasso \citep{mateos2010distributed,lee2017communication} and empirical risk minimization \citep{zhang2013communication}. Nevertheless, in the ultra-high dimensional regime that we consider, subsampling features (vertically partitioning the data) seems more suitable.
\end{sloppypar} 

% The ADMM framework \citep{boyd2011distributed} can be used to fit, in a distributed fashion, a regression model on vertically partitioned data, provided that the regularizer is separable at the level of the blocks of features; this is, e.g., the case in lasso. The DECO framework of \cite{wang2016decorrelated}, after arbitrarily partitioning the features, performs a decorrelation step via the singular value decomposition of the design matrix and, as a result, each subset of data after feature partitioning, can produce consistent estimates (despite having missing features within each subset). Each computing node locally fits a lasso model, before the computed coefficients are centrally combined (and possibly refined). Several hybrid {\color{mr}methods which} combine screening with ideas from the distributed literature have also been developed \citep{yang2016feature,zhou2014parallel,song2015split}.

\subsection{Advances in Decision Trees} \label{subsec:advances-trees}
In this section, we present recent advances in the decision tree problem. 

Decision trees are one of the most popular and interpretable methods in machine learning. At a high level, decision trees recursively partition the feature space into disjoint regions and assign to each resulting partition either a label, in the context of classification, or a constant or linear prediction, in the context of regression. The leading work for decision tree methods in classification is the classification and regression trees framework (CART), proposed by \cite{breiman1984classification}, which takes a top-down approach to determining the partitions. Briefly, the CART method operates in two phases: 
\begin{itemize}
    \item[-] A top-down induction phase, where, starting from a root node, a split is determined by minimizing an ``impurity measure'' (e.g., entropy), and then recursively applying the process to each of the resulting child nodes until no more splits are possible.
    \item[-] A pruning phase, where the learned tree is pruned to penalize more complex structures that might not generalize well.
\end{itemize}
The tree induction process of CART has several limitations. First, the process is one-step optimal and not overall optimal. Second, it does not directly optimize over the actual objective (e.g., misclassification error). {\color{mr}Third, although \cite{breiman1984classification} discuss multivariate splits (i.e., splits that involve multiple features) in their original work (dubbed ``variable combinations''), their proposed approach in identifying such splits is greedy and computationally inefficient. As a result, most modern implementations of CART only include univariate splits (i.e., splits that involve a single feature); see, e.g., \cite{scikit-learn}.} %it only considers univariate splits, i.e., splits that involve a single feature.
Despite of the aforementioned limitations, {\color{mr}CART or variants thereof have been} widely used as building blocks in ensemble learning methods based on bagging and boosting; examples include the random subspace method \citep{ho1998random}, random forest \citep{breiman2001random}, and, more recently, xgboost \citep{chen2016xgboost}. Indeed, such methods enhance the performance of the resulting classifier, sacrificing, however, the interpretability of the model. 

\cite{bertsimas2017optimal} formulate the decision tree problem using MIO and propose a tailored coordinate descent-based solution method. The resulting optimal trees framework (OT) overcomes many of the limitations of CART (optimization is over the actual objective, multivariate splits are possible), while still leading to highly interpretable models. The key observation that led to OT is that, when learning a decision tree, there is a number of discrete decisions and outcomes we need to consider:
\begin{itemize}
    \item[-] Whether to split at any node and which feature to split on.
    \item[-] Which leaf node a point falls into and (for classification problems) whether this point is correctly classified based on its label.
\end{itemize}
At each branch node, a split of the form $\*a^T \*x < b$ is applied. Points that satisfy this constraint follow the left branch of the tree, whereas those that violate the constraint follow the right branch. Each leaf node is assigned a label, and each point is assigned the label of the leaf node into which the point falls. The resulting MIO formulation aims to make the aforementioned decisions in such a way that the linear combination of an error metric (e.g., misclassification error) and a tree complexity measure is minimized. The formulation also includes a number of constraints ensuring that the resulting tree is indeed valid and consistent. 
{\color{changecolor} The resulting MIO formulation is solved using a tailored coordinate descent approach; the problem is nonconvex, so the solution process is repeated from a variety of starting trees that are generated randomly and, in the end, the one with the lowest overall objective function is selected.}{\color{mr} A long stream of literature has followed the original work by \cite{bertsimas2017optimal} in trying to optimally or near-optimally solve the decision tree problem, including, e.g., \cite{hu2019optimal,verwer2019learning,aghaei2020learning,blanquero2020sparsity,blanquero2021optimal}. The aforementioned works develop a variety of optimization-based approaches to the decision tree problem, which range from flow-based MIO formulations to modeling using continuous optimization; we refer the interested reader to \cite{carrizosa2021mathematical} for a review. Our proposed framework can be used in combination with any of the above approaches.}

% The resulting MIO formulation is solved using a tailored coordinate descent approach, which starts with an initial tree solution and repeatedly visits the nodes of the tree one at time and in a random order. For each branch node, the algorithm considers deleting the split at that node and finding the optimal split to use at that node. For each leaf node, it considers creating a new split at that node. If any of these changes improve the overall objective value of the tree, then the modification is accepted. The algorithm continues until no possible improvements are found, meaning that this tree is a local minimum. The problem is nonconvex, and therefore, this coordinate descent process is repeated from a variety of starting trees that are generated randomly. From this set of trees, the one with the lowest overall objective function is selected as the final solution.

\paragraph{Decision Trees in Ultra-High Dimensions.} 
Decision trees are known to suffer from the curse of dimensionality and to not perform well in the presence of many irrelevant features \citep{almuallim1994learning}. Among the most notable attempts to scale decision trees to ultra-high dimensional problems is the recent work by \cite{liu2017making}, who develop a sparse version of the perceptron decision trees framework \citep{bennett2000enlarging} and solve problems with $p\sim10^6$ features.

{\color{mr}
\subsection{Ultra-High Dimensional Machine Learning Beyond Sparse Regression and Decision Trees}

Several popular machine learning models beyond sparse regression and decision trees have been modified to address ultra-high dimensional problems. \cite{peng2016error} investigate the statistical performance of the $\ell_1$-regularized SVM in the ultra-high dimensional regime and \cite{lian2017divide} propose and analyze a divide and conquer approach for solving it; \cite{liu2019sparse} develop an efficient method for solving sparse SVM with $\ell_0$ norm approximation, which yields sparser models. \cite{radovanovic2009nearest, liu2018doubly} study and propose solutions for the k-nearest neighbor classifier in ultra-high dimensional problems, where the number of distances that need to be calculated is huge, whereas \cite{wang2018randomized, yang2020pase} focus on the implementation of systems for the aforementioned problem.
}

\subsection{Backbones in Optimization} \label{subsec:review-backbones}
\cite{schneider1996searching}, \cite{walsh2001backbones}, and the many references therein use the term backbone of a discrete optimization problem 
to refer to a set of frozen decision variables that \emph{must be set to one in any optimal solution} {\color{mr}(assuming multiple optimal solutions exist)}. Thus, the backbone can be obtained as the intersection of all optimal solutions to the problem. For example, {\color{mr}using the aforementioned interpretation of a backbone,} in the satisfiability decision problem, the backbone of a formula is the set of literals which are true in every {\color{mr}feasible solution}. This definition is fundamentally different from our notion of a backbone and, from a practical perspective, identifying such a backbone can be as hard as solving the actual problem. In our approach, the term backbone refers to all variables that \emph{are set to one in at least one near-optimal solution}. Thus, the backbone can be obtained as the union of all near-optimal solutions. To the best of our knowledge, the notion of backbone that we examine in this paper first appeared in \cite{bertsimas2019online} in the context of online vehicle routing.

\subsection{Outline \& Contributions} \label{subsec:outline}
Our key contributions can be summarized as follows:
\begin{itemize}
    \item[-] We develop the backbone method, a novel, two-phase framework that, as we empirically show, performs highly effective feature selection and enables sparse machine learning models to scale to ultra-high dimensions. 
    As mixed integer optimization offers a natural way to model sparsity, we focus on MIO-based machine learning methods. 
    Nonetheless, our framework is developed in full generality and can be applied to any sparsity-inducing machine learning method.%, as well as to problems beyond machine learning that exhibit sparse optimal solutions.
    \item[-] We apply the backbone method to the sparse regression problem. {\color{changecolor} We show that, under certain assumptions and with high probability, the backbone set consists of the {\color{mr}truly relevant} features. Our computational results on both synthetic and real-world data indicate that the backbone method outperforms or competes with state-of-the-art methods for ultra-high dimensional problems, accurately scales to problems with $p\sim10^7$ in minutes and problems with $p\sim10^8$ in hours. In problems with $p\sim10^5$, where exact methods apply and hence we can compare with near-optimal solutions, the backbone method, by drastically reducing the problem size, achieves, faster (by $15-20\%$) and with fewer data points {\color{mr}(i.e., with a smaller number of training samples)}, the same levels of accuracy as exact methods, while providing optimality guarantees, albeit for the reduced problem.}
    \item[-] We apply the backbone method to the decision tree problem. {\color{changecolor} Our computational results indicate that the backbone method scales to problems with $p\sim10^5$ in minutes, outperforms CART, and competes with random forest, while still outputting a single, interpretable tree. In problems with $p\sim10^3$, the backbone method can accurately filter the feature set and compute, substantially faster (by $10-100$ times), decision trees with comparable out-of-sample performance to that of the decision tree obtained by applying the optimal trees framework to the entire problem.}
\end{itemize}

The structure of the paper is as follows. {\color{changecolor} In Section \ref{sec:backbone}, we develop the generic framework for the backbone method in a general supervised learning setting; we unfold the components of the method and discuss topics that include hyperparameter selection, termination, complexity, and parallel implementation. In Section \ref{sec:sr}, we apply the backbone method to the sparse regression problem, discuss several implementation aspects and challenges, and present a theoretical result on the method's accuracy, which can provide guidance in tuning the method's hyperparameters. In Section \ref{sec:trees}, we apply the backbone method to the decision tree problem and again discuss several implementation aspects and challenges. Section \ref{sec:results-synthetic} investigates, via experiments on synthetic datasets, the backbone method's scalability, performance in various regimes, and sensitivity to the method's hyperparameters and components. In Section \ref{sec:results-real}, we evaluate the method through computational experiments on real-world datasets. Finally, Section \ref{sec:concl} concludes the paper.}

%%%%%%%%%%%%%%%%%%%%%%%%%%%%%%%%%%%%%%%%%%%%%%%%%%%%%%%%%%%%%%%%%%%%%%%%%%%%%%%%%%%%%%%%%%%%%%%%%%%%%%%%%%%%%%
%%%%%%%%%%%%%%%%%%%%%%%%%%%%%%%%%%%%%%%%%%%%%%%%%%%%%%%%%%%%%%%%%%%%%%%%%%%%%%%%%%%%%%%%%%%%%%%%%%%%%%%%%%%%%%
%%%%%%%%%%%%%%%%%%%%%%%%%%%%%%%%%%%%%%%%%%%%%%%%%%%%%%%%%%%%%%%%%%%%%%%%%%%%%%%%%%%%%%%%%%%%%%%%%%%%%%%%%%%%%%

\section{The Backbone Method} \label{sec:backbone}
In this section, we describe the backbone method in a general supervised learning context. The main idea is that many real-world ultra-high dimensional machine learning problems are indeed sparse, which means that {\color{mr}a small subset of features is sufficient for performing regression or classification and hence the remaining features need not be considered when solving the actual problem's formulation.}%most features are uninformative for performing regression or classification and hence need not be considered when solving the actual problem's formulation. 
Therefore, the core of the two-phase backbone method is the construction of the \emph{backbone set}, which consists of features identified as potentially relevant. The two phases of the backbone method are:
\begin{enumerate}
    \item[1.] \textbf{Backbone Construction Phase:} Form a series of tractable subproblems that can be solved efficiently while still admitting sparse solutions, by reducing the dimension and/or relaxing the original problem. Construct the backbone set by collecting all decision variables that participate in the solution to at least one subproblem.
    \item[2.] \textbf{Reduced Problem Solution Phase:} Solve the reduced problem to near-optimality, considering only the decision variables that were included in the backbone set.
\end{enumerate}
The backbone method differs from existing feature selection methods in that we propose to filter uninformative features via a procedure closely related to the actual problem. For example, to perform feature selection for the {\color{mr}optimal} decision tree learning problem, we train more tractable decision trees in the subproblems and select those features that are identified as relevant within the subproblems' trees. A major difference between the backbone method and heuristic solution methods is that we aim to reduce the problem dimension as much as possible so that the resulting reduced problem can actually be solved to near-optimality. From an optimization perspective, the backbone method can loosely be viewed as a heuristic branch and price approach.

\subsection{A Generic Hierarchical Backbone Algorithm} \label{subsec:backbone-algorithm}
% {\color{red} (This has been rearranged.)} 
We proceed by applying the backbone method in a general sparse supervised learning setting (Algorithm \ref{alg:backbone}). Let $\*X \in \mathbb{R}^{n \times p}$ be the input data ($n$ and $p$ denote the number of data points and features respectively) and $\*y$ the vector of responses (for regression problems) or class labels (for classification problems). At the end of the backbone construction phase, the backbone set, which we denote by $\mathcal{B}$, must be small enough so that we can actually solve the resulting reduced problem. Therefore, we implement the backbone construction phase in Algorithm \ref{alg:backbone} hierarchically, in a bottom-up divide and conquer fashion, until the size of the backbone set is sufficiently small.

The notation used in Algorithm \ref{alg:backbone} also includes the following. The set $\mathcal{U}_t$ consists of all features that are candidates to enter the backbone during iteration $t$. Initially, all features are considered, so $\mathcal{U}_0=[p]=\{1,...,p\}$ (unless the \texttt{screen} function, which will be explained shortly, is used). When the data matrix $\*X$ is indexed by a set $\mathcal{C} \subseteq [p]$, denoted by $\*X_{\mathcal{C}}$, it is implied that only the features contained in $\mathcal{C}$ have been selected.

As part of the input to Algorithm \ref{alg:backbone}, we need to specify the application-specific functions that will be used within our framework. In particular, these include:
\begin{sloppypar}
\begin{itemize}
    \item[-] {\color{changecolor} Function \verb|screen|: The function used to eliminate features that are highly likely to be irrelevant. This function is optionally called before the backbone construction phase. The output of this function is the set of features that have not been eliminated, along with their marginal utilities.}
    \item[-] Function \verb|construct_subproblems|: The function used to construct more tractable subproblems. Since $p$ is assumed to be very large, the \verb|construct_subproblems| function acts by selecting a subset of features $\mathcal{P}$ for each subproblem. Thus, the output of this function is a collection of features for every subproblem.
    \item[-] Function \verb|fit_subproblem|: The model fitting function to be used within each subproblem. Fits a sparse model of choice (regression, trees, etc.) to the data that are given as input to the subproblem. This function has to be highly efficient for the backbone construction to be fast, so we propose that a relaxed/surrogate formulation or heuristic solution method is used. For simplicity in notation, we assume that all input hyperparameters (e.g., for lasso, the regularization weight/sequence of weights) are ``hidden'' within the definition of the function. The output of this function is the learned model.
    \item[-] Function \verb|extract_relevant|: The function that takes a sparse model as input and extracts all features that the model identifies as relevant.
    \item[-] Function \verb|fit|: The target fitting function that fits a sparse model of choice to the data given as input. As this function is to be applied to the (small) backbone set, it needs not be extremely efficient, so we propose that a method that comes with optimality guarantees is used. For simplicity in notation, we assume that all input hyperparameters are ``hidden'' within the definition of the function. The output of this function is the learned model.
\end{itemize}
\end{sloppypar}
{\color{changecolor} Algorithm \ref{alg:backbone} has a number of hyperparameters; in the sequel, we discuss how the hyperparameters can be tuned in practice (Section \ref{subsec:backbone-discussion}) and in theory (Section \ref{subsec:sr-theory}), and examine how sensitive the backbone method is with respect to each of them (Section \ref{subsec:results-synth-sensitivity}). The hyperparameters are the following:
\begin{itemize}
    \item[-] $M \in \mathbb{N}$: The number of subproblems to solve in each iteration (or hierarchy). {\color{mr}To speed up the algorithm and since after each iteration the number of features that are candidate to enter the backbone set decreases, we reduce the number of subproblems to solve in each iteration by a factor of 2.}
    \item[-] $\beta \in (0,1]$: The fraction of features to include in each subproblem's feature set.
    \item[-] $\alpha \in (0,1]$: The fraction of features to keep after applying the \texttt{screen} function.
    \item[-] $B_{\max} \in \mathbb{N}$: The maximum allowable backbone size.
\end{itemize}

}

\begin{algorithm*}
\caption{Generic Hierarchical Backbone Algorithm}
\label{alg:backbone}
\begin{algorithmic}
    \REQUIRE Data $(\*X,\*y)$, hyperparameters $(M,\beta,\alpha,B_{\max})$, functions (\verb|screen|, \verb|construct_subproblems|, \verb|fit_subproblem|, \verb|extract_relevant|, \verb|fit|).
    \ENSURE Learned model.
    \STATE $t \leftarrow 0$
    \STATE {\color{changecolor} $\mathcal{U}_0, \*s \leftarrow \verb|screen|(\*X,\*y,\alpha)$}
    \REPEAT
        \STATE $\mathcal{B} \leftarrow \emptyset$
        \STATE $(\mathcal{P}_m)_{m\in\left[\lceil \frac{M}{2^t} \rceil\right]} \leftarrow \verb|construct_subproblems|(\mathcal{U}_t,\*s,\lceil \frac{M}{2^t} \rceil,\beta)$
        \FOR{$m \in \left[\lceil \frac{M}{2^t} \rceil\right]$}
            \STATE $\text{model}_m \leftarrow \verb|fit_subproblem|(\*X_{\mathcal{P}_m},\*y)$
            \STATE $\mathcal{B} \leftarrow \mathcal{B} \cup \verb|extract_relevant|(\text{model}_m)$
        \ENDFOR
        \STATE $t \leftarrow t+1$
        \STATE $\mathcal{U}_t \leftarrow \mathcal{B}$
    \UNTIL{$|\mathcal{B}| \leq B_{\max}$ (or other termination criterion is met)}
    \STATE model $\leftarrow$ \verb|fit|$(\*X_{\mathcal{B}},\*y)$
    \RETURN model 
\end{algorithmic}
\end{algorithm*}

The backbone method provides a generic framework that can be used to address a large number of different problems. As illustrated {\color{mr}by the applications that we} investigate in the following sections, depending on the problem at hand, there are several design choices that affect the performance of the method. For example:
\begin{itemize}
    \item[-] How to form the subproblems, so that they are tractable and, at the same time, useful for the original problem? We address this question in Section \ref{subsec:backbone-construct} by developing a generic framework for regression and classification problems.
    \item[-] How to solve the subproblems and, in particular, what is the impact of applying heuristics within each subproblem for the quality of the backbone set? We examine these questions for the sparse regression and the decision tree problems in Sections \ref{sec:sr} and \ref{sec:trees}, respectively.
\end{itemize}

{\color{changecolor} 

\subsection{The \texttt{screen} Function: Sure Independence Screening at a Glance} \label{subsec:backbone-screen}
In this section, we describe the \verb|screen| component of the backbone method. We first briefly present the sure independence screening (SIS) framework \citep{fan2008sure}, which plays an important role in our approach.

SIS is a two-phase sparse learning framework. In the first phase, we rank the features based on their marginal utilities $s_j$ and {\color{mr}retain only the top $\lceil \alpha p \rceil$ ones, i.e., $$\mathcal{M} = \{ 1 \leq j \leq p: \ s_j \text{ is among the $\lceil \alpha p \rceil$ largest features' utilities} \}.$$} Given a convex loss function $\ell$ (e.g., ordinary least squares or logistic loss), we measure the marginal utility $s_j$ of each feature $ j \in [p]$ using the empirical maximum marginal likelihood estimator, i.e.,
\begin{equation} \label{eqn:screen}
    (\hat{w}_{0,j},\hat{w}_j) = \text{argmin}_{w^0_j,w_j} \frac{1}{n} \sum_{i=1}^n \ell( y_i, w^0_j + w_j X_{i,j} ).
\end{equation}
We then estimate the marginal utility of feature $j$ as either the magnitude of the maximum marginal likelihood estimator $s_j := | \hat{w}_j |$ or the minimum of the loss function $s_j := \frac{1}{n} \sum_{i=1}^n \ell( y_i,\hat{w}_{0,j} + \hat{w}_j X_{i,j} ).$
In practice, we usually choose {\color{mr}$\alpha$} using cross validation. In the second phase, we conduct learning and inference in the reduced feature space consisting only of features in ${\mathcal{M}}$. Under certain conditions, SIS possesses the sure screening property, which requires that all relevant features are contained in the set ${\mathcal{M}}$ with probability tending to one \citep{fan2008sure,fan2009ultrahigh,fan2010sure}.

% Following \cite{fan2010sure}, we measure the marginal utility $s_j$ of each feature using the empirical maximum marginal likelihood estimator. More specificially, given a convex loss function $\ell$, we compute, $\forall j \in [p],$
% \begin{equation}
%     (\hat{w}_{0,j},\hat{w}_j) = \text{argmin}_{w^0_j,w_j} \frac{1}{n} \sum_{i=1}^n \ell( y_i, w^0_j + w_j X_{i,j} ).
% \end{equation}
% We then estimate the marginal utility of feature $j$ as either the magnitude of the maximum marginal likelihood estimator $s_j := | \hat{w}_j |$ 
% or the minimum of the loss function $s_j := \frac{1}{n} \sum_{i=1}^n \ell( y_i,\hat{w}_{0,j} + \hat{w}_j X_{i,j} ).$

\begin{algorithm*}
\caption{Function \texttt{screen}}
\label{alg:screen}
\begin{algorithmic}
    \REQUIRE Data $(\*X,\*y)$, hyperparameter $\alpha$.
    \ENSURE Screened features $\mathcal{M}$, marginal utilities $\*s$.
    \STATE $ \bm{s} \leftarrow \bm{0}_p$
    \FOR{$j \in [p]$}
        \STATE $(\hat{w}_{0,j},\hat{w}_j) \leftarrow \text{argmin}_{w^0_j,w_j} \frac{1}{n} \sum_{i=1}^n \ell( y_i, w^0_j + w_j X_{i,j} )$
        \STATE $s_j \leftarrow | \hat{w}_j |$
    \ENDFOR
    \STATE $\bm{r} \leftarrow \texttt{arg\_sort}(\bm{s})$ \hfill $ \triangleright$ Return a permutation of $[p]$ that sorts $\bm s$ in decreasing order.
    \STATE $\mathcal{M} \leftarrow \{ r_i \}_{i=1}^{\lceil \alpha p \rceil }$  \hfill $ \triangleright$ Keep top $\lceil \alpha p \rceil$ features.
    \RETURN $\mathcal{M}, \*s $ 
\end{algorithmic}
\end{algorithm*}

We present the \texttt{screen} function that we use within our framework in Algorithm \ref{alg:screen}. Our proposed \texttt{screen} function computes the marginal utility of each feature and eliminates the $(1-\alpha)p$ lowest ranked features. Concerning the choice of the loss function $\ell$ in Equation \eqref{eqn:screen} or, more generally, the scoring function we use, we make the following remarks:
\begin{itemize}
    \item[-] In linear regression problems, we use the squared loss function \citep{fan2008sure}. By doing so, the marginal utility measure given by $s_j := | \hat{w}_j |$ simplifies, {\color{changecolor} under standard assumptions such as standardizing the data,} to the absolute marginal empirical correlation between feature $j$ and the response, namely, $s_j := |\widehat{\text{cor}} \left( \*X_j, \*y \right)|$ (note that $\widehat{\text{cor}}$ represents the empirical correlation). % which is the standard marginal utility measure for linear regression as per \cite{fan2008sure}
    
    \item[-] In binary classification problems, we use the logistic loss function, whereas in multi-class classification problems, the multi-category SVM loss function \citep{fan2009ultrahigh}.
    
    \item[-] In highly nonlinear problems, we use the nonparametric screening approach by \cite{fan2011nonparametric} or the entropy-based approach by \cite{ni2016entropy}, which is particularly suited for decision tree problems.
\end{itemize} 

This approach provides a unified scoring framework for both regression and classification problems. Since we only use \texttt{screen} as a preprocessing step, we do not perform cross validation and, instead, select a slightly larger value for $\alpha$; e.g., we pick $\alpha$ such that $\alpha p \sim 10 n$, which is consistent with the theoretical and empirical analysis of \cite{fan2008sure} and, as we empirically show, is a good choice in practice. As we discussed in the Introduction of this paper, by using feature marginal utility measures that satisfy the sure screening property, we obtain strong theoretical guarantees that we will not eliminate any relevant feature during this step.

\subsection{The \texttt{construct\_subproblems} Function} \label{subsec:backbone-construct}
In this section, we describe the \texttt{construct\_subproblems} component of the backbone method, which aims to construct more tractable subproblems. 

Having computed each feature's marginal utility as per Section \ref{subsec:backbone-screen}, we construct the feature set of subproblem $m \in M$ by sampling $\lceil \beta \alpha p \rceil$ features among the $\lceil \alpha p \rceil$ features that survived the screening step of Algorithm \ref{alg:screen}. Within each subproblem, we sample the features without replacement and with probability that increases exponentially according to each feature's marginal utility. Algorithm \ref{alg:construct_subproblems} provides pseudocode for the proposed approach.

\begin{algorithm*}
\caption{Function \texttt{construct\_subproblems}}
\label{alg:construct_subproblems}
\begin{algorithmic}
    \REQUIRE Candidate features $\mathcal{U}$, feature marginal utilities $\*s$, hyperparameters $M,\beta$.
    \ENSURE Subproblem feature sets $(\mathcal{P}_m)_{m\in[M]}$.
    \STATE $\bm{\pi} \leftarrow \bm 0_{|\mathcal{U}|}$
    \FOR{$j \in \mathcal{U}$}
        \STATE $s_j \leftarrow \frac{s_j}{\max_{i \in \mathcal{U}} s_i}$ \hfill $ \triangleright$ Normalize utilities.
        \STATE $\pi_j \leftarrow \exp(s_j+1)$
    \ENDFOR
    \STATE $ (\mathcal{P}_m)_{m\in[M]} \leftarrow (\emptyset)_{m\in[M]}$
    \FOR{$m \in [M]$}
        \STATE $\mathcal{P}_m \leftarrow \texttt{sample}(\mathcal{U},\beta |\mathcal{U}|,\bm{\pi})$ \hfill $ \triangleright$ Sample $\beta |\mathcal{U}|$ features from $\mathcal{U}$, each w/ prob $\propto \pi_j$.
    \ENDFOR
    \RETURN $(\mathcal{P}_m)_{m\in[M]}$ 
\end{algorithmic}
\end{algorithm*}

The main benefits of breaking the problem into subproblems, as per Algorithm \ref{alg:construct_subproblems}, after -possibly- applying the screening step of Algorithm \ref{alg:screen}, are as follows:
\begin{sloppypar}
\begin{itemize}
    \item[-] \emph{Tractability}: Since $p$ is assumed to be very large and $\lceil \alpha p \rceil$ can also be large, by selecting a subset of features $\mathcal{P}$ for each subproblem, we perform an additional dimensionality reduction step. In particular, assume that all $k$ relevant features survived the screening step and that we randomly sample each subproblem's $\lceil \beta \alpha p \rceil$ features. Then, in expectation, we will have $\beta k$ relevant features within each subproblem. Thus, the subproblems clearly become more tractable, since, even with an exhaustive search procedure, we end up having to check $\binom{\beta \alpha p}{\beta k}$ subsets of features, instead of the initial $\binom{\alpha p}{k}$ or $\binom{p}{k}$, which can be astronomically larger. 
    
    \item[-] \emph{Ensemble learning}: Our proposed approach, of sampling the features that form each subproblem's feature set with probability that increases exponentially according to each feature's marginal utility, is partly inspired by the exponential mechanism of \cite{mcsherry2007exponential}. We aim to bridge the gap between the popular random subspace method \citep{ho1998random} and its variants, whereby, within each subproblem, a feature subset is selected uniformly at random, and the deterministic subspace method by \cite{koziarski2017deterministic}, which ranks features according to a predefined function and selects the top ones. 
    
    Using our proposed approach, we still enjoy the benefits of ensemble learning, whereby each feature gets examined multiple times and as part of different feature sets, while, at the same time, we avoid having subproblems with few or no relevant features. The absence of relevant features from a subproblem's feature set makes the subproblem harder to solve, {\color{mr}since any relevant feature that is not sampled and hence not contained within the subproblem's feature set is essentially} viewed as noise within this subproblem. In Section \ref{subsec:results-synth-sensitivity}, we empirically show that our proposed approach has an edge over sampling features uniformly at random or with probabilities proportional to their screening scores.

\end{itemize}
\end{sloppypar}

\subsection{Discussion: Hyperparameters, Termination, Complexity, and Parallel Implementation} \label{subsec:backbone-discussion}
In this section, we discuss several other aspects of the backbone method: how we choose the hyperparameter values in practice, how we ensure the algorithm's termination, what is the computational complexity of the method, and how the algorithm can be implemented in parallel.

\paragraph{Hyperparameter selection.} The hyperparameters $M, \beta, \alpha, B_{\max}$ of the backbone method can in practice be tuned via cross validation using a multi dimensional grid search. However, we remark that, in practical applications, $\beta$ and $B_{\max}$ are partly determined by the available computational resources (e.g., available memory) and, more specifically, by the size of the problems that we can solve using the \verb|fit_subproblem| function and the \verb|fit| function, respectively. The hyperparameter $\alpha$, which determines how many features the \texttt{screen} function will eliminate, is also partly dependent on the available resources, although we do have more freedom in tuning it; \cite{fan2008sure} provide both theoretical and empirical insights on how to choose $\alpha$. Finally, the number of subproblems $M$ can be selected dynamically: we pick a large value for $M$; we keep solving subproblems and add their solutions to the backbone set; if the backbone set does not change after two consecutive subproblems, we stop. 

In Section \ref{subsec:sr-theory}, we provide theoretical guidance on how to choose $M$ and $\alpha$ as function of the remaining problem and algorithm parameters. In Section \ref{subsec:results-synth-sensitivity}, we present a detailed sensitivity analysis of the backbone method with respect to each of the hyperparameters.

\paragraph{Termination.} To ensure finite termination of Algorithm \ref{alg:backbone}, we require that the \texttt{fit\_subproblem} function returns a model with at most $k_{\max} \leq k$ relevant features. In the context of sparse regression, we can directly control $k_{\max}$. In the context of decision trees, we indirectly control $k_{\max}$ via the tree's depth. Further, we reduce the number of subproblems to be solved in each iteration by a factor of $2$. Therefore, the number of iterations that Algorithm \ref{alg:backbone} will perform is at most
$$ H = \left\lceil \log_2 \frac{M k_{\max}}{B_{\max}} + 1\right\rceil. $$

%  Provided that $Mk_{\max}$ is sufficiently small, the size of $\mathcal{U}$ reduces by a factor of $1-\frac{Mk_{\max}}{|\mathcal{U}_{t-1}|}$ after each iteration.

\paragraph{Computational Complexity.} We denote by $f_{\texttt{screen}}(n,p)$, $f_{\texttt{construct}}(p,\beta)$, and $f_{\texttt{solve}}(n,p)$ the complexity of the \texttt{screen} function, the \texttt{construct\_subproblems} function, and the \texttt{solve\_subproblem} function, respectively. Then, the overall complexity of the backbone construction phase of Algorithm \ref{alg:backbone} is
\begin{equation*}
\begin{split}
    & O \left[ f_{\texttt{screen}}(n,p) + \sum_{t=1}^H \frac{M}{2^{t-1}} f_{\texttt{construct}}(\alpha p,\beta) + \sum_{t=1}^H \frac{M}{2^{t-1}} f_{\texttt{solve}}(n,\alpha \beta p) \right] \\
    & = O \left[ f_{\texttt{screen}}(n,p) + M f_{\texttt{construct}}(\alpha p,\beta) + M f_{\texttt{solve}}(n,\alpha \beta p) \right] .
\end{split}
\end{equation*} 

As an example, consider a sparse regression problem where we use the empirical correlation between each feature and the response as the scoring function for \texttt{screen}, use a simple heap-sampling implementation for \texttt{construct\_subproblems}, and solve subproblems using the LARS algorithm for the lasso formulation \citep{efron2004least}. Then, the resulting complexity is $O \left[ np + M \alpha \beta p\log(\alpha p) + M (\alpha \beta p)^2 n \right]$.

}

\paragraph{Parallel \& Distributed Implementation.} 
An important feature of the backbone method is that it can be naturally executed in parallel by assigning different subproblems to different computing nodes and then centrally collecting all solutions and solving the reduced problem. This is particularly important in distributed applications, where a massive dataset is ``vertically partitioned'' among different nodes, i.e., each node has access to a subset of the features in the data. This regime, despite being common in practice, has not been considered much in the literature.

\section{The Backbone of Sparse Regression} \label{sec:sr}

Given data $\{(\*x_i,y_i)\}_{i\in[n]}$, where, $\forall i\in[n]$, $\*x_i \in \mathbb{R}^p$ (with $n \ll p$), and $y_i \in \mathbb{R}$ (for regression) or $y_i \in \{-1,1\}$ (for binary classification), and a convex loss function $\ell$, consider the sparse empirical risk minimization problem with Tikhonov regularization and an explicit sparsity constraint, outlined as
\begin{equation}
    \label{eqn:sr}
    \begin{split}
        \min_{\*w \in \mathbb{R}^p} & 
        \quad ~\sum_{i=1}^n \ell( y_i, \*w^T \*x_i ) + \frac{1}{2\gamma} \Vert \*w \Vert_2^2
        \qquad \mbox{ s.t. } 
        \quad \Vert \*w \Vert_0 \leq k.
    \end{split}
\end{equation}
Problem (\ref{eqn:sr}) can be reformulated as a MIO problem, by introducing a binary vector encoding the support of the regressor $\*w$. The optimal cost for the equivalent formulation is then given by
\begin{equation}
    \label{eqn:sr-mio}
    \begin{split}
        \min_{\*z \in \{0,1\}^p: \sum_{j=1}^p z_j \leq k} 
        \quad \min_{\*w \in \mathbb{R}^p} & 
        \quad ~\sum_{i=1}^n \ell( y_i, \*w^T \*x_i ) + \frac{1}{2\gamma} \Vert \*w \Vert_2^2 \\
        \qquad \mbox{ s.t. }  &
        \quad z_j=0 \Rightarrow w_j=0 \quad \forall j \in [p].
    \end{split}
\end{equation}
As we already noted in the introduction, exact MIO {\color{mr}methods which} solve Problem (\ref{eqn:sr-mio}) scale up to $p \sim 10^5$. We next apply the backbone method to the sparse regression Problem (\ref{eqn:sr}).

During the backbone construction phase, we form the $M$ subproblems, whose feature sets are denoted by $\mathcal{P}_m,\ m \in [M]$, using the \verb|construct_subproblems| function (Algorithm \ref{alg:construct_subproblems}). Within the $m$-th subproblem, $m \in [M]$, we may use a subset of data points $\mathcal{N}_m \subseteq [n]$ and different hyperparameter values $k_{m}$ and $\gamma_{m}$. We form the backbone set as follows:
\begin{equation}
    \label{eqn:sr_backbone_construction}
    \begin{split}
        \mathcal{B} = \bigcup_{m=1}^M 
        \bigg \{ j: \quad
        & w_j^{(m)} \not= 0,  \\
        & \*w^{(m)} = \underset{\substack{ \*w \in \mathbb{R}^p:\\ \Vert \*w \Vert_0 \leqslant k_m,\\ w_j=0 \ \forall j \not\in \mathcal{P}_m}}{\text{argmin}}
         ~\sum_{i \in \mathcal{N}_m} \ell( y_i, \*w^T \*x_i ) 
        + \frac{1}{2\gamma_m} \Vert \*w \Vert_2^2 \bigg \}.
    \end{split}
\end{equation}
\begin{sloppypar}
We note that, instead of solving the sparse regression formulation in each subproblem, as shown in Equation (\ref{eqn:sr_backbone_construction}), there is a possibility of solving a relaxation or a closely-related surrogate problem, such as the elastic net formulation \citep{zou2005regularization}. We discuss this possibility in more detail in Section \ref{subsec:sr-implementation-backbone}.
\end{sloppypar}

Finally, once the backbone construction phase is completed, we solve Problem \eqref{eqn:sr}, considering only the features that are contained in the backbone set (i.e., exact solution computation in subset of backbone features), namely,
\begin{equation}
\label{eqn:sr_bb}
\begin{split}
\min_{\*w \in \mathbb{R}^p} & \quad ~\sum_{i=1}^n \ell( y_i, \*w^T \*x_i ) + \frac{1}{2\gamma} \Vert \*w \Vert_2^2 \\
\mbox{ s.t. } & \quad \Vert \*w \Vert_0 \leqslant k, \\
&  \quad w_j = 0 \quad \forall j \not\in \mathcal{B}.
\end{split}
\end{equation}

{\color{mr}\subsection{Implementation Details: Forming the Backbone Set} \label{subsec:sr-implementation-backbone}}
In this section, we discuss some additional implementation aspects of the backbone method for the sparse regression problem.\\

\paragraph{Solving Subproblems via the Sparse Regression Formulation.} 
Our first proposed approach to solve the subproblems relies on the actual sparse regression Formulation (\ref{eqn:sr}). In our implementation, we use the subgradient method of \cite{bertsimas2020sparseempirical}, which, {\color{mr}besides being fast}, is especially strong in recovering the true support; it should be clear that any of the exact or heuristic solution methods discussed in Section \ref{subsec:advances-sr} can be used. We empirically found this approach to be the most effective in terms of feature selection, namely, it achieves the lowest false detection rate within the backbone set.

A critical design choice in this approach concerns the \emph{number of relevant features that should be extracted from each subproblem}. Unless our sampling procedure is perfectly accurate, it is possible that some subproblems will end up containing $k_{\text{s}} < k$ relevant features. If this is the case, we empirically observed that solution methods for Problem (\ref{eqn:sr}) quickly recover the $k_{\text{s}}$ truly relevant features but struggle to select the remaining $k-k_{\text{s}}$, i.e., they spend a lot of time trying to decide which irrelevant features to pick. It is therefore crucial to either use cross validation within each subproblem (like the one we describe shortly), or apply an incremental selection procedure (e.g., forward stepwise selection). In our implementation, we use the following \emph{incremental cross validation scheme for the subproblem support size}, which relies on progressively fitting less sparse models. We start at a small number of relevant features $k_0$, increment to $k_1 = k_0 + k_{\text{step}}$, $k_2 = k_0 + 2 k_{\text{step}}$, and so forth. We stop after $i$ steps if the improvement in terms of validation error of a model with $k_i+k_{\text{step}}$ and $k_i+2k_{\text{step}}$ features is negligible compared to the error of a model with $k_i$ features. Crucially, when training a model with $k_i>k_0$ features during cross validation, we use the best model with $k_i-k_{\text{step}}$ features as a warm-start. 

Concerning the regularization parameter $\gamma$ in Formulation (\ref{eqn:sr}), we apply a simple grid search (with grid length $l$) cross-validation scheme, between a small value, e.g., $\gamma_0 = \frac{p}{k \ n \ \max_i \|\*x_i\|^2}$, and $\gamma_l=\frac{1}{\sqrt{n}}$, which is considered a good choice for most regression instances \citep{chen2012linear}.

\paragraph{Solving Subproblems via Randomized Rounding.} 
Instead of solving the MIO sparse regression Formulation (\ref{eqn:sr-mio}), one can consider its boolean relaxation, whereby the integrality constraints $z_i \in \{0,1\}$ are relaxed to $z_i \in [0,1]$ \citep{xie2020scalable}. The resulting solution $\hat{z}^{\text{rel}}$ will, in general, not be integral, so we propose to randomly round it by drawing $\hat{z}_i^{\text{bin}} \sim \text{Bernoulli}(\hat{z}_i^{\text{rel}})$. We form the backbone set by repeating the rounding procedure multiple times for each subproblem and collecting all features that had their associated binary decision variable set to $1$ in at least one realization of the Bernoulli random vector. By doing so, we reduce the number of subproblems we solve and, instead, perform multiple random roundings per subproblem. As a result, the overall method is sped up, sacrificing, however, its effectiveness in terms of feature selection. Moreover, in this approach, we cannot directly control the number of relevant features extracted from each subproblem, so we heuristically keep the features that correspond to the $k_{\max}$ largest regressors. For these reasons, we did not use this approach in the computational results we present.

\paragraph{Solving Subproblems via Surrogate Formulations.} 
In our second proposed approach to solve the subproblems, we formulate each subproblem using surrogate formulations, such as the lasso estimator \citep{tibshirani1996regression} and its extensions. In particular, we utilize the elastic net formulation \citep{zou2005regularization} to solve the $m$-th subproblem, $m \in [M],$ namely,
\begin{equation} \label{eqn:enet}
    \*w^{(m)} = \underset{\substack{\*w \in \mathbb{R}^{p}: \\ w_j=0 \ \forall j \not\in\mathcal{P}_m}}{\text{argmin}} \quad ~\sum_{i \in \mathcal{N}_m \subseteq [n]} \ell( y_i, \*w^T \*x_i ) + \lambda_m \left[ \mu_m \Vert \*w \Vert_1 + \tfrac{1-\mu_m}{2} \|\*w\|_2^2 \right].
\end{equation} 
The popularity of $\ell_1$-regularization is justified by the fact that it enjoys a number of properties that are particularly useful in practical applications. First and foremost, its primary mission is to robustify solutions against noise in the data and, specifically, against feature-wise perturbations. Second, the convexity of the $\ell_1$-norm makes the task of optimizing over it significantly easier. Third, $\ell_1$-regularization provides sparser solutions than $\ell_2$-regularization. 
We refer the interested reader to \cite{bertsimas2018characterization} and \cite{xu2009robust} for a detailed discussion on the equivalence of regularization and robustification.

We pick the hyperparameters of the elastic net formulation via cross validation. We select $\mu$ using grid search in the $[0,1]$ interval. For each fixed $\mu$, we perform grid search for $\lambda \in [\lambda_{\min},\lambda_{\max}]$, where $\lambda_{\max}$ is the value of $\lambda$ that leads to an empty model and $\lambda_{\min}$ is the value of $\lambda$ that leads to a model consisting of $k_{\max}$ nonzero coefficients. Once the best elastic net hyperparameters are found, we refit and add to the backbone set those features whose associated coefficients are above a user-specified threshold.

% \paragraph{Comparisons.} 
% Extensive comparisons between the aforementioned methods can be found in \cite{hastie2017extended} and, more recently, in \cite{bertsimas2020sparseempirical}. A summary of the conclusions of \cite{bertsimas2020sparseempirical} is as follows. In terms of accuracy in recovering the true support, non-convex methods should be preferred over $\ell_1$-regularization, since they provide better feature selection. In terms of false detection rate, cardinality-constrained formulations improve substantially, even when the true support size $k$ is unknown. These empirical observations are aligned with recent theoretical work in statistics \citep{gamarnik2017high} which has identified regimes where Lasso fails to recover the true support even though support recovery is possible from an information theoretic point of view. Finally, as far as the effect of noise is concerned, while lasso performs poorly in low noise settings, it competes and sometimes dominates other methods as noise increases; this observation supports the view that $\ell_1$-regularization is, first and foremost, a robustness story \citep{bertsimas2018characterization}.

{\color{mr}
\subsection{Implementation Details: Solving the Reduced Problem} \label{subsec:sr-implementation-reduced}}

After having formed the backbone set, we solve Problem (\ref{eqn:sr_bb}) using the cutting planes method by \cite{bertsimas2020sparse}. As \cite{bertsimas2020sparseempirical} observed, the computational time required for the MIO Formulation (\ref{eqn:sr-mio}) to solve to provable optimality is highly dependent on $\gamma$; for smaller values of $\gamma$, problems with $p\sim10^5$ can be solved in minutes, whereas, for larger values, it might take a huge amount of time to solve to provable optimality (although the optimal solution is usually attained fast). To address this issue, we impose a time limit on the cutting planes method for each value of $\gamma$ during the cross validation process {\color{mr}(typically in the order of minutes)}. In practice, we observe that the cutting planes method applied to the backbone set recovers the correct support in seconds (provided that the backbone set is sufficiently small{\color{mr}, i.e., it consists of few hundreds of features}). \\

{\color{changecolor}
\subsection{Theoretical Justification}\label{subsec:sr-theory}

In this section, we present a theoretical result, which guarantees that, under certain conditions and with high probability, the {\color{mr}truly relevant} features are selected in the backbone set. Our result, given in Theorem \ref{theo:backbone}, theoretically justifies the proposed approach and, most importantly, as explained above, provides guidance for the selection of the hyperparameters of the method.

{\color{mr}
\paragraph{Model and Assumptions.} We consider the model outlined in Assumption \ref{asmn:model}: 
\begin{asmn} \label{asmn:model}
    Let $p > n \geq k > 0$ be integers. Further, let
    \begin{itemize}
        \item[-] $\bm X \in \mathbb{R}^{n\times p}$ be a random design matrix such that each row $\bm x_i, i \in [n],$ is an iid copy of the random vector $X = (X_1,...,X_{p}) \sim \mathcal{N}(0,\bm I_{p})$.
        \item[-] $\bm \beta \in \{-1,0,1\}^{p}$ be fixed but unknown regressors that satisfy $\|\bm \beta \|_0 = k$ and let $\mathcal{S}^{\text{true}}$ denote the set of indices that correspond to the support of the true regressor $\bm \beta$.
        \item[-] $\bm y = \bm X \bm \beta + \bm \varepsilon$ be a response vector where the noise term $\bm \varepsilon$ consists of iid entries $\varepsilon_i \sim \mathcal{N}(0,\sigma^2), i \in [n].$
    \end{itemize}
\end{asmn}
%We have a random design matrix $\bm X \in \mathbb{R}^{n\times p}$ such that each row $\bm x_i, i \in [n],$ is an iid copy of the random vector $X = (X_1,...,X_{p}) \sim \mathcal{N}(0,\bm I_{p})$. 
% We have fixed but unknown regressors $\bm \beta \in \{-1,0,1\}^{p}$ that satisfy $\|\bm \beta \|_0 \leq k$. 
% Let $\mathcal{S}^{\text{true}}$ denote the set of indices that correspond to the support of the true regressor $\bm \beta$. 
% The response vector is $\bm y = \bm X \bm \beta + \bm \varepsilon$ where the noise term $\bm \varepsilon$ consists of iid entries $\varepsilon_i \sim \mathcal{N}(0,\sigma^2), i \in [n].$ Thus, each entry in $\bm Y$ is distributed as $Y = X^{\top} \bm \beta + \varepsilon \sim \mathcal{N}(0,k+\sigma^2)$ (assuming that $\|\bm \beta \|_0 = k$). 
% We further assume that $p>n\geq k$ and $\log p = O(n^\xi)$ for some $\xi \in (0,1).$ 
When we make asymptotic arguments, we take $p \rightarrow \infty.$ We analyze a simplified version of the backbone method, which we outline in Appendix \ref{sec:appendix-proof}. Our goal is to show that, with high probability, $\mathcal{S}^{\text{true}} \subseteq \mathcal{B},$ that is, the truly relevant features are selected in the backbone set $\mathcal{B}$.

The model we examine is indeed very simple and rather unrealistic; nevertheless, such assumptions are standard for analyzing exact sparse regression methods (see, e.g., \cite{pilanci2015sparse,david2017high,reeves2019all,zadik2019computational,bertsimas2020sparse}). Moreover, by the following result, we primarily aim to provide guidance for the selection of the hyperparameters of the method. We believe that the main contribution of the paper is its practical relevance.

\paragraph{Main Result.} 
Our main result consists of two parts. 
First, we give conditions under which the probability that the \texttt{solve\_subproblem} function fails to recover all relevant features that are included in the feature set of an arbitrary subproblem converges to zero. 
Second, we give additional conditions under which the probability that the (simplified version of the) backbone method fails converges to zero; namely, with high probability, $\nexists j \in [p]$ such that $j \in \mathcal{S}^{\text{true}}$ and $j \not\in \mathcal{B}$.
Specifically, we prove the following:
% We denote by $\varepsilon_2$ an upper bound on the probability that the \texttt{solve\_subproblem} function fails to recover all relevant features that are included in the feature set of an arbitrary subproblem. We denote by $\mathcal{E}$ the event that the (simplified version of the) backbone method fails, namely, $\exists j \in [p]$ such that $j \in \mathcal{S}^{\text{true}}$ and $j \not\in \mathcal{B}$. We prove the following result:

\begin{theo} \label{theo:backbone}
    Consider the model described in Assumption \ref{asmn:model} and assume $\log p = O(n^\xi)$, for some $0<\xi<1$.
    If the fraction of features screened satisfies $0 < \alpha \leq 1$ and $\alpha = \Theta\left(\frac{n^{1-\phi}}{p}\right)$, for some $ \phi<1$, the fraction of features per subproblem satisfies $0 < \beta \leq 1,$ and the regularization weight satisfies $\gamma = \frac{1}{n}$, then, for sufficiently many samples $n$, %samples $n \geq (\sigma^2+2k) \log(\alpha p) \log(\beta \alpha p)$:
    \begin{itemize}
        \item[(a)] As $p\rightarrow \infty$, the probability $p_{\text{subproblem}}$ that the \texttt{solve\_subproblem} function fails to recover all relevant features that are included in the feature set of an arbitrary subproblem satisfies $p_{\text{subproblem}} \rightarrow 0$.
    \end{itemize}
    Further, if the number of subproblems satisfies $M = \Omega \left( \frac{\log \left(\alpha p k\right)}{\log\left( \frac{1}{1-\beta+\beta p_{\text{subproblem}}} \right)} \right)$, then
    \begin{itemize}
        \item[(b)] As $p\rightarrow \infty$, the probability $p_{\text{backbone}}$ that the backbone method fails satisfies $p_{\text{backbone}} \rightarrow 0$.
    \end{itemize}
\end{theo}

% \begin{theo} \label{theo:backbone}
%     Suppose that $p > n\geq k$, $\log p = O(n^\xi)$ for some $\xi \in (0,1)$,
%     and the data is generated as discussed above. 
%     Let $\alpha = O\left(\frac{n^{1-\phi}}{p}\right)$, for some $ \phi<1$, and $\gamma = \frac{1}{n}$.
%     Then, for all $\theta \geq 1$, for samples $n \geq \theta (\sigma^2+2k) \log(\beta \alpha p)$, it holds that $\varepsilon_2 = O( e^{-\theta}).$ 
%     Further, let $M = O \left( \frac{\log \left(\alpha p k\right)}{\log\left( \frac{1}{1-\beta+\beta \varepsilon_2} \right)} \right)$.
%     Then, $\mathbb{P}(\mathcal{E}) \rightarrow 0$ as $p\rightarrow \infty$, that is, the (simplified version of the) backbone method recovers with high probability the {\color{mr}truly relevant} features in the backbone set.
% \end{theo}

The proof is included in Appendix \ref{sec:appendix-proof}. 
This result asserts that the backbone method recovers with high probability the truly relevant features in the backbone set.
Furthermore, it provides guidance concerning the backbone method's hyperparameters. 
Specifically, we get the order of $\alpha$ so that the screening step does not miss any relevant feature. 
In addition, the numerator in the asymptotic expression for $M$ suggests that the number of subproblems increases logarithmically with the number of features that we sample from in each subproblem and with the number of relevant features. 
From the denominator in the asymptotic expression for $M$ and, specifically, the term $1-\beta$, we get that, since $\beta$ controls the subproblem size, the larger the subproblems are, the fewer subproblems we need to solve.
Finally, the term $\beta p_{\text{subproblem}}$ corresponds to the fact that, as we solve subproblems more accurately, we again need to solve fewer subproblems.
% {\color{red}
% Give values to parameters in theorem as fun only of p and interpret.
% }
}   
}

%%%%%%%%%%%%%%%%%%%%%%%%%%%%%%%%%%%%%%%%%%%%%%%%%%%%%%%%%%%%%%%%%%%%%%%%%%%%%%%%%%%%%%%%%%%%%%%%%%%%%%%%%%%%%%
%%%%%%%%%%%%%%%%%%%%%%%%%%%%%%%%%%%%%%%%%%%%%%%%%%%%%%%%%%%%%%%%%%%%%%%%%%%%%%%%%%%%%%%%%%%%%%%%%%%%%%%%%%%%%%
%%%%%%%%%%%%%%%%%%%%%%%%%%%%%%%%%%%%%%%%%%%%%%%%%%%%%%%%%%%%%%%%%%%%%%%%%%%%%%%%%%%%%%%%%%%%%%%%%%%%%%%%%%%%%%

\section{The Backbone of Decision Trees} \label{sec:trees}

Given data $\{(\*x_i,y_i)\}_{i\in[n]}$, where, $\forall i\in[n]$, $\*x_i \in \mathbb{R}^p$ (with $n \ll p$) and $y_i \in [K]$, decision trees recursively partition the feature space and assign a class label from $[K]$ to each partition. Formally, let $T$ be a decision tree, $\mathcal{T}_B$ the set of branch nodes and $\mathcal{T}_L$ the set of leaf nodes. At each branch node $t \in \mathcal{T}_B$, a split of the form $\*a_t^T \*x < b_t$ is applied, {\color{mr}where, typically, $\*{a_t} \in \{0,1\}^p$ and $\sum_{j=1}^p a_{tj} = 1$.} Each leaf node $l \in \mathcal{T}_L$ is assigned a class label, typically via a majority vote among the class labels $y_i$ of all data points that fall into leaf $l$ after traversing the tree. The function $N(l;\*X)$ counts the number of data points in leaf $l$ and $N_{\min}$ is a ``minbucket'' parameter that controls the minimum number of data points that are allowed to fall into any leaf. Finally, $g(T; \*X,\*y,\lambda) = \text{error}(T; \*X,\*y)+\lambda |T|$ is the objective and consists of two components. The first, $\text{error}(T; \*X,\*y),$ is typically the misclassification error of the tree $T$ on the training data $(\*X,\*y)$. The second, $|T|$, is typically the number of branch nodes in the tree $T$. Then, at a high level, the decision tree problem can be stated as
\begin{equation} \label{eqn:oct}
    T^{\star} = \text{argmin}_{T: \text{depth}(T) \leq D} 
    \quad  g(T; \*X,\*y,\lambda)
    \quad \text{ s.t. } N(l;\*X) \geq N_{\min} \ \forall l \in \mathcal{T}_L.
\end{equation}

Ideally, we would like to exactly solve Problem (\ref{eqn:oct}) and therefore obtain a tree that achieves global optimality; in practice, however, this is still beyond the reach of MIO solvers. The optimal classification trees (OCT) framework provides a MIO formulation along with a set of heuristics that enable us to approximately solve Problem (\ref{eqn:oct}) in significantly larger dimensions than what was known. Nevertheless, despite the success of OCTs, the number of features $p$ remains their primary bottleneck; currently, OCTs scale up to $p$ in the 1,000s. 

As we pointed in the introduction, the decision tree induction process has traditionally been addressed via scalable heuristic methods, such as CART. Nonetheless, CART's training process has little to do with the actual misclassification objective and, as a result, the algorithm often settles with trees that are far from optimal for the original problem. The popularity of decision tree classifiers led to their wide use in ensemble models, such as bagging and boosting. Random forest, for example, combines multiple independently trained decision trees and typically boosts their performance, enjoying many of the benefits of ensemble learning, at the cost, however, of sacrificing interpretability.

The aforementioned limitations of decision tree-based methods, along with the fact that decision trees are indeed sparse models, are our main motivations in developing a backbone method for OCTs. 

\paragraph{Relevant Features.} 
Intuitively, each split in a decision tree is associated with a feature $j \in [p]$, so there are at most $2^{D}-1$ relevant features in tree of max depth $D$. We define that feature $j$ relevant if at least one split performed on it. During the backbone construction phase, we again form $M$ distinct and tractable subproblems. 

\paragraph{Constructing Subproblems.} 
Similarly to regression, in the $m$-th subproblem, $m \in [M]$, we only consider the features included in the subproblem's feature set $\mathcal{P}_m$ (constructed using the \verb|construct_subproblems| function of Section \ref{subsec:backbone-construct}) and, possibly, randomly sample a subset of data points $\mathcal{N}_m \subseteq [n]$. Note that, in formulating each subproblem, we may use different hyperparameter values $\lambda_m, N_{\min,m}$ and $D_{m}$. 

\paragraph{Forming the Backbone Set.} 
Let us denote by $\mathcal{T}(\*X,\*y,D,N_{\min})$ the set of all feasible trees on input data $(\*X,\*y)$ and by $\*{a_t}$ the split performed at branch node $t$. Then, the backbone set can be written as the union of the solutions to all subproblems, namely,
\begin{equation}
    \label{eqn:oct_backbone_construction}
    \begin{split}
        \mathcal{B} = \bigcup_{m=1}^M 
        \bigg \{ j: \quad
        & \sum_{t \in \mathcal{T}_B^{(m)}} a^{(m)}_{tj} \geq 1,  \\
        & T^{(m)} = \underset{ \substack{T \in \mathcal{T}(\*X_{\mathcal{N}_m,\mathcal{P}_m},\\\*y_{\mathcal{N}_m},D_m,N_{\min,m})} }{\text{argmin}}
        % _{T \in \mathcal{T}(\*X_{\mathcal{N}_m,\mathcal{P}_m},\*y_{\mathcal{N}_m},D_m,N_{\min,m})} 
        \quad g(T; \*X_{\mathcal{N}_m,\mathcal{P}_m},\*y_{\mathcal{N}_m},\lambda_m) \bigg \}.
\end{split}
\end{equation}
Importantly, in solving each of the subproblems in (\ref{eqn:oct_backbone_construction}), we need not solve the OCT formulation, shown in (\ref{eqn:oct}); instead, we propose solving each subproblem using scalable heuristic methods, such as CART. In fact, we empirically found that applying the OCT framework to subproblems does not significantly improve the support recovery accuracy in the backbone set (i.e., fraction of relevant features included in the backbone set). We use cross validation within each subproblem $m \in [M]$ to tune the hyperparameters $D_m,N_{\min,m},\lambda_m$.

\paragraph{Solving the Reduced Problem.} 
Once the backbone set $\mathcal{B}$ is constructed, we solve the OCT Formulation (\ref{eqn:oct}), considering only the features in $\mathcal{B}$, i.e.,
\begin{equation} \label{eqn:oct_bb}
    T^{\star} = \text{argmin}_{T: \text{depth}(T) \leq D} 
    \quad  g(T; \*X_{\mathcal{B}},\*y,\lambda)
    \quad \text{ s.t. } N(l;\*X_{\mathcal{B}}) \geq N_{\min} \ \forall l \in \mathcal{T}_L.
\end{equation}

\paragraph{Extension to Optimal Classification Trees with Hyperplane Splits.} 
Instead of limiting the tree learning method to univariate splits, where, $\*{a_t} \in \{0,1\}^p$ and $\sum_{j=1}^p a_{tj} = 1$ for any branch node $t$, multivariate splits can also be used. If this is the case, it is important that the number of features that participate in each split is artificially constrained (which translates to a sparsity constraint on $\*{a_t}$), otherwise the backbone set will generally not be small enough to substantially reduce the problem dimension.

\paragraph{Connections with the Random Subspace Method and Random Forest.} 
Feature bagging methods have had significant success when applied to ensembles of decision trees. The random subspace method \citep{ho1998random} relies on training each decision tree in the ensemble on a random subset of the features instead of the entire feature set. In random forest \citep{breiman2001random}, a subset of the features is considered in each split of each decision tree. Our backbone method for OCTs relies on the \verb|construct_subproblems| function, so each tree induced during the backbone construction phase is also trained on a subset of features. Thus, our approach enjoys many of the benefits of feature bagging (e.g., parallelizeability), while, at the same time, its output is a single, interpretable tree. {\color{mr}Moreover, as our computational results suggest, our approach does not lose much in terms of predictive power compared to random forest.}

%%%%%%%%%%%%%%%%%%%%%%%%%%%%%%%%%%%%%%%%%%%%%%%%%%%%%%%%%%%%%%%%%%%%%%%%%%%%%%%%%%%%%%%%%%%%%%%%%%%%%%%%%%%%%%
%%%%%%%%%%%%%%%%%%%%%%%%%%%%%%%%%%%%%%%%%%%%%%%%%%%%%%%%%%%%%%%%%%%%%%%%%%%%%%%%%%%%%%%%%%%%%%%%%%%%%%%%%%%%%%
%%%%%%%%%%%%%%%%%%%%%%%%%%%%%%%%%%%%%%%%%%%%%%%%%%%%%%%%%%%%%%%%%%%%%%%%%%%%%%%%%%%%%%%%%%%%%%%%%%%%%%%%%%%%%%

{\color{changecolor}
\section{Computational Results on Synthetic Data} \label{sec:results-synthetic}
In this section, we investigate the performance of the backbone method on synthetic datasets generated according to ground truth models that are known to be sparse. We start by describing the data generating methodology, the metrics, and the algorithms that we use throughout this (as well as the following) section. Then, we explore the scalability and performance of the method, as well as the sensitivity to the method's hyperparameters and components. Our primary goal in this section is to shed light on the behavior of the proposed backbone method and not to benchmark the backbone method compared to state-of-the-art alternatives; this is, in fact, the focus of Section \ref{sec:results-real}.
}
% Show scalability compared to very scalable heuristic, while having perfect/good performance (can be checked since ground truth is known, e.g., support recovery).

\subsection{Data Generating Methodology} \label{subsec:results-synth-data}

{\color{changecolor} 
In all our synthetic experiments throughout this section, we generate data according to the following methodology. 
}

\paragraph{Design Matrix.} We assume that the \emph{input data} $\*X = (\*x_1 , ... , \*x_n)$ are i.i.d. realizations from a $p$-dimensional zero-mean normal distribution with covariance matrix $\mathbf{\Sigma}$, i.e., $\*x_i \sim \mathcal{N}(\mathbf{0}_p, \mathbf{\Sigma}), i \in [n]$. The covariance matrix $\mathbf{\Sigma}$ is parameterized by the correlation coefficient $\rho \in [0, 1)$ as $\Sigma_{ij} = \rho^{|i-j|}, \forall i,j \in [p]$. As $\rho \rightarrow 1$, the columns of the data matrix $\*X$, i.e., the features, become more alike which should impede the discovery of nonzero components of the true regressor $\*w_{\text{true}}$ by obfuscating them with highly correlated look-alikes. In our experiments, we focus on high correlation regimes (e.g., $\rho=0.6$ or even $\rho=0.9$).

\paragraph{Sparse Linear Regression Data.} For linear regression, the unobserved true regressor $\*{w_{\text{true}}}$ is constructed at the beginning of the process and has exactly $k$-nonzero components at indices selected uniformly without replacement from $[p]$. Likewise, the nonzero coefficients $\*{w_{\text{true}}}$ are drawn uniformly at random from the set $\{-1,+1\}$. We next generate the \emph{response vector} $\*y$, which satisfies the linear relationship $\*y = \*X \*{w_{\text{true}}} + \*{\varepsilon}$, where $\varepsilon_i, i \in [n],$ are i.i.d. noise components from a normal distribution, scaled according to a chosen signal-to-noise ratio $\sqrt{\text{SNR}} = \| \*X \*{w_{\text{true}}} \|_2 / \| \*{\varepsilon} \|_2$. Evidently as the $\text{SNR}$ increases, recovery of the unobserved true regressor $\*{w_{\text{true}}}$ from the noisy observations can be done with higher precision. 

\paragraph{Sparse Logistic Regression Data.} For logistic regression, the true regressor is constructed in the exact same manner. The signal $\*y$ is computed according to $$\*y = \text{sign}(\*X \*{w_{\text{true}}} + \*{\varepsilon}).$$ 

\paragraph{Classification Tree Data.} 
For classification trees, we first create a full binary tree $T_{\text{true}}$ (\emph{ground truth tree}) of given depth $D$. The structure of $T_{\text{true}}$ is determined as follows.
\begin{itemize}
    \item[-] \emph{Relevant features:} We randomly pick $k$ relevant features among the entire feature set. At each split node, we select a relevant feature to split on. {\color{mr}To ensure that all selected features are actually relevant,} we require that each of them appears in at least $r$ split nodes in the tree. {\color{mr}(In other words, the fact that we select a feature to be relevant, does not really guarantee that this feature will actually be relevant; e.g., it could only be used in a split node that only ``touches'' very few data points.)} Thus, the number of relevant features $k$ must satisfy $k \leq \frac{2^D-1}{r}$.
    \item[-] \emph{Split thresholds:} Within each split node, we randomly pick a split threshold, taking care to maintain consistency of feature ranges across paths in the tree {\color{mr}(e.g., if at node $t$ we split on feature $x \leq x_0$, then it would not make sense to pick a split threshold of $x_1>x_0$ for feature $x$ at node $t$'s left ancestors since we already know that $x\leq x_0$)}. Moreover, let $x \in [x_{\min},x_{\max}]$ be the feature on which we split at node $t$. To ensure that splits are reasonably balanced, we require that the split threshold $b_t \in [x_{\min}+\frac{x_{\max}-x_{\min}}{2} \cdot f,\ x_{\max} - \frac{x_{\max}-x_{\min}}{2} \cdot f],$ where $f \in [0,1]$ is the parameter that determines how balanced the splits are. In our experiments, we use $f=\frac{1}{2}.$
    \item[-] \emph{Labels:} We assign class labels to leaf nodes in such a way that no sibling leaves correspond to the same class. We denote by $\mathcal{C}$ the set of all class labels and by $c_l \in \mathcal{C}$ the class label of leaf $l$. 
    %Given a global ``noise'' parameter $\rho \in [0,1]$, we associate with each leaf $l$ a local ``noise'' parameter $\rho_l$ drawn randomly from $[0,\rho]$. $\rho_l$ corresponds to the probability of flipping the label of data points in leaf $l$. 
    Furthermore, let $K=|\mathcal{C}|$ the total number of classes. In our experiments, we use $K=2$ and hence examine binary classification problems.
\end{itemize}
Next, we generate data from $T_{\text{true}}$ by setting, for each $i \in [n]$, $y_i=c_l$, where $l$ is the leaf where data point $\*x_i$ falls after traversing the tree.
% Next, we generate data from $T_{\text{true}}$ as follows. Let $l$ be the leaf where data point $\*x_i,\ i \in [n],$ falls after traversing the tree. With probability $1-\rho_l$, we set $y_i=c_l$, whereas, with probability $\rho_l$, we draw $y_i$ uniformly at random from $\mathcal{C} \setminus \{c_l\}$.

\subsection{Metrics}\label{subsec:results-synth-metrics}

{\color{changecolor} 
In our computational study in this section and in Section \ref{sec:results-real}, we evaluate the quality of each method based on the following metrics: }
\begin{itemize}
    \item[-] \emph{Support recovery accuracy} (SR-ACC): Measures the fraction of the $k$ relevant features that were actually selected by the estimator. For example, in the context of regression, we have $$\dfrac{|\{j: w_j \neq 0, \; w_{\text{true},j} \neq 0\}|}{k}.$$ 
    \item[-] \emph{Support recovery false alarm rate} (SR-FA): Measures the ratio of number irrelevant features selected over total number of features selected.
    \item[-] \emph{Fraction of features used that are relevant:} Measures the ratio of the number of relevant features used over the total number of features used in the learned model. This metric simultaneously captures support recovery accuracy and model simplicity, and is particularly useful in decision tree models, whereby one feature might be part of the ground truth tree and yet have little impact on the classification task.
    \item[-] \emph{Prediction accuracy} (R$^2$ or AUC): Evaluates the out-of-sample performance of the estimator. We use the R$^2$ statistic for linear regression and the area under the curve (AUC) for logistic regression and for classification trees (since we deal with binary classification problems).
    \item[-] \emph{Optimality gap} (OG): Measures the gap between the lower and upper objective bound during the solution process of an MIO problem. {\color{mr}When presenting computational results on OG, we explicitly clarify which MIO formulation we are referring to (specifically, in the context of sparse regression, we state whether we are referring to the original Problem \eqref{eqn:sr-mio} or the reduced Problem \eqref{eqn:sr_bb}).}
    \item[-] \emph{Learned tree depth:} Reports the depth of the learned decision tree model.
    \item[-] \emph{Computational time} (T): Total amount of time used (in seconds).
\end{itemize}

Additionally, to assess the quality of each component of the backbone method, we also record the following statistics:
\begin{itemize}
    \item[-] {\color{mr}\emph{Support recovery accuracy in the backbone set}}: Measures the fraction of the $k$ relevant features that were selected in the backbone set. %Formally, $$\dfrac{|\{j: j \in \mathcal{B}, \; w_{\text{true},j} \neq 0\}|}{k}.$$
    \item[-] \emph{Backbone size}: Number of features included in the backbone set.
    % \item[-] \emph{Backbone construction time}: Time required for the first phase of the backbone method (in seconds).
    \item[-] {\color{mr}\emph{Support recovery accuracy in the $m$-th subproblem/in total after $m$ subproblems}: Measures the fraction of the $k$ relevant features that were selected in subproblem $m$/in total after $m$ subproblems.} %Formally, $$\dfrac{|\{j: j \in \bigcup_{m=1}^M \mathcal{P}_m, \; w_{\text{true},j} \neq 0\}|}{k}.$$
    % \item[-] \emph{Subproblem feature sets size}: Number of distinct features that were included in at least one subproblem.
\end{itemize}

\begin{sloppypar}
All experimental results were obtained over $10$ independently generated datasets. In each experiment, we report both the mean and standard deviation of each metric. All out-of-sample metrics were obtained from independently generated test sets of size $n_{\text{test}}=2,000.$
\end{sloppypar}

\subsection{Algorithms \& Software}\label{subsec:results-synth-algorithms}

{\color{changecolor} 
In this section, we summarize the algorithms and software that we use in our experiments. For the sparse regression problem, we use the following algorithms:
\begin{itemize}
    \item[-] \verb|SR|: Implementation of the cutting planes method by \cite{bertsimas2020sparse} in Julia using the commercial MIO solver Gurobi \citep{gurobi2016gurobi}. Uses the subgradient method \cite{bertsimas2020sparseempirical} to compute a warm start. Solves the sparse regression formulation to optimality or near-optimality. 
    \item[-] \verb|SR-REL|: Implementation of the subgradient method by \cite{bertsimas2020sparseempirical} in Julia using the commercial Interpretable AI software package \citep{InterpretableAI}. Solves the sparse regression heuristically by considering its boolean relaxation and iteratively alternating between a sub-gradient ascent step and a projection step.
    \item[-] \verb|ENET|: \verb|glmnet| Fortran implementation \citep{friedman2009glmnet}, using the Julia wrapper. Solves the elastic net formulation using cyclic coordinate descent.
\end{itemize}
For the decision tree problem, we use the following algorithms:
\begin{itemize}
    \item[-] \verb|OCT|: Implementation of the OCT-learning method from the OT framework \citep{bertsimas2017optimal} in Julia using the commercial Interpretable AI software package \citep{InterpretableAI}. Solves the decision tree formulation via a tailored local search procedure.
    \item[-] \verb|CART|: Implementation of the CART heuristic \citep{breiman1984classification} using the Julia wrapper for the scikit-learn package \citep{scikit-learn}. 
    \item[-] \verb|RF|: Implementation of the random forest classifier \citep{breiman2001random} using the Julia wrapper for the scikit-learn package \citep{scikit-learn}.
\end{itemize}
To address high-dimensional regimes, we consider the following {\color{mr}methods which} incorporate a feature selection component:
\begin{itemize}
    \item[-] \verb|SIS-ENET|: Implementation of the sure independence screening \citep{fan2008sure} feature selection heuristic in Julia, as per Section \ref{subsec:backbone-screen}, followed by \verb|ENET| on the reduced feature set.
    \item[-] \verb|RFE|: Implementation of the popular recursive feature elimination algorithm \citep{guyon2002gene} using a Julia wrapper for the scikit-learn package \citep{scikit-learn}. For sparse regression, we use linear regression to eliminate features, with a step of 100 features; then, we apply \verb|SR| on the selected features (tuned using cross-validation). For classification trees, we use \verb|CART| to eliminate features, with a step of 100 features; then, we apply \verb|CART| on the selected features (tuned using cross-validation).
    \item[-] \verb|DECO|: Implementation of the DECO framework by \cite{wang2016decorrelated} in Julia. We partition the feature space into $5$ subsets and, after the de-correlation step, we perform feature selection in each subset using lasso. Then, we apply \verb|ENET| on the selected features (tuned using cross-validation).
\end{itemize}
Finally, our proposed backbone method is implemented as outlined below:
\begin{itemize}
    \item[-] \verb|BB|: Implementation of the backbone method in Julia. The components and parameterization of the method are discussed in each experiment separately. We remark that, in the results that we present, we did not make an effort to fine-tune the method's parameters; instead, we selected them based on Theorem \ref{theo:backbone} and on empirical evidence.
\end{itemize}
All experiments were performed on a standard Intel(R) Xeon(R) CPU E5-2690 @ 2.90GHz running CentOS release 7. Moreover, all methods' hyperparameters are tuned using the holdout method for cross validation, whereby we split the training set into actual training and validation data at a $0.7$ ratio.
}

{\color{changecolor} 

\subsection{Scalability with the Number of Features}\label{subsec:results-synth-scalability-features}

In this experiment, we examine the scalability of \verb|BB| as the number of features $p$ increases. We show that \verb|BB| accurately scales to ultra-high dimensional problems and notably outperforms baseline heuristics.

\paragraph{Sparse Linear Regression.} We consider a sparse linear regression problem with $n=5,000$ data points, $k=50$ relevant features, $\text{SNR}=2$, and correlation $ \rho = 0.9.$ We vary the number of features $p \in \{10^6, 3\cdot10^6, \dots, 3\cdot10^8$\}.

We tune \verb|BB| as follows. We select the \texttt{screen} function's parameter $\alpha$ such that all but $4 \cdot n = 20,000$ features are eliminated. We set $\beta=0.5$ and solve $M=10$ subproblems. We set $B_{\max}=250.$ We solve the subproblems using \texttt{SR-REL}; in the $m$-th subproblem, we cross-validate 3 values for $k_m \in \{ \lceil \frac{k}{3} \rceil, \lceil \frac{2k}{3} \rceil, k \}$ and set $\gamma_m = \frac{1}{\sqrt{n_m}}$. We solve the reduced problem using \texttt{SR} with a time limit of 5 minutes; we cross-validate 5 values for the hyperparameter $\gamma$. As a baseline, we compare \verb|BB| with \verb|SIS-ENET|, whereby we select $\alpha p$ features using \verb|SIS| and then apply \verb|ENET|. We tune \verb|ENET| as described in Section \ref{subsec:sr-implementation-backbone} and, specifically, we cross-validate 5 values for the hyperparameter $\mu \in [0,1]$ (the pure lasso model is included in the cross-validation procedure). We discard from the final model any feature whose corresponding regressor has magnitude $\leq 10^{-6}$.

Figure \ref{fig:regression_scalability} presents the results for this experiment. \verb|BB| achieves near-perfect accuracy for problems with up to 300 million features and substantially outperforms \verb|SIS-ENET|, in terms of both support recovery accuracy (in that \verb|BB| recovers almost the entire true support with near zero false positives, whereas \verb|SIS-ENET| recovers the true support at the cost of a large number of false positives) and out-of-sample predictive performance. As far as the computational time is concerned, we observe that the overhead of \verb|BB| over \verb|SIS-ENET| is by no means prohibitive; we are able to solve problems with 10 million features in less than an hour and problems with 300 million features in less than 10 hours.

\begin{figure}[htbp] 
    \centering
    \subfigure[Support recovery accuracy]{\includegraphics[width=0.49\textwidth]{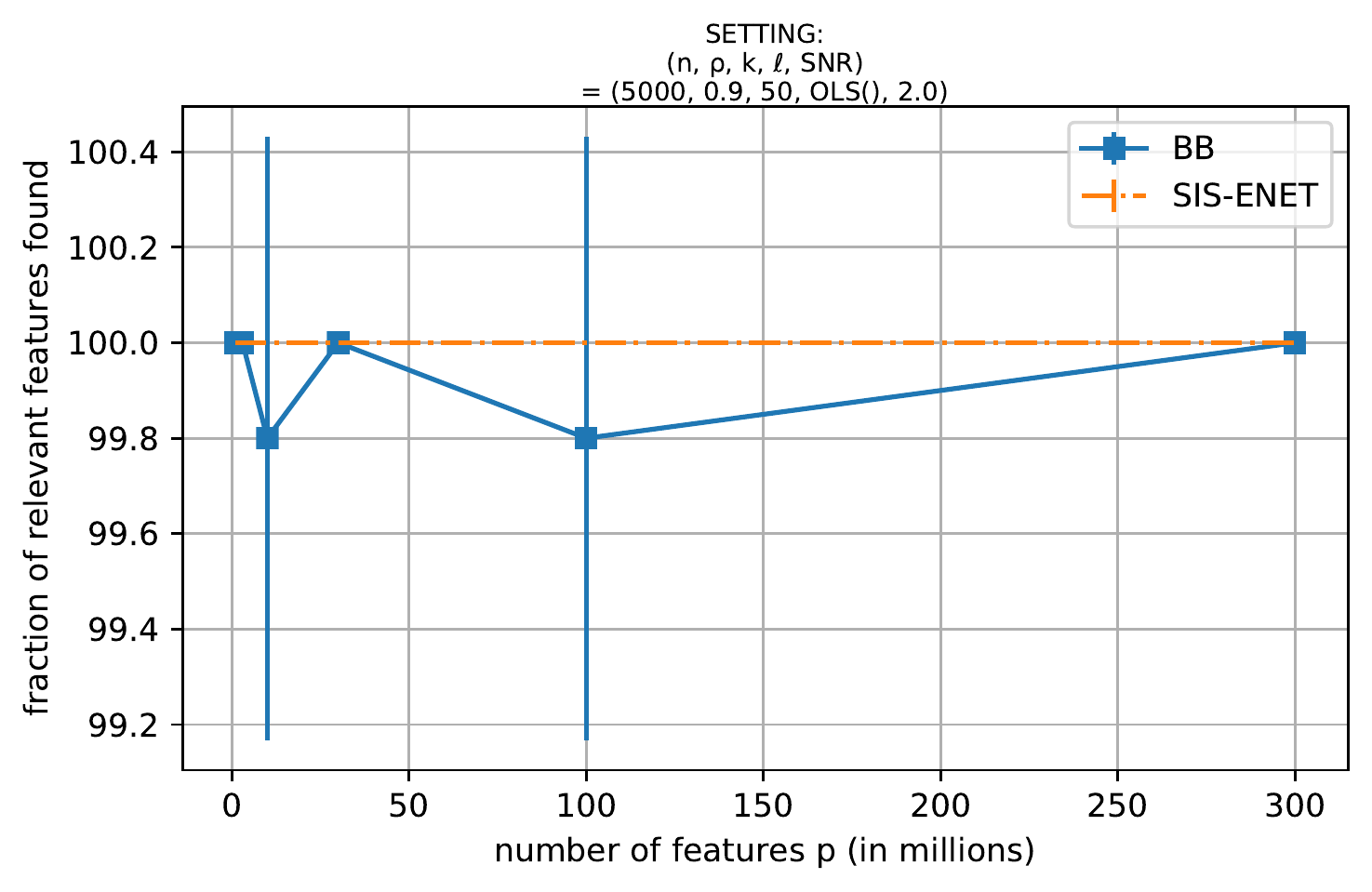}}
    \subfigure[Support recovery false alarm rate]{\includegraphics[width=0.49\textwidth]{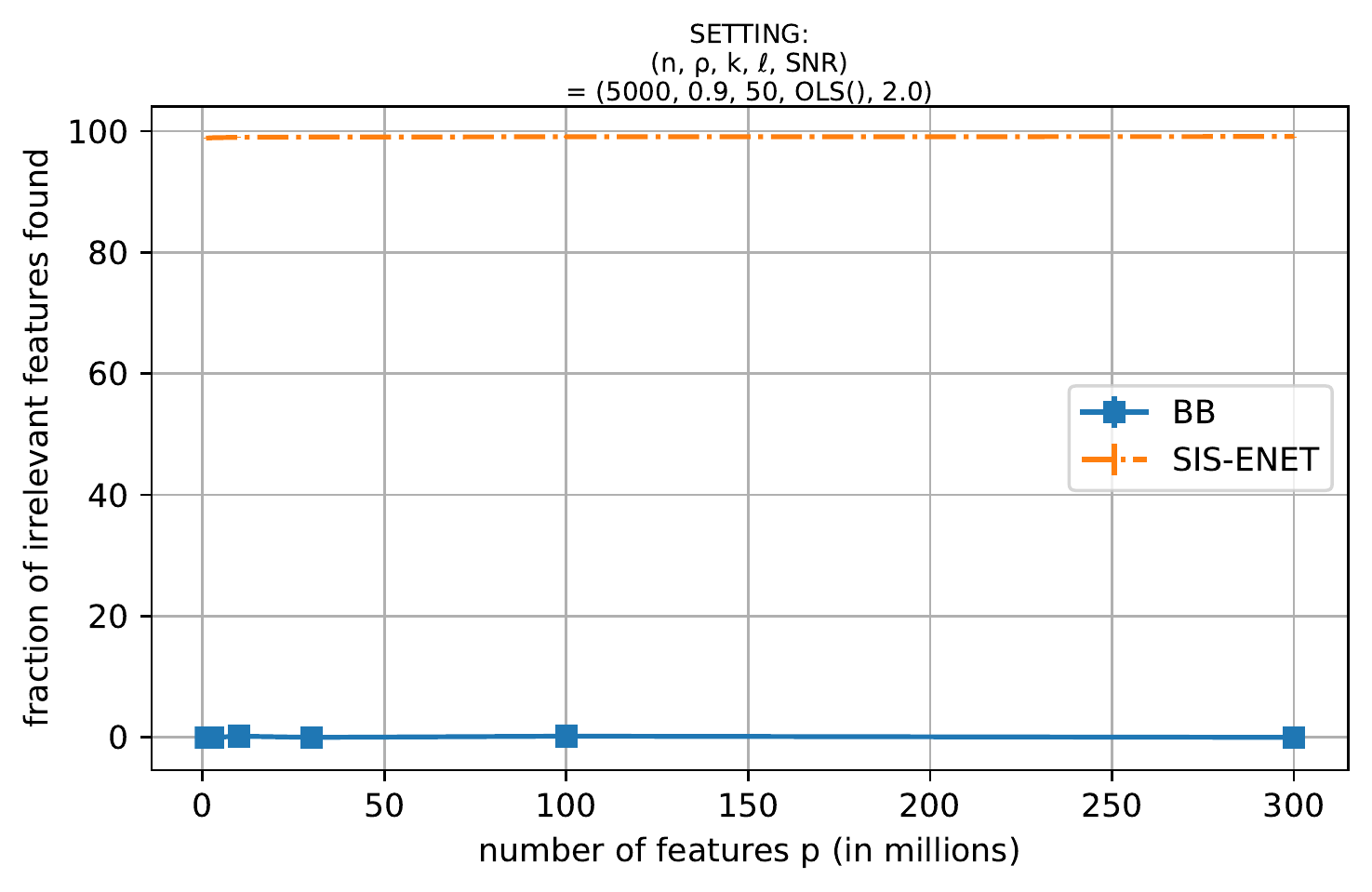}} 
    \subfigure[Out-of-sample R$^2$]{\includegraphics[width=0.49\textwidth]{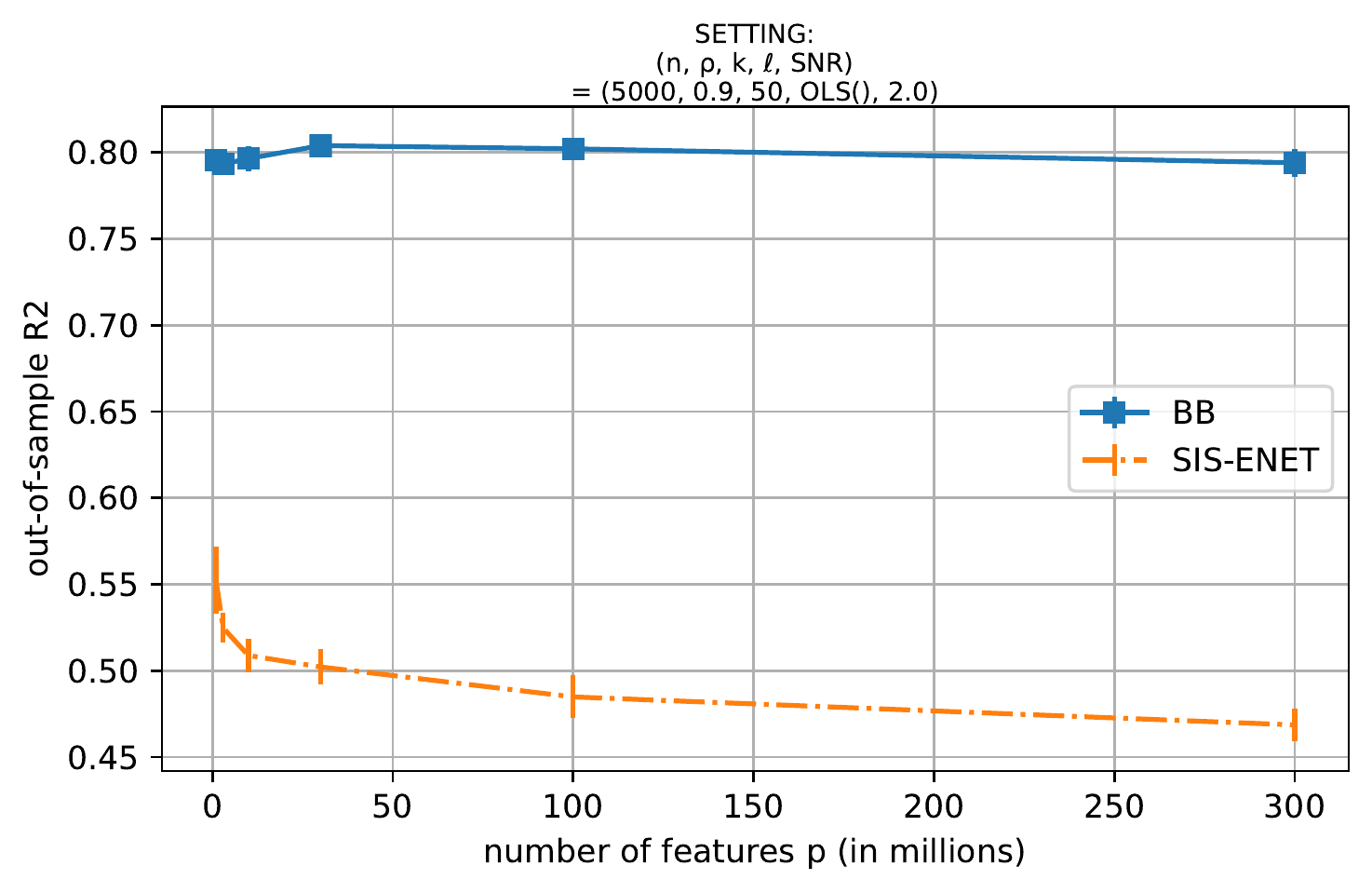}}
    \subfigure[Computational Time (sec)]{\includegraphics[width=0.49\textwidth]{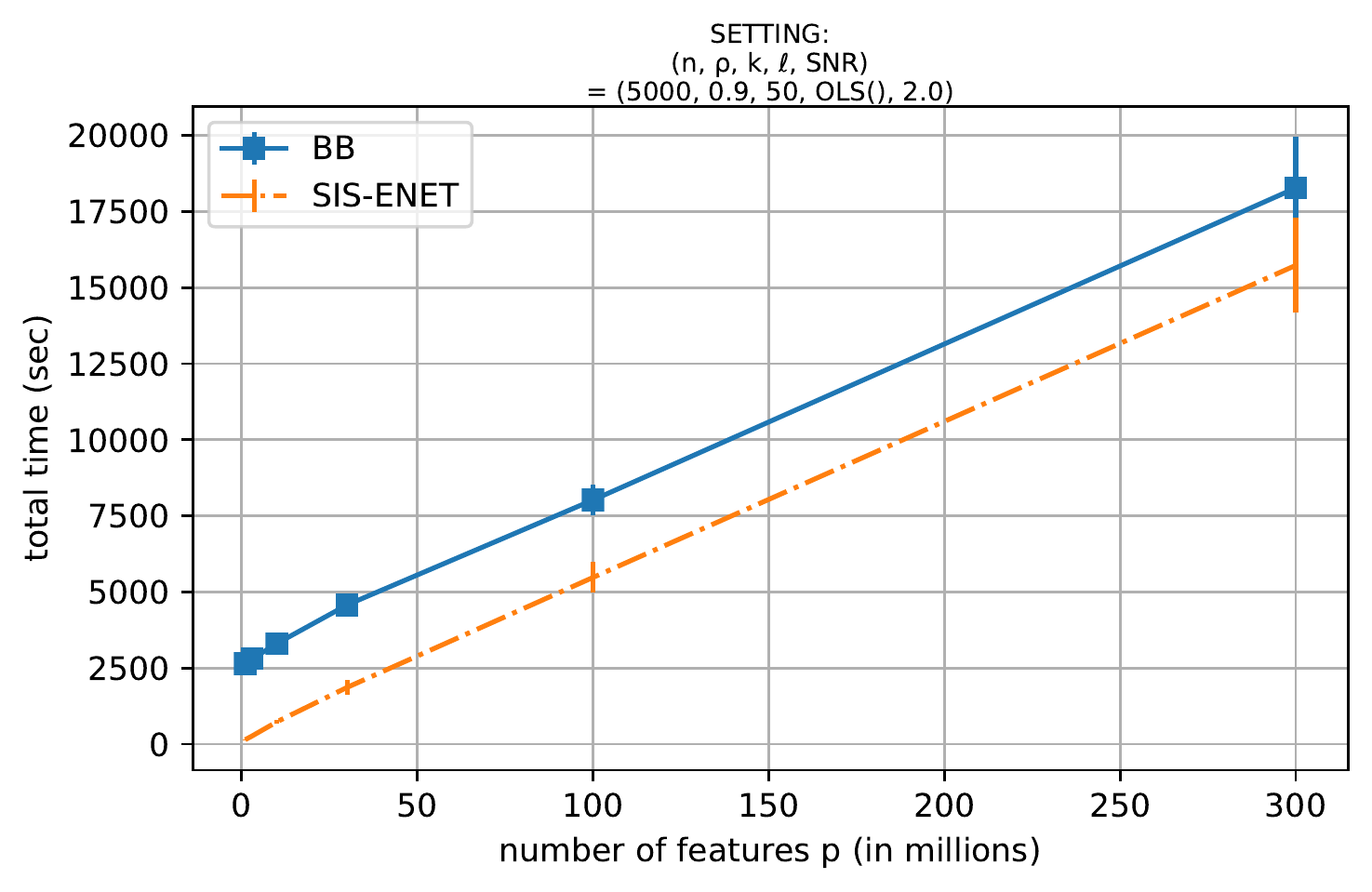}}
    \caption{Scalability with the number of features for sparse linear regression.}
    \label{fig:regression_scalability}
\end{figure}

\paragraph{Classification Trees.} We consider a classification tree problem with $n=5,000$ data points, tree depth of $D=5$ and $k=31$ relevant features, and correlation $\rho = 0.9.$ We vary the number of features $p \in \{2\cdot10^4, 4\cdot10^4, \dots, 10^5$\}.

We tune \verb|BB| as follows. We select the \texttt{screen} function's parameter $\alpha=1$ such that no features are eliminated. We set $\beta=0.1$ and solve $M=50$ subproblems. We set $B_{\max}=250.$ We solve the subproblems using \texttt{CART}; in the $m$-th subproblem, we cross-validate $D_m \in \{ 2, \dots, D-2 \}$ (we also cross-validate the minbucket and complexity parameter of \verb|CART|). We solve the reduced problem using \texttt{OCT}; we cross-validate $D \in \{ 3, \dots, D+2 \}$ (we also cross-validate the minbucket and complexity parameter of \verb|OCT|). As a baseline, we compare \verb|BB| with \verb|CART|, tuned in the exact same way as \verb|OCT| in solving the reduced problem for \verb|BB|.

As can be observed in Figure \ref{fig:trees_scalability}, \verb|BB| outperforms \verb|CART| in terms of out-of-sample predictive performance for problems with up to $100,000$ features. Moreover, among the features that are used in the split nodes of the learned classification tree, \verb|BB| selects a substantially higher fraction of relevant ones and, at the same time, the learned tree is simpler (i.e., of smaller depth). This particular configuration of \verb|BB| solves problems with $100,000$ features in approximately an hour and the computational time scales linearly with the number of features.

\begin{figure}[htbp] 
    \centering
    \subfigure[Fraction of features used that are relevant]{\includegraphics[width=0.49\textwidth]{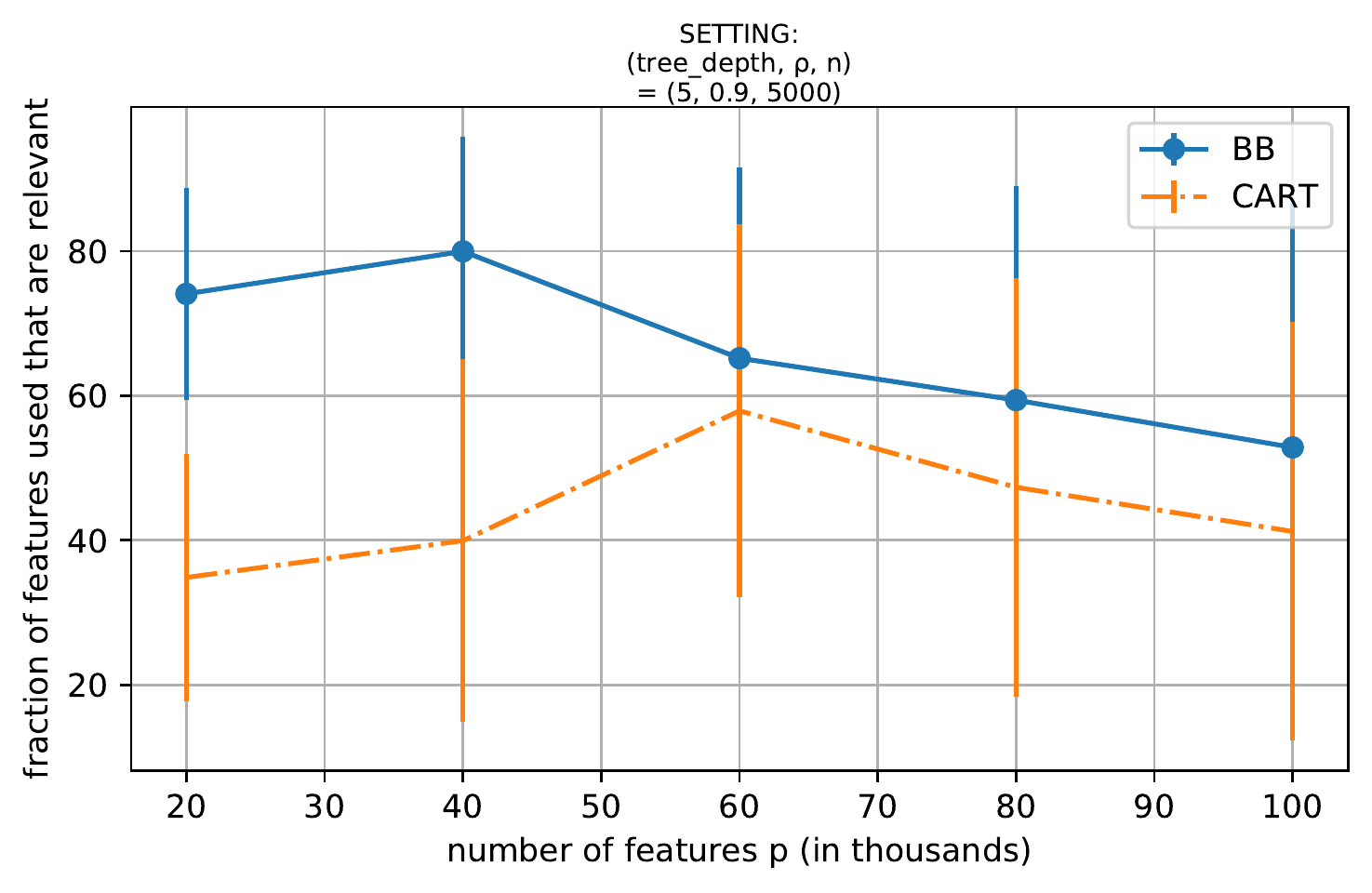}}
    \subfigure[Learned tree depth]{\includegraphics[width=0.49\textwidth]{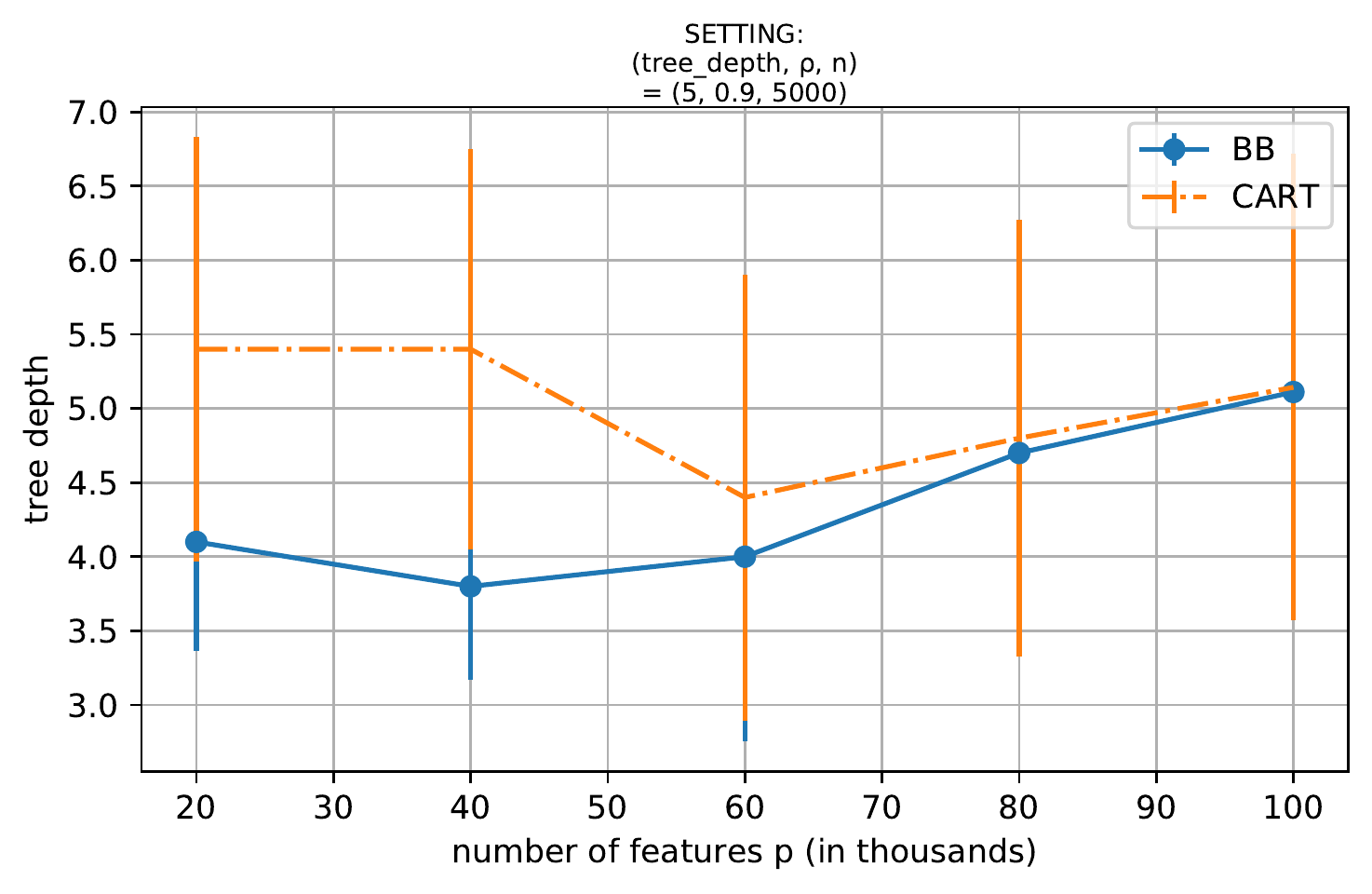}} 
    \subfigure[Out-of-sample AUC]{\includegraphics[width=0.49\textwidth]{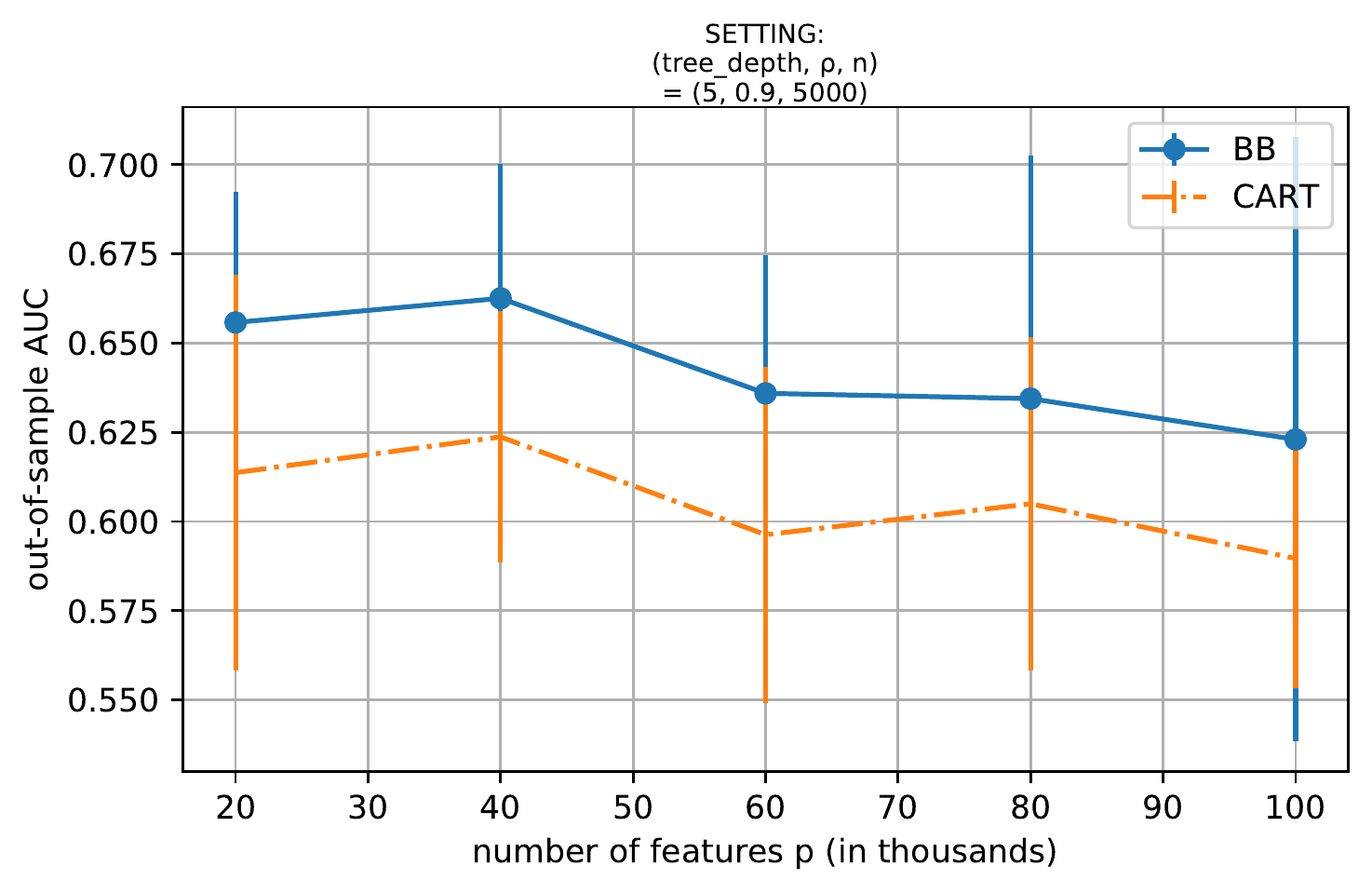}}
    \subfigure[Computational Time (sec)]{\includegraphics[width=0.49\textwidth]{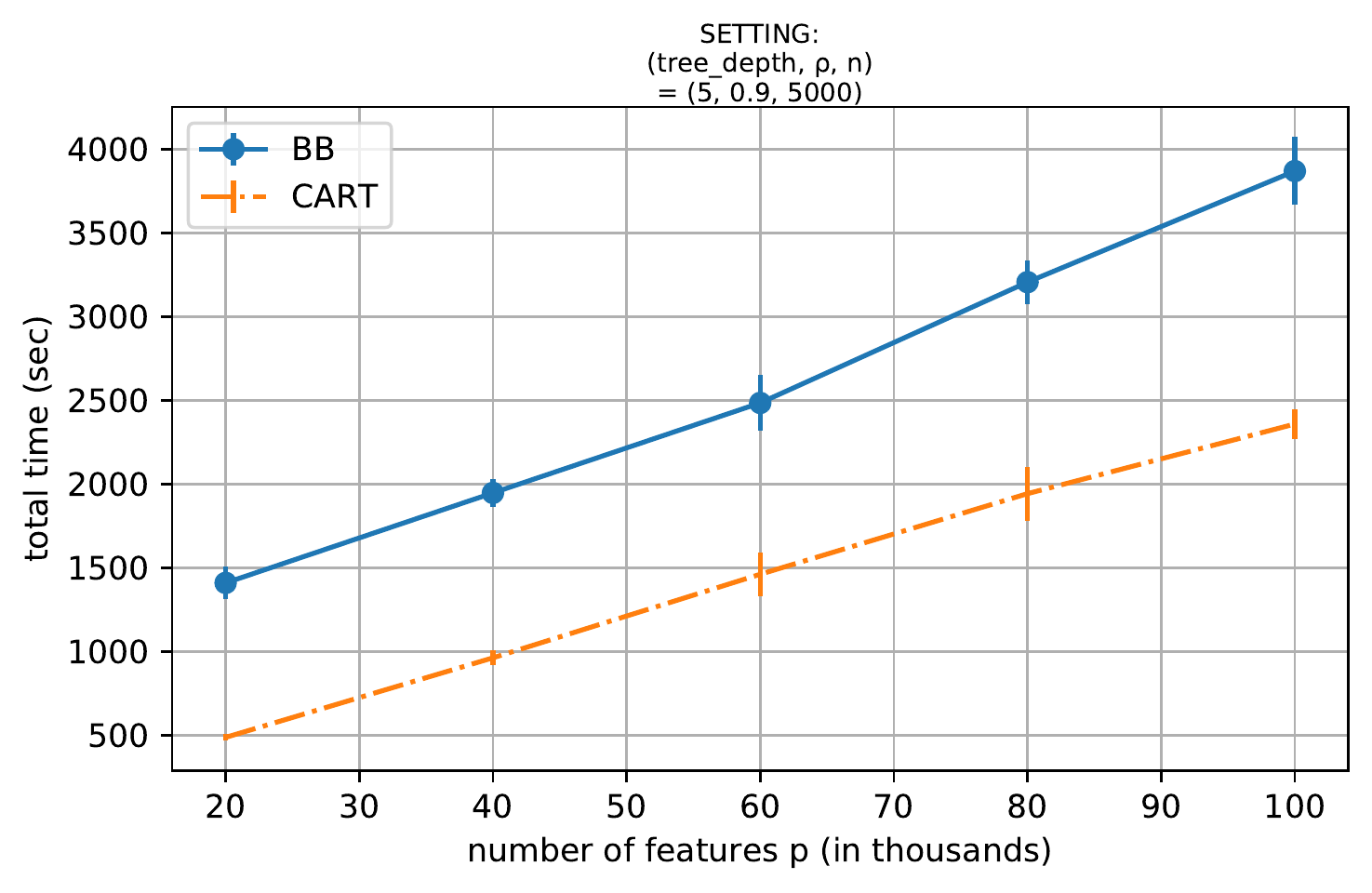}}
    \caption{Scalability with the number of features for classification trees.}
    \label{fig:trees_scalability}
\end{figure}

\subsection{Scalability with the Number of Samples} \label{subsec:results-synth-scalability-samples}

In this experiment, we examine the scalability of \verb|BB| as the number of samples $n$ increases. We show that \verb|BB| outperforms baseline heuristics in the context of sparse linear regression, sparse logistic regression, and classification trees.

\paragraph{Sparse Linear Regression.} We consider a sparse linear regression problem with $p=10^6$ features, $k=50$ relevant features, $\text{SNR}=2$, and correlation $ \rho = 0.9.$ We vary the number of data points $n \in \{1,000, 3,000, \dots, 9,000$\}.

We tune \verb|BB| and \verb|SIS-ENET| as described in the first experiment in Section \ref{subsec:results-synth-scalability-features}, under the following modifications. For \verb|BB|, we now select the \texttt{screen} function's parameter $\alpha$ such that all but $10,000$ features are eliminated, $\beta=0.5$ and solve $M=15$ subproblems.

Figure \ref{fig:regression_regimes} presents the results for this experiment. \verb|BB| outperforms \verb|SIS-ENET| as the number of samples increases, in terms of both support recovery accuracy and out-of-sample predictive performance, and solves problems with 1 million features and $9,000$ samples in less than an hour.

\begin{figure}[htbp] 
    \centering
    \subfigure[Support recovery accuracy]{\includegraphics[width=0.49\textwidth]{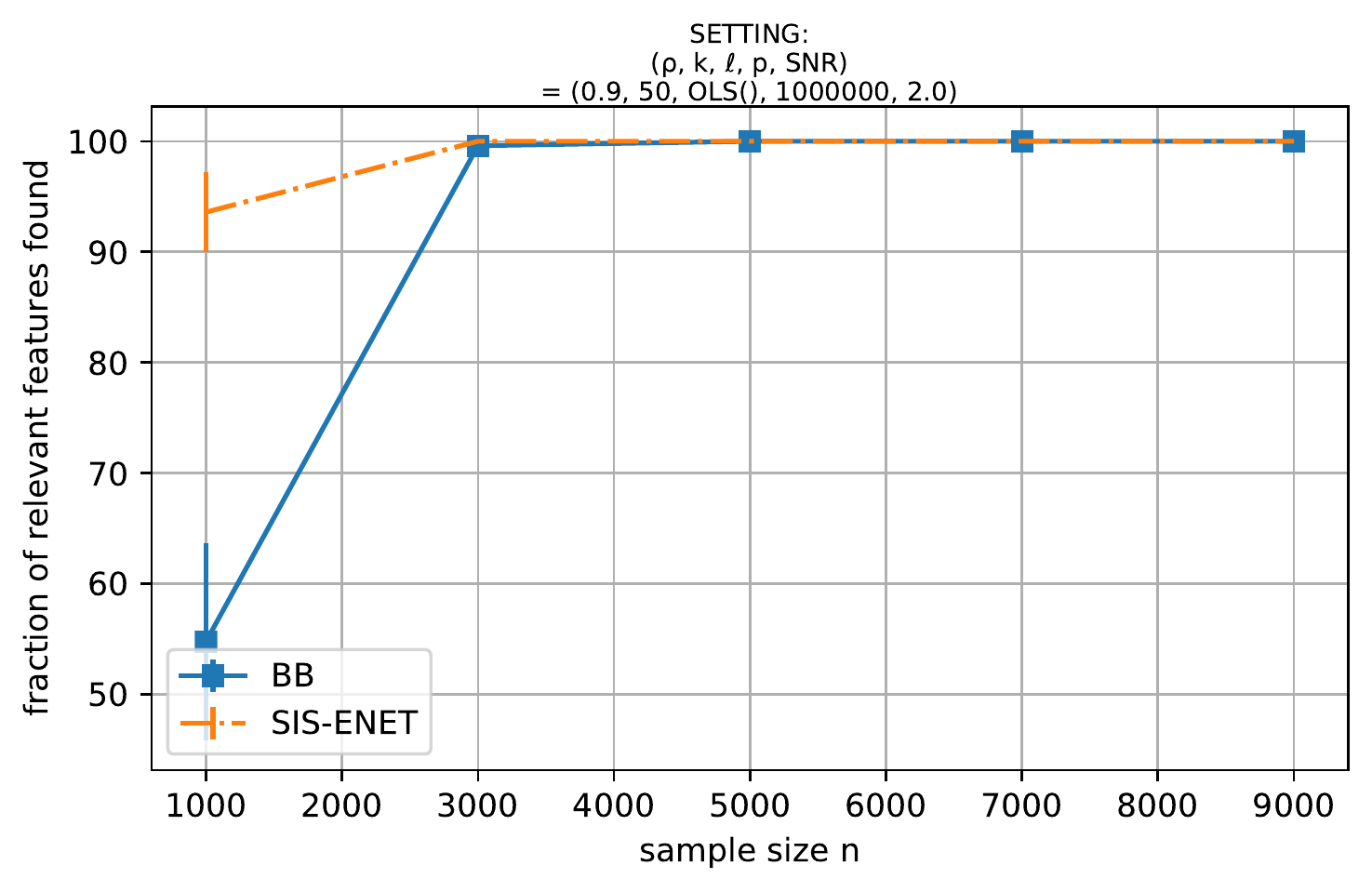}}
    \subfigure[Support recovery false alarm rate]{\includegraphics[width=0.49\textwidth]{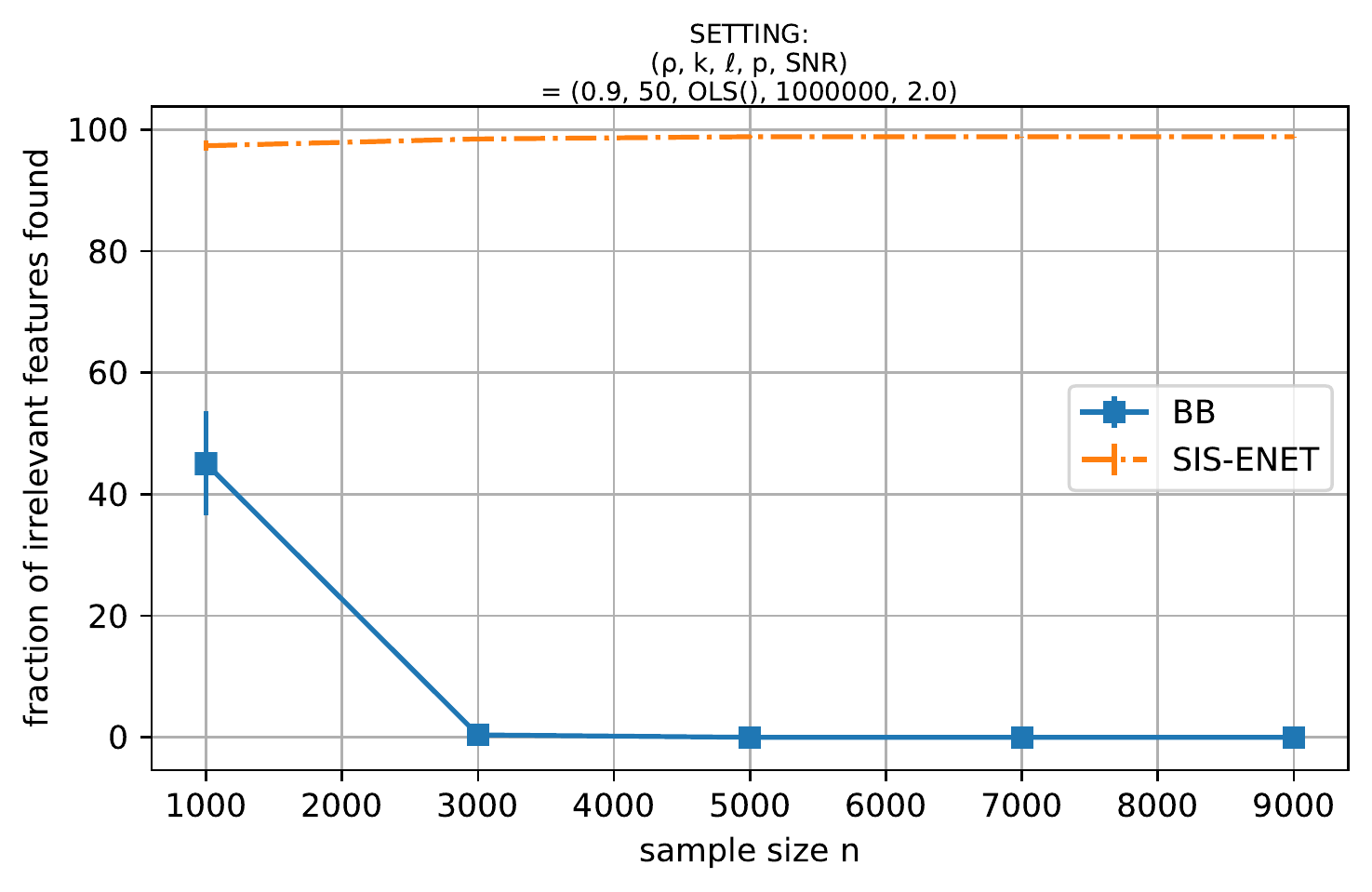}} 
    \subfigure[Out-of-sample R$^2$]{\includegraphics[width=0.49\textwidth]{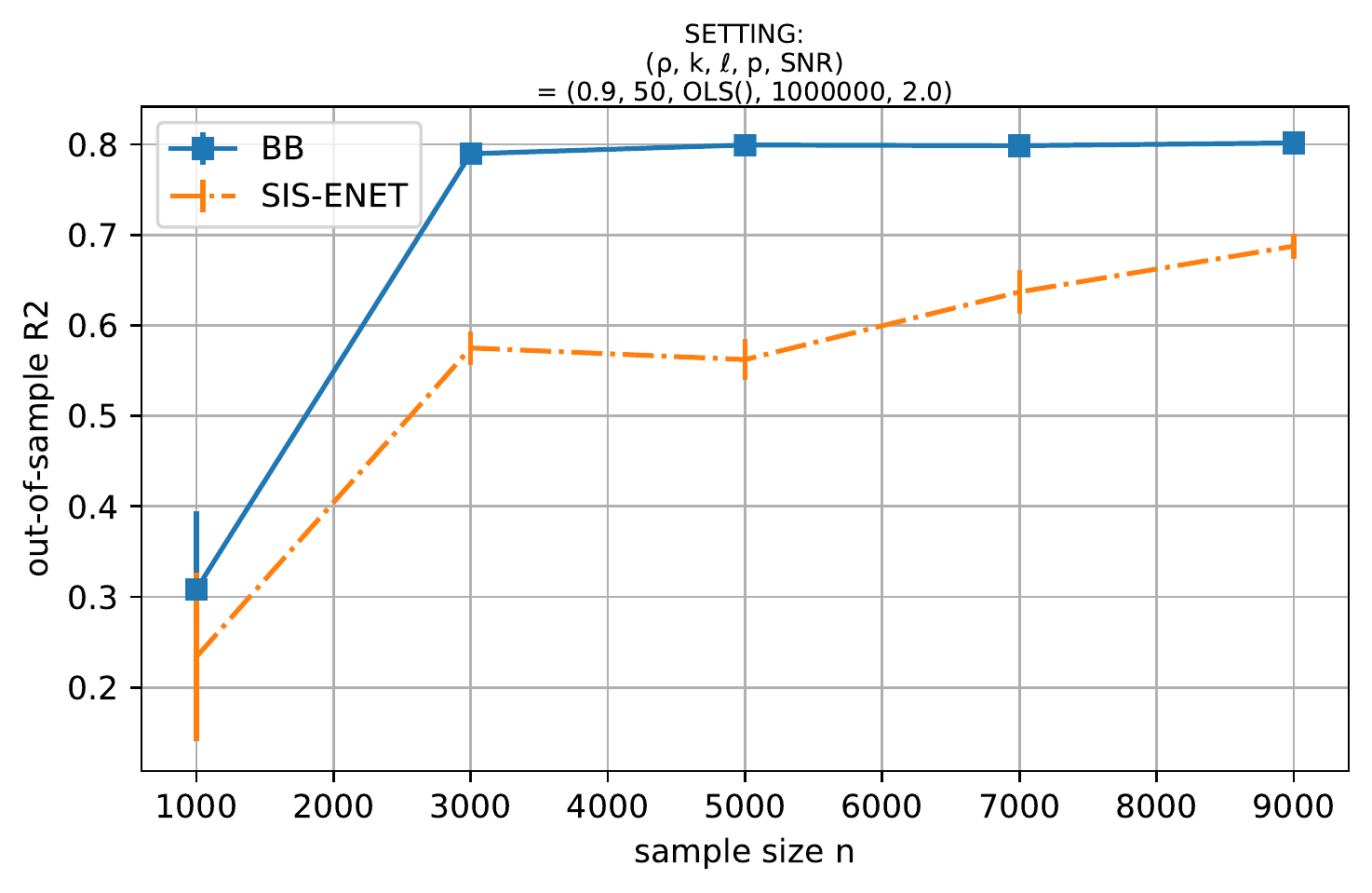}}
    \subfigure[Computational Time (sec)]{\includegraphics[width=0.49\textwidth]{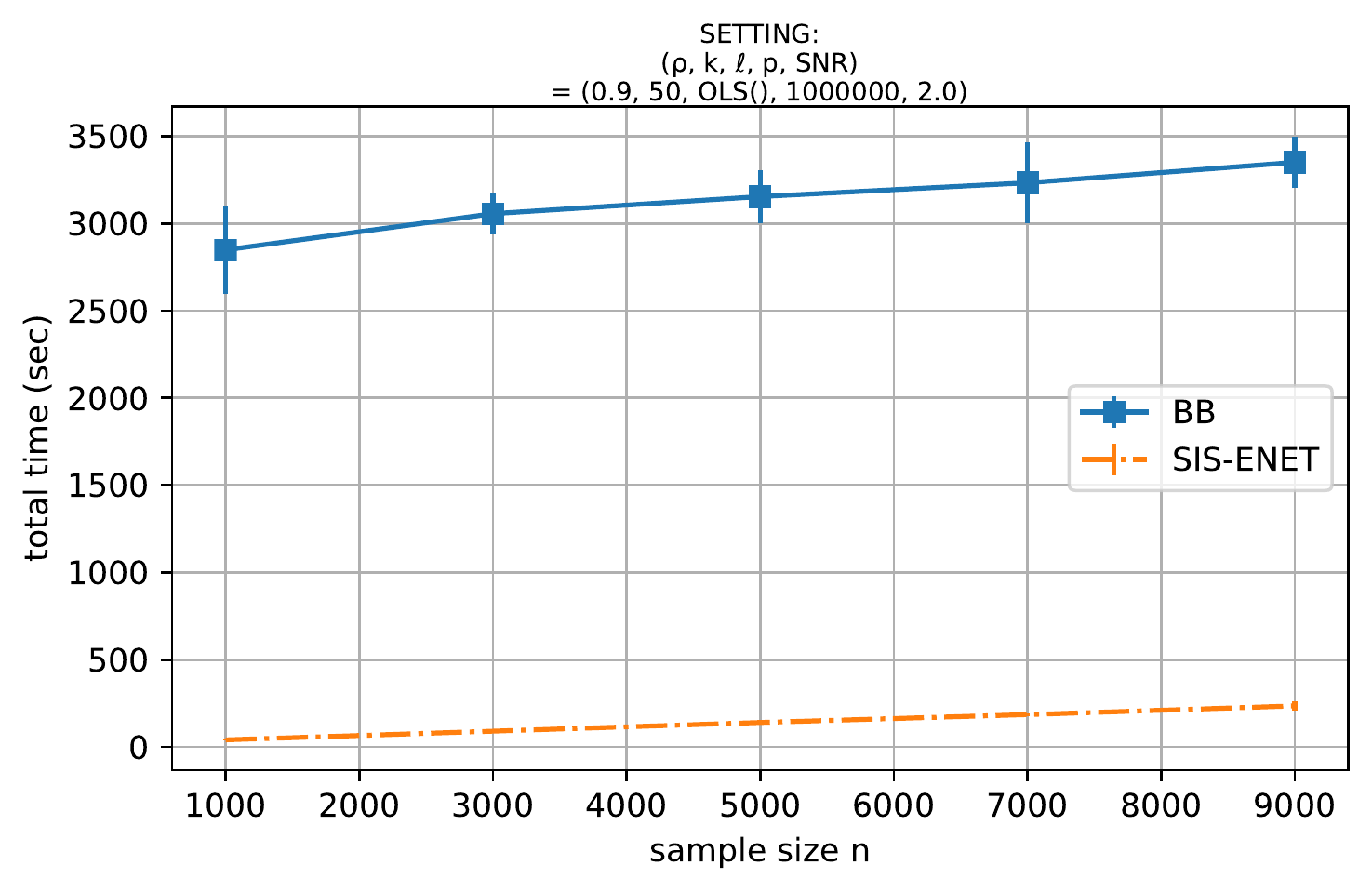}}
    \caption{Scalability with the number of samples for sparse linear regression.}
    \label{fig:regression_regimes}
\end{figure}

\paragraph{Sparse Logistic Regression.} We consider a sparse logistic regression problem with $p=10^6$ features, $k=50$ relevant features, $\text{SNR}=2$, and correlation $ \rho = 0.9.$ We vary the number of data points $n \in \{3,000, 5,000, \dots, 11,000$\}.

We tune \verb|BB| and \verb|SIS-ENET| as described in the sparse linear regression experiment preceding this one (Section \ref{subsec:results-synth-scalability-samples}), under the following modifications. For \verb|BB|, we now cross-validate the hyperparameter $\gamma_m$ within each subproblem and increase the time limit for the reduced problem to $15$ minutes.

In Figure \ref{fig:regression_logistic}, we show that \verb|BB| outperforms \verb|SIS-ENET| as the number of samples increases, in terms of both support recovery accuracy and out-of-sample predictive performance, and solves sparse logistic regression problems with 1 million features and $11,000$ samples in less than five hours. This experiment illustrates that \verb|BB| performs equally well in classification problems, whereby the logistic loss is used instead of the least squares loss.

\begin{figure}[htbp] 
    \centering
    \subfigure[Support recovery accuracy]{\includegraphics[width=0.49\textwidth]{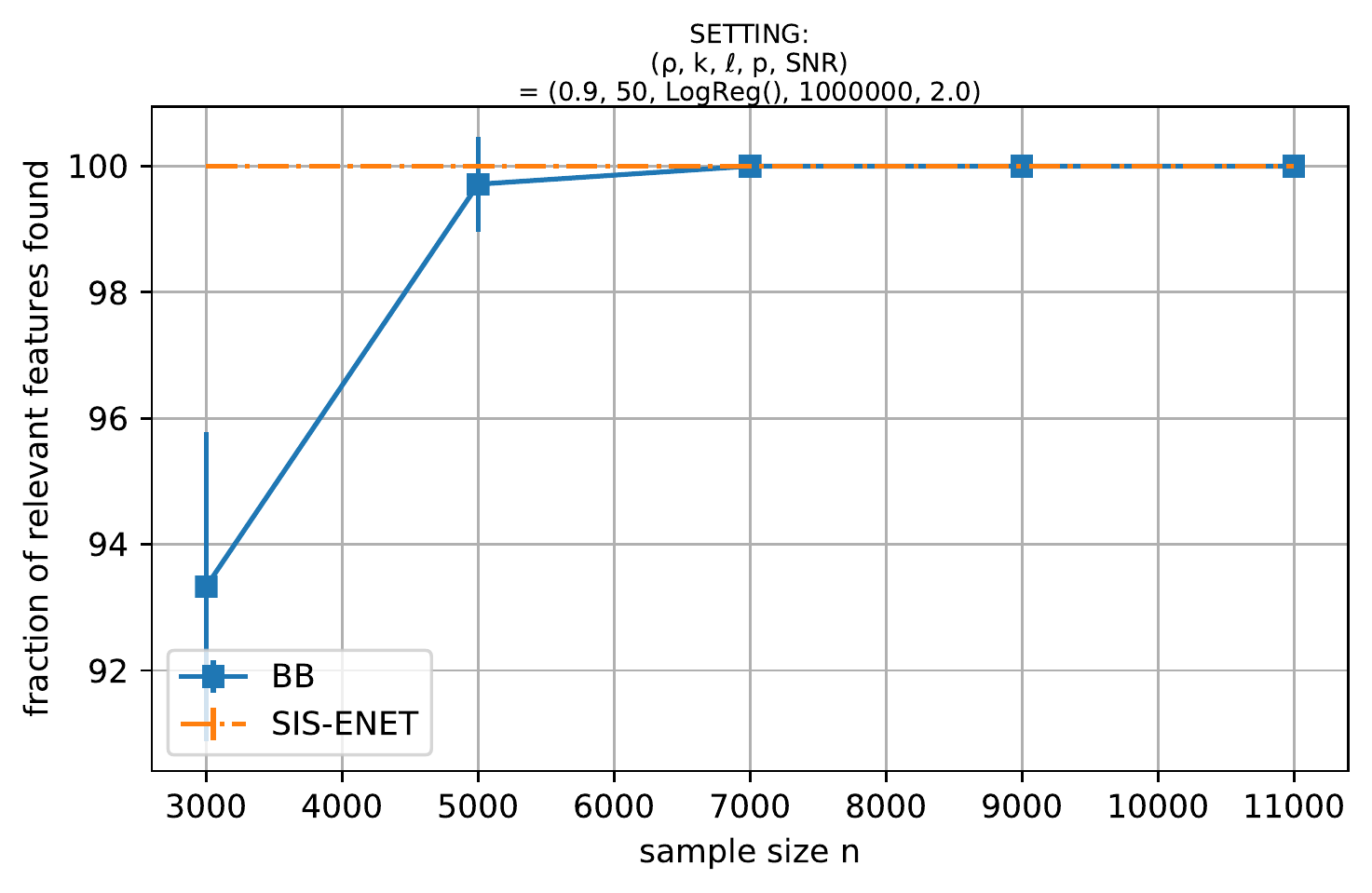}}
    \subfigure[Support recovery false alarm rate]{\includegraphics[width=0.49\textwidth]{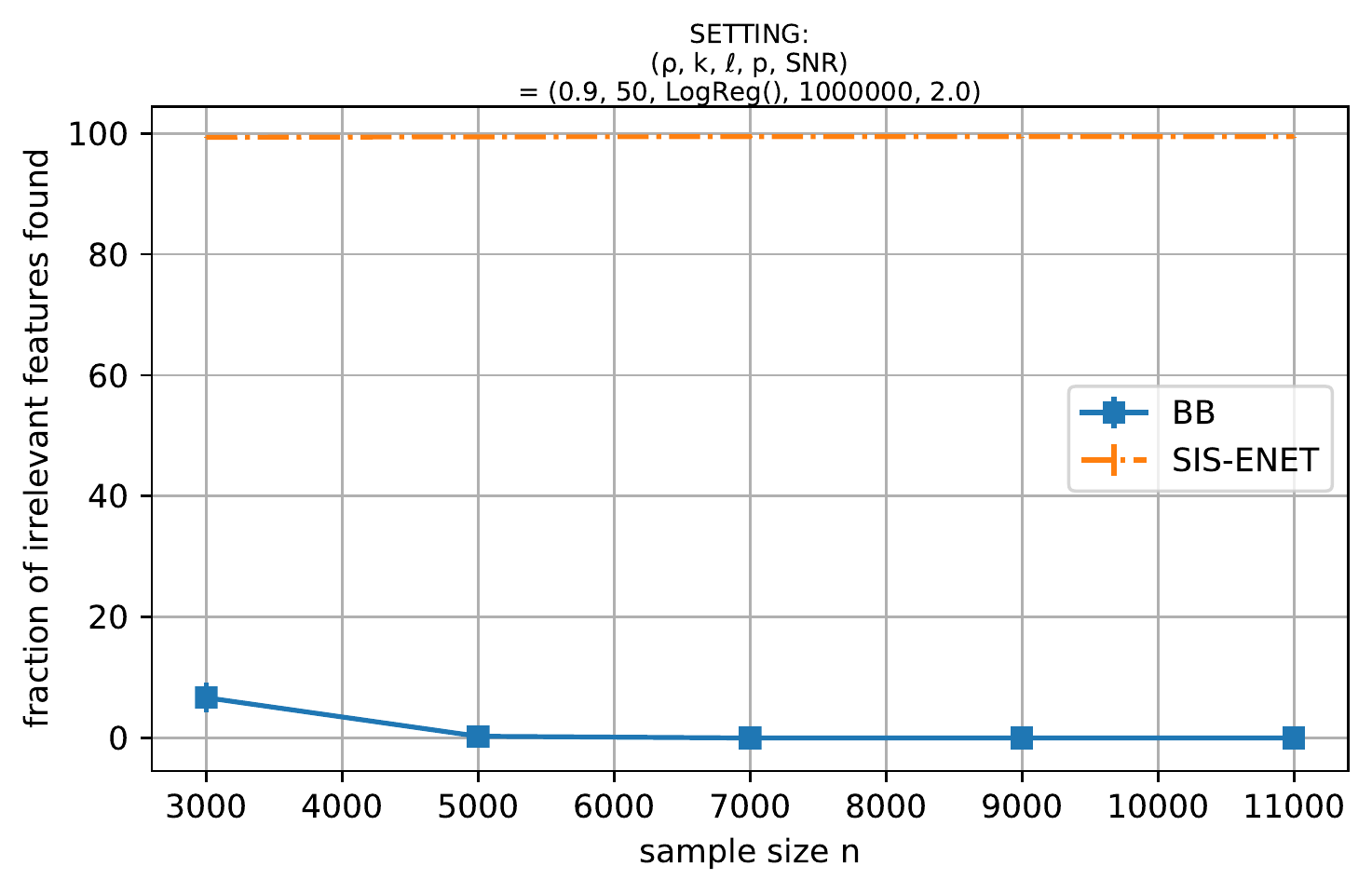}} 
    \subfigure[Out-of-sample AUC]{\includegraphics[width=0.49\textwidth]{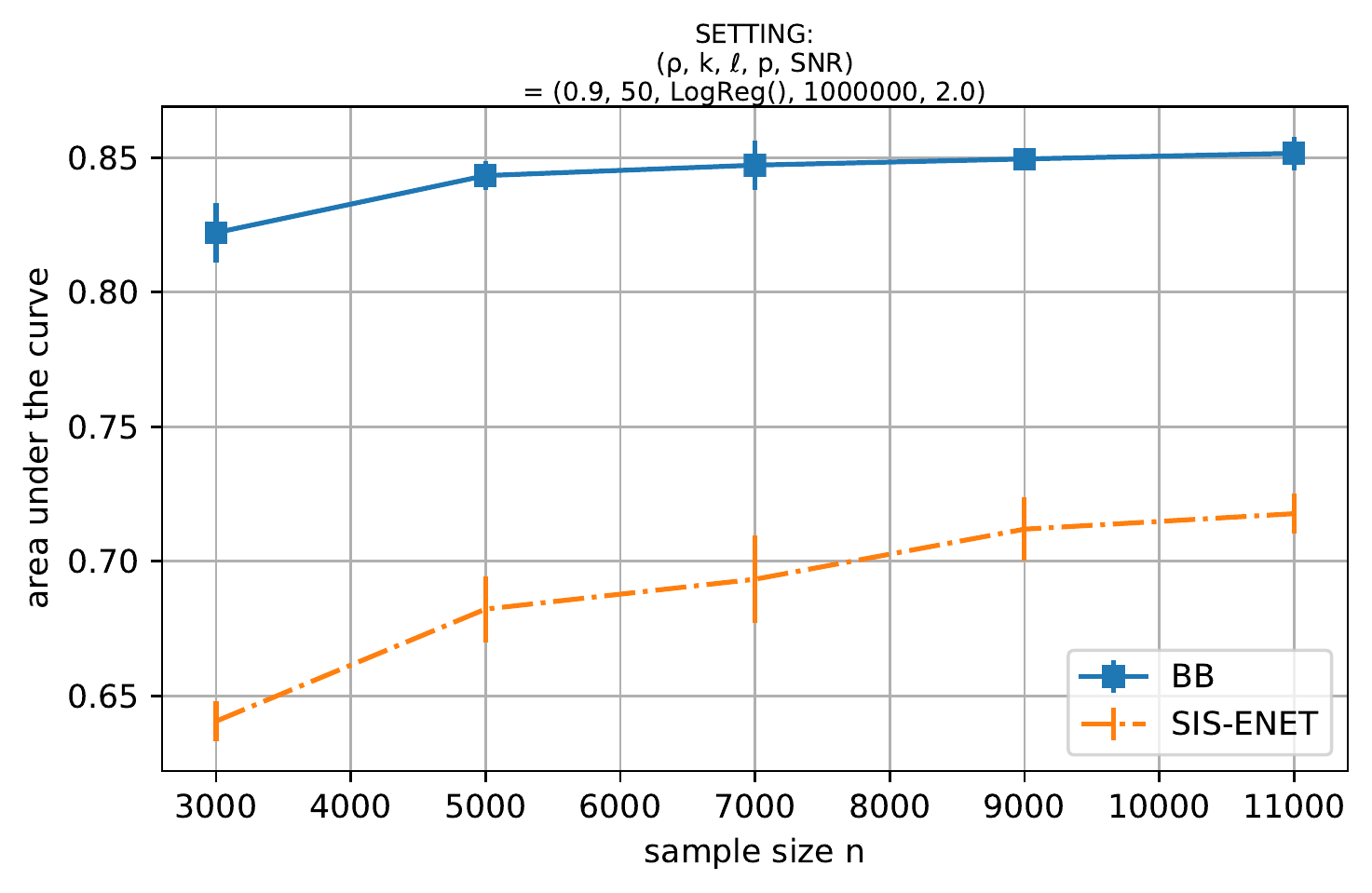}}
    \subfigure[Computational Time (sec)]{\includegraphics[width=0.49\textwidth]{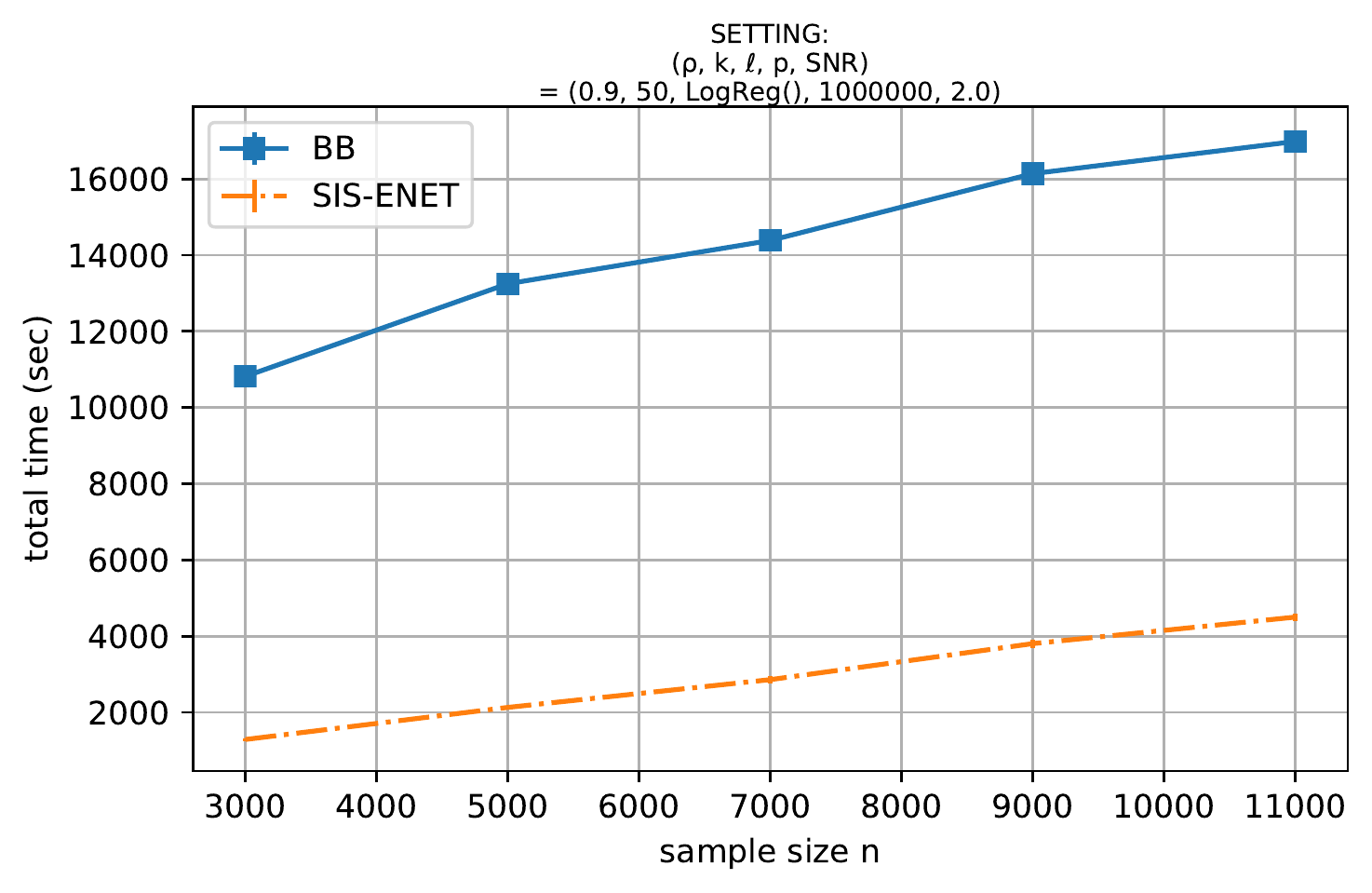}}
    \caption{Scalability with the number of samples for sparse logistic regression.}
    \label{fig:regression_logistic}
\end{figure}

\paragraph{Classification Trees.} We consider a classification tree problem with $p=100,000$ features, tree depth of $D=3$ and $k=7$ relevant features, and correlation $\rho = 0.6.$ By considering a much simpler ground truth tree than in Section \ref{subsec:results-synth-scalability-features}, the methods under investigation will hopefully be able to learn a tree that is closer to the truth. We vary the number of data points $n \in \{3,000,5,000,7,000\}$.

We tune \verb|BB| and \verb|CART| as described in the classification tree experiment in Section \ref{subsec:results-synth-scalability-features}, under the following modifications. For \verb|BB|, we now select $\beta=0.5$ and solve $M=10$ subproblems (i.e., we solve fewer, larger subproblems compared to Section \ref{subsec:results-synth-scalability-features}).

The results are presented in Figure \ref{fig:trees_regimes}. Both methods achieve near perfect out-of-sample AUC and \verb|BB| is computationally more intensive; this is likely due to the fact that \verb|OCT|, which is used to solve the reduced problem, is more sensitive to the number of samples in the data. Nevertheless, \verb|BB| results in trees that are much simpler and much closer to the ground truth, in that the fraction of features used that are relevant is close to $100\%$ and the learned tree's depth is, on average, within 1 of the ground truth tree's depth.

\begin{figure}[htbp] 
    \centering
    \subfigure[Fraction of features used that are relevant]{\includegraphics[width=0.49\textwidth]{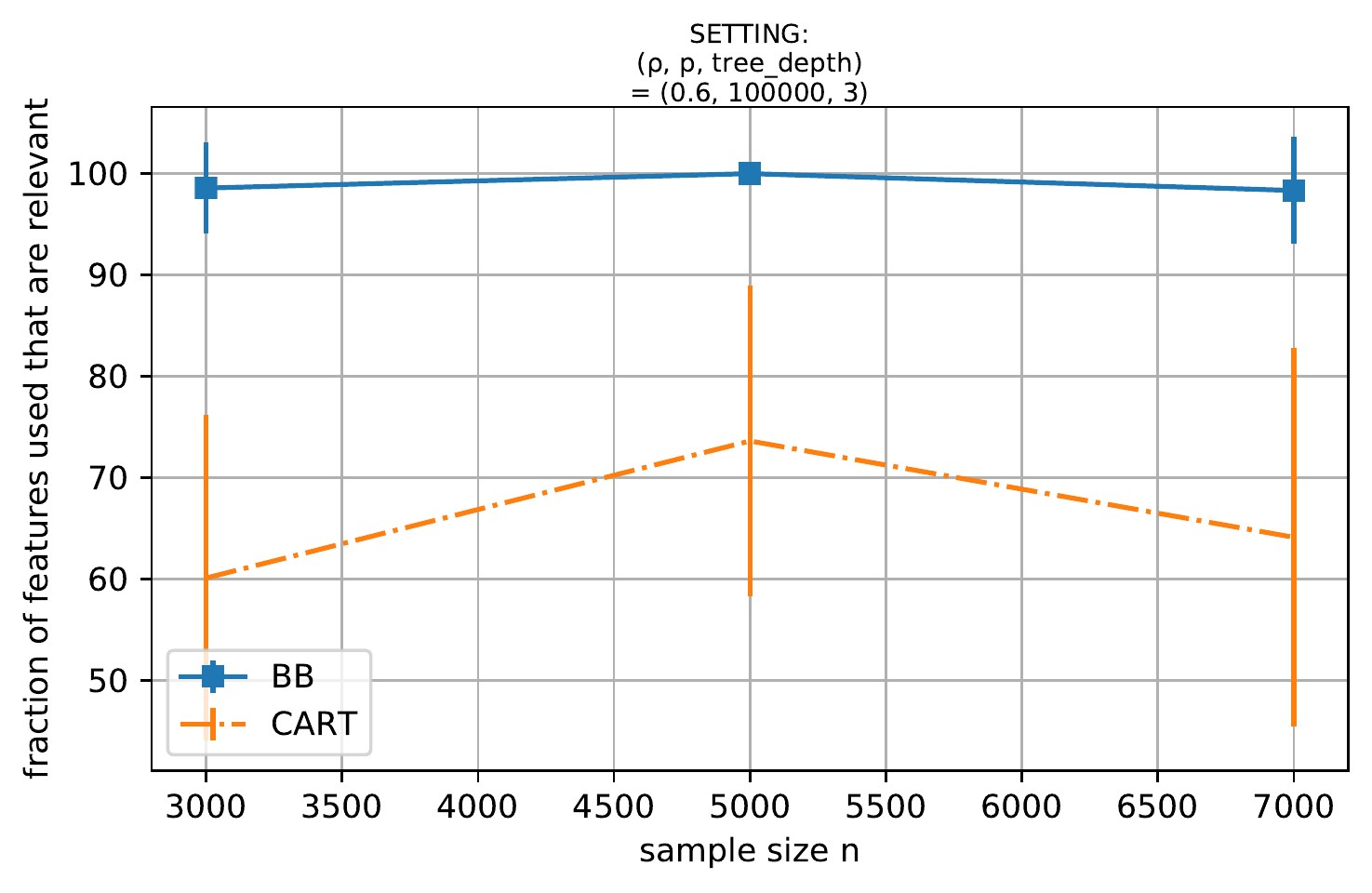}}
    \subfigure[Learned tree depth]{\includegraphics[width=0.49\textwidth]{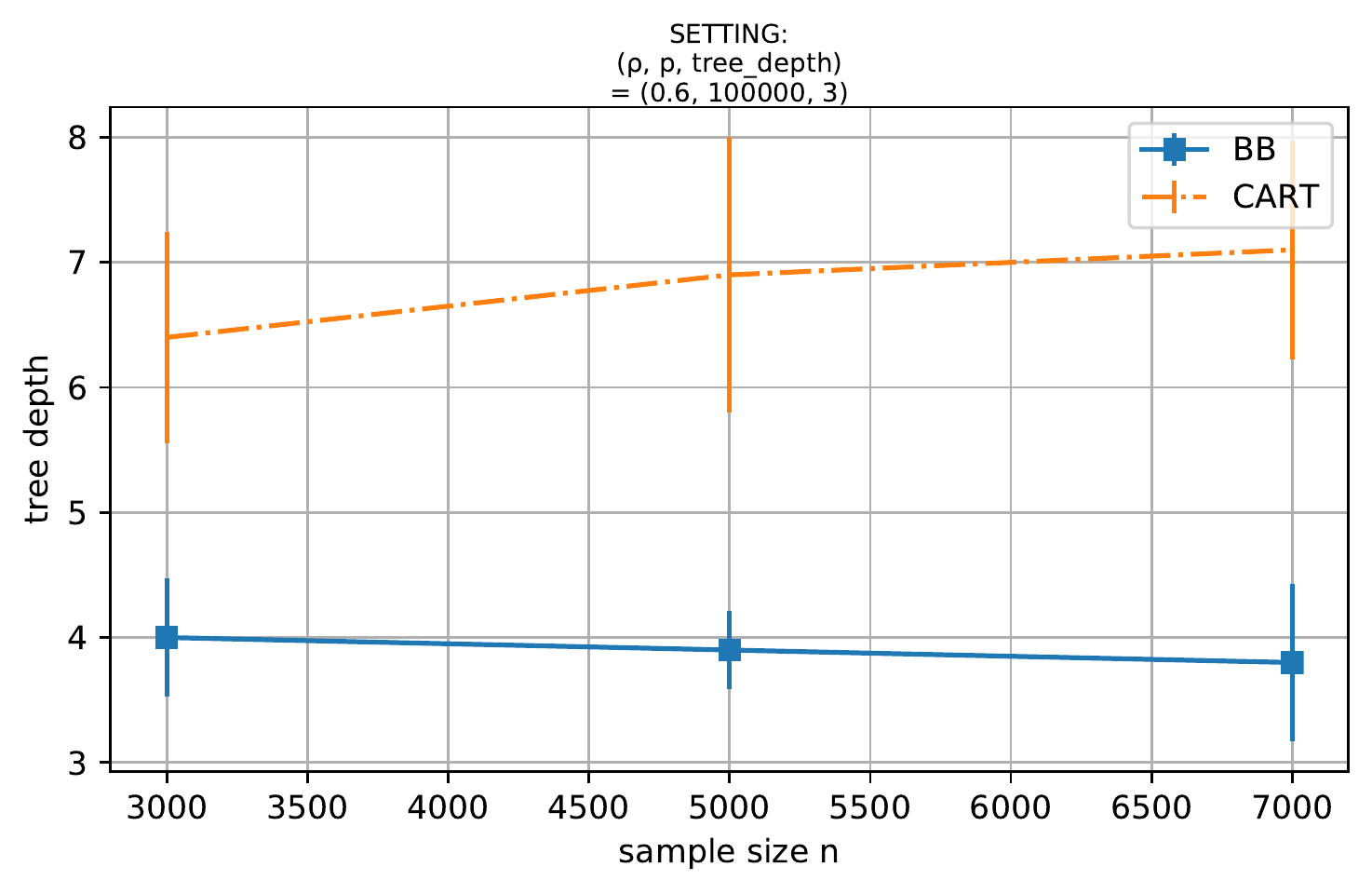}} 
    \subfigure[Out-of-sample AUC]{\includegraphics[width=0.49\textwidth]{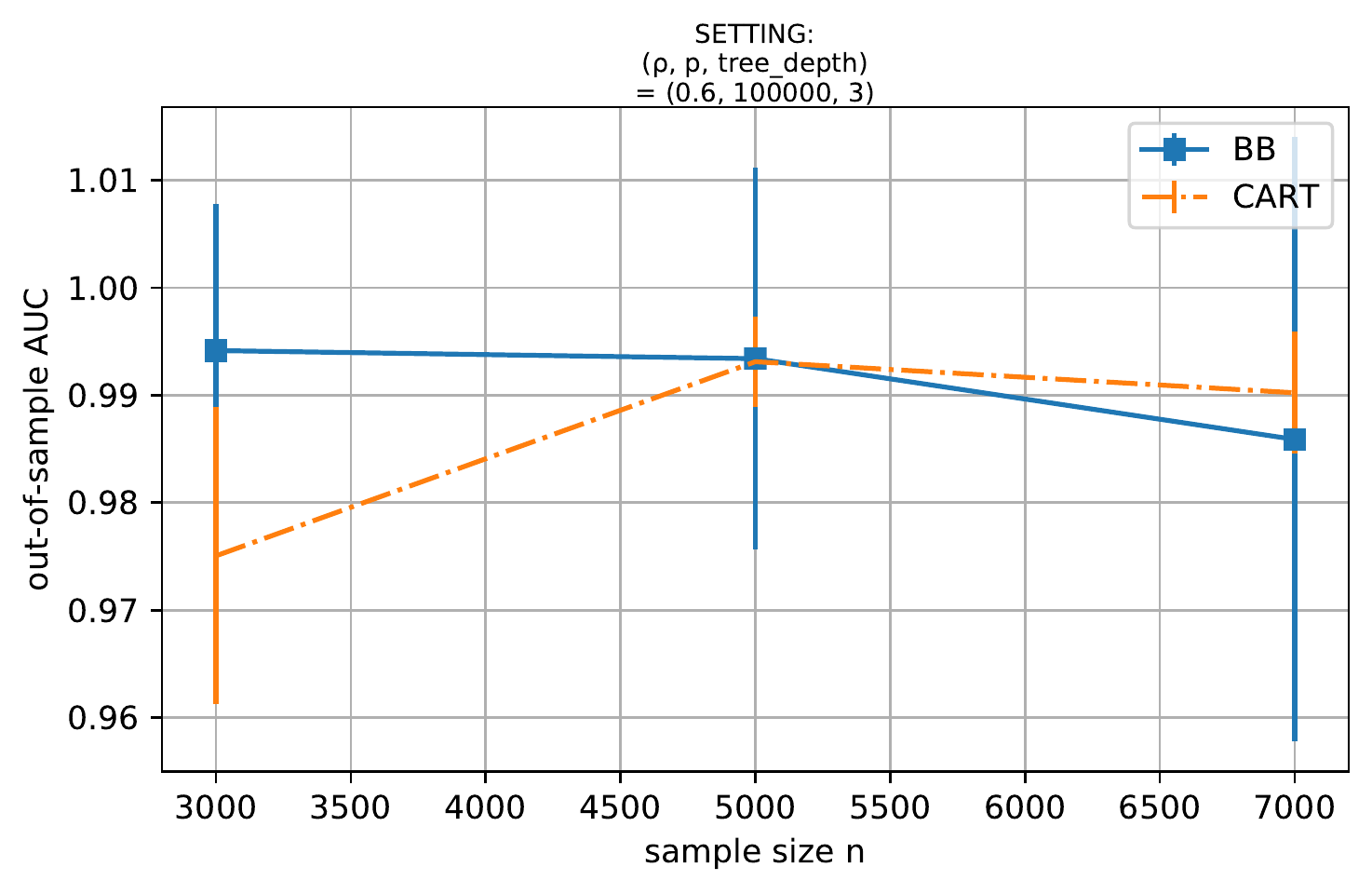}}
    \subfigure[Computational Time (sec)]{\includegraphics[width=0.49\textwidth]{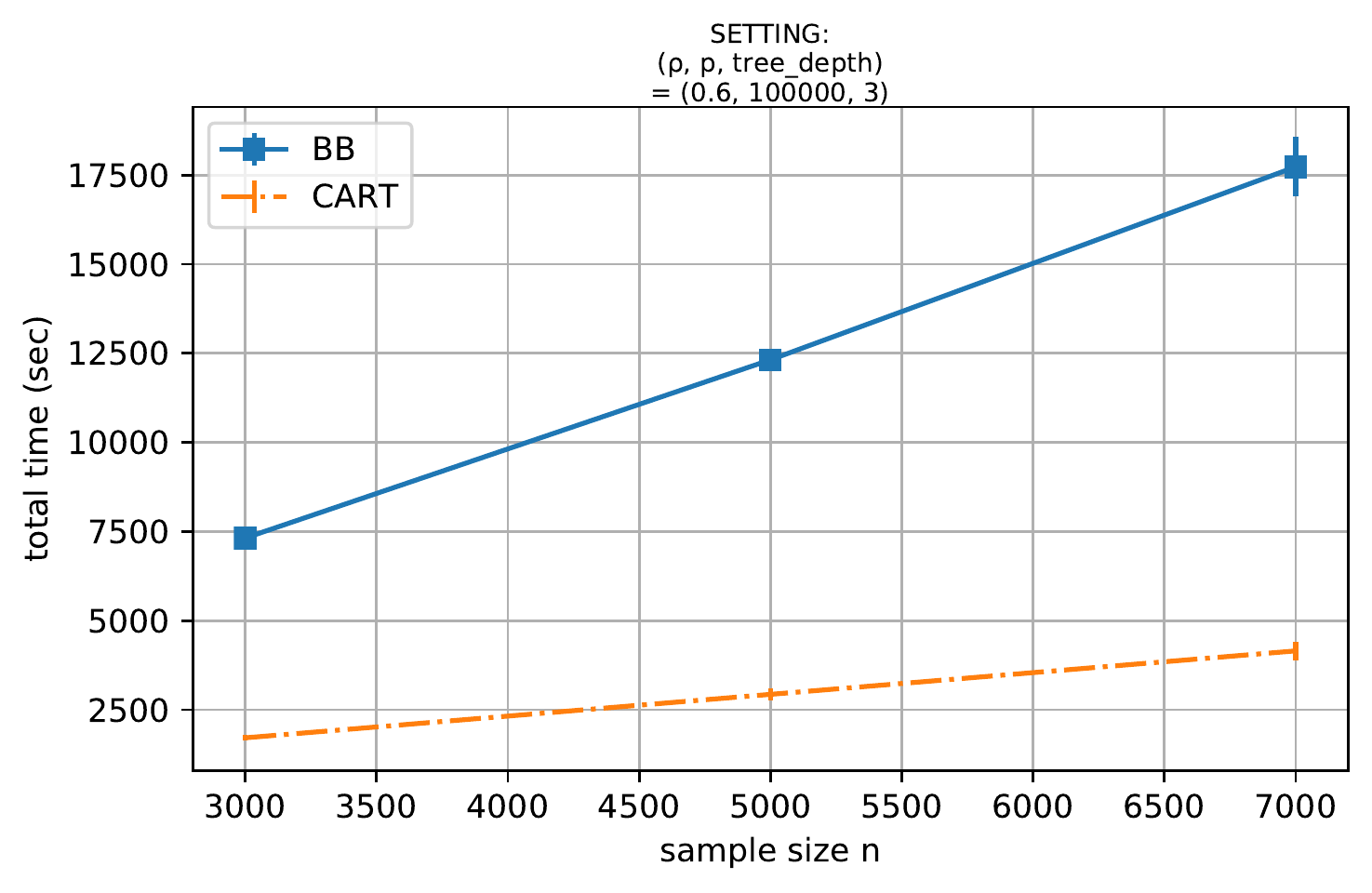}}
    \caption{Scalability with the number of samples for classification trees.}
    \label{fig:trees_regimes}
\end{figure}

\subsection{Comparison with Exact Methods Applied to the Entire Feature Set}\label{subsec:results-synth-exact}

In this experiment, we compare \verb|BB| with \verb|SR| or \verb|OCT| applied to the entire feature set in problems that are sufficiently small and \verb|SR| or \verb|OCT| scale. We show that, in such regimes, \verb|BB| competes with optimal or near-optimal solutions, while substantially reducing the computational time and/or MIO optimality gap.

\paragraph{Sparse Linear Regression.} We consider a sparse linear regression problem with $p=20,000$ features, $k=50$ relevant features, $\text{SNR}=2$, and correlation $ \rho = 0.9.$ We vary the number of data points $n \in \{1,000, 2,000, \dots, 5,000$\}.

We tune \verb|BB| as described in the first experiment in Section \ref{subsec:results-synth-scalability-features}, under the following modifications. For \verb|BB|, we now select the \texttt{screen} function's parameter $\alpha$ such that all but $5,000$ features are eliminated, we set $\beta=0.4$, and solve $M=10$ subproblems. We compare \verb|BB| with \verb|SR| applied to the entire feature set and tuned exactly as when solving the reduced problem in \verb|BB|.

Figure \ref{fig:regression_optimal} reports the support recovery accuracy, MIO optimality gap, and computational time of each method as function of the sample size $n$. 
Both methods achieve near perfect support recovery accuracy at similar rates; however, \verb|BB| seems to have an edge when it comes to support recovery false alarm rate. One possible explanation for this is that the first phase in \verb|BB| eliminates irrelevant features that \verb|SR| ends up selecting. Additionally, while the solution returned by \verb|SR| comes with no optimality guarantee, the optimality gap for \verb|BB| quickly drops to 0 as the number of samples increases, albeit for the reduced problem. In terms of computational time, \verb|BB| indeed performs effective feature selection and enables us to solve the reduced problem's sparse regression MIO formulation to near-optimality faster.

\begin{figure}[htbp] 
    \centering
    \subfigure[Support recovery accuracy]{\includegraphics[width=0.49\textwidth]{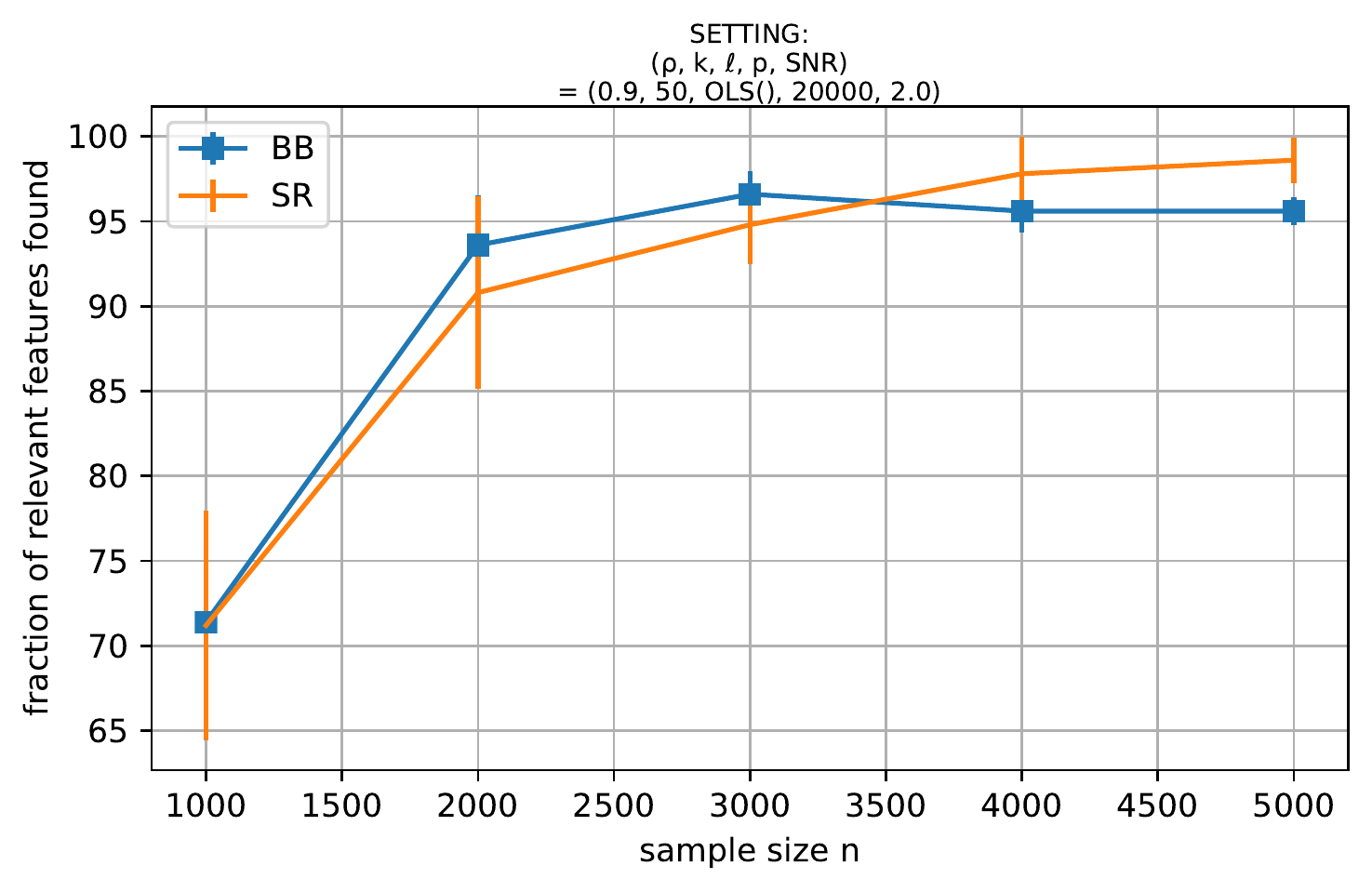}}
    \subfigure[Support recovery false alarm rate]{\includegraphics[width=0.49\textwidth]{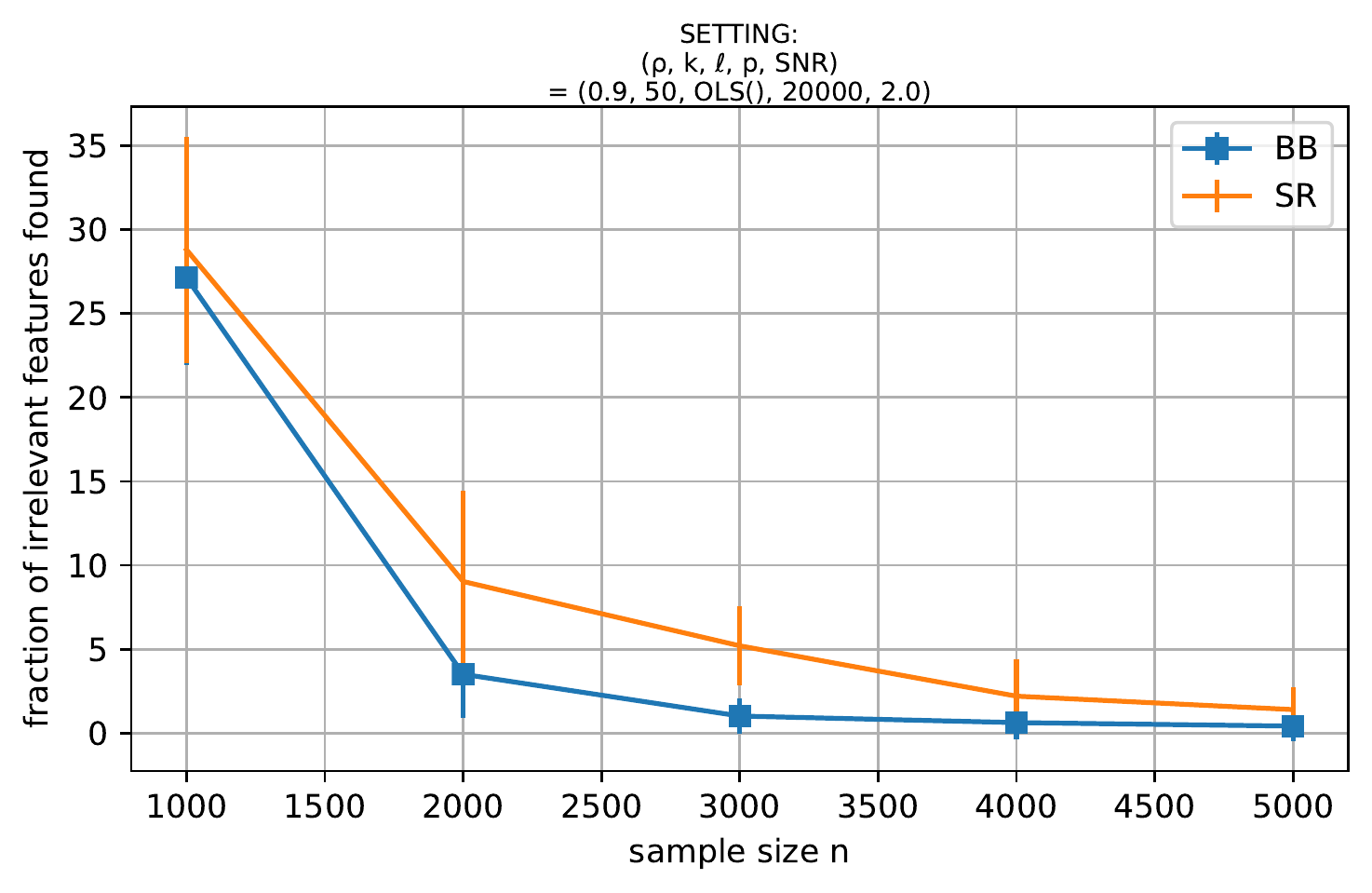}} 
    \subfigure[Optimality Gap]{\includegraphics[width=0.49\textwidth]{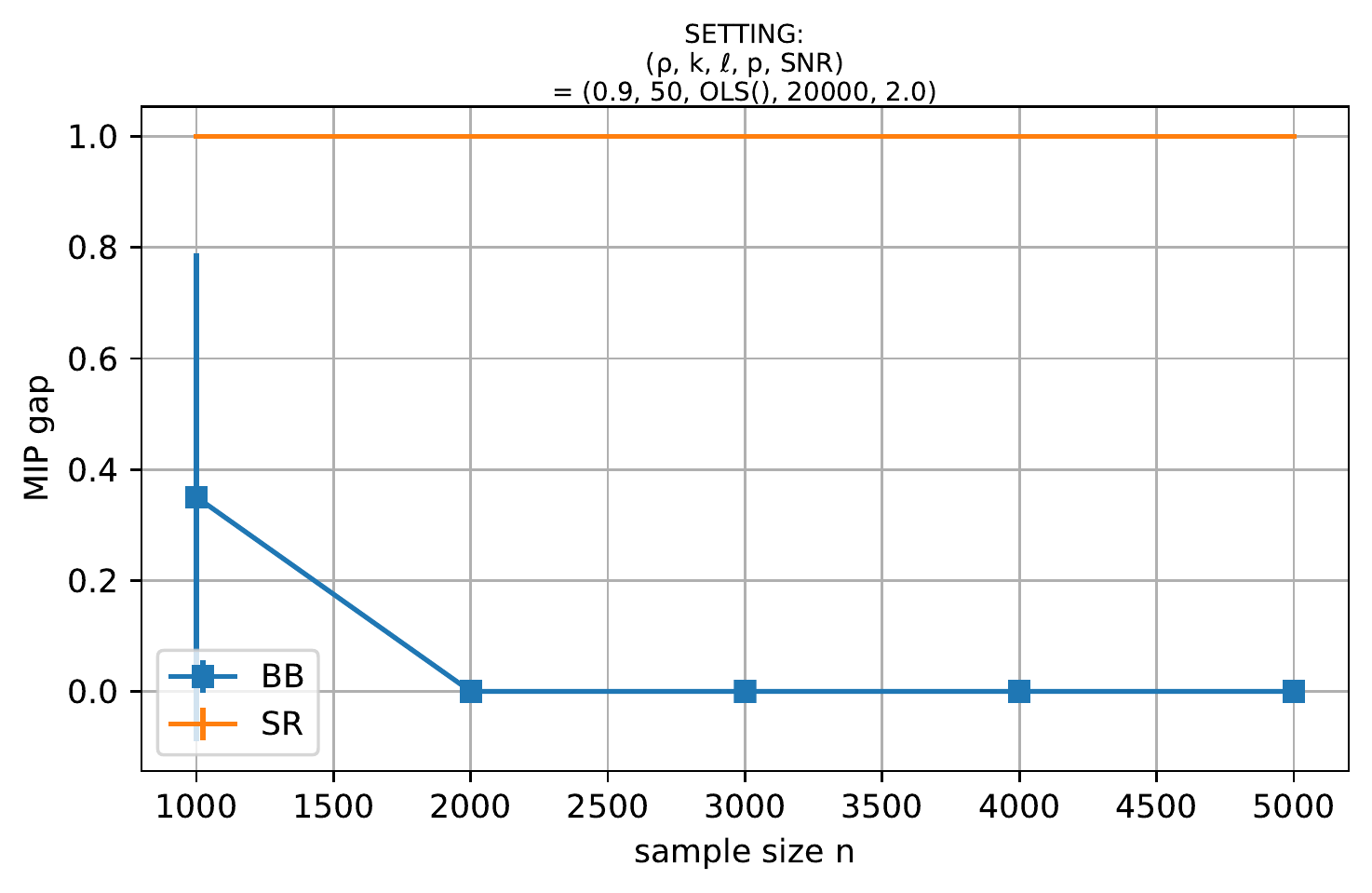}}
    \subfigure[Computational Time (sec)]{\includegraphics[width=0.49\textwidth]{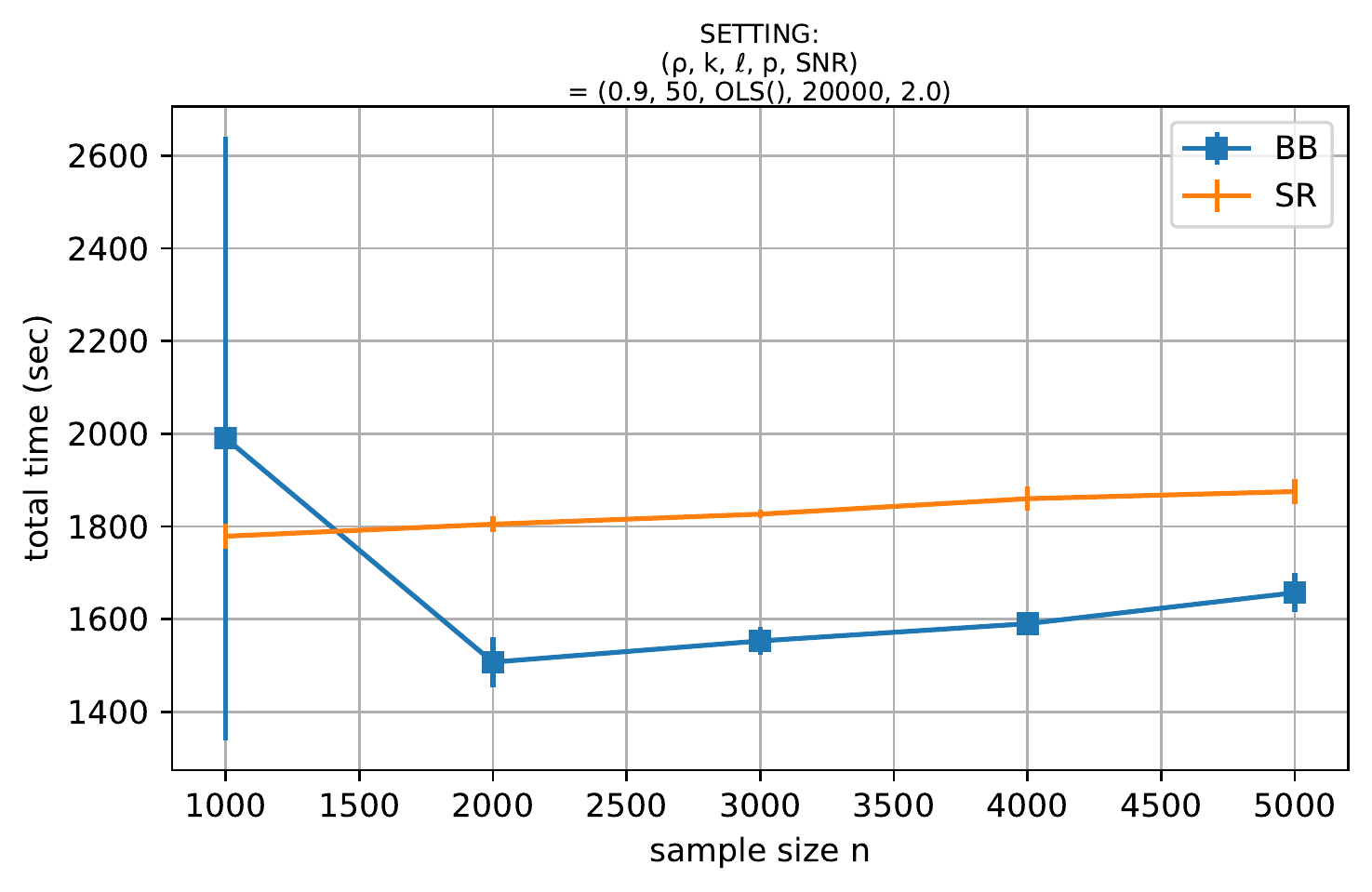}}
    \caption{Comparison with sparse linear regression applied to the entire feature set.}
    \label{fig:regression_optimal}
\end{figure}

\paragraph{Classification Trees.} We consider a classification tree problem with $p=2,000$ features, tree depth of $D=5$ and $k=31$ relevant features, and correlation $\rho = 0.9.$ We vary the number of data points $n \in \{250,500,1,000,1,500,2,000\}$.

We tune \verb|BB| as described in the classification tree experiment in Section \ref{subsec:results-synth-scalability-features}, under the following modifications. Since our focus now is solely on feature selection, we incorporate the screening step into \verb|BB|. Thus, we select $\alpha=0.5$, we screen features and construct subproblems using the logistic loss, we set $\beta=0.5$, and solve $M=10$ subproblems.

We present the results in Figure \ref{fig:trees_optimal}. In terms of support recovery and structure of the learned tree, the two methods performs similarly. Although \verb|OCT| has a slight edge in terms of predictive power, the gains of \verb|BB| in terms of computational time are tremendous.

\begin{figure}[htbp] 
    \centering
    \subfigure[Fraction of features used that are relevant]{\includegraphics[width=0.49\textwidth]{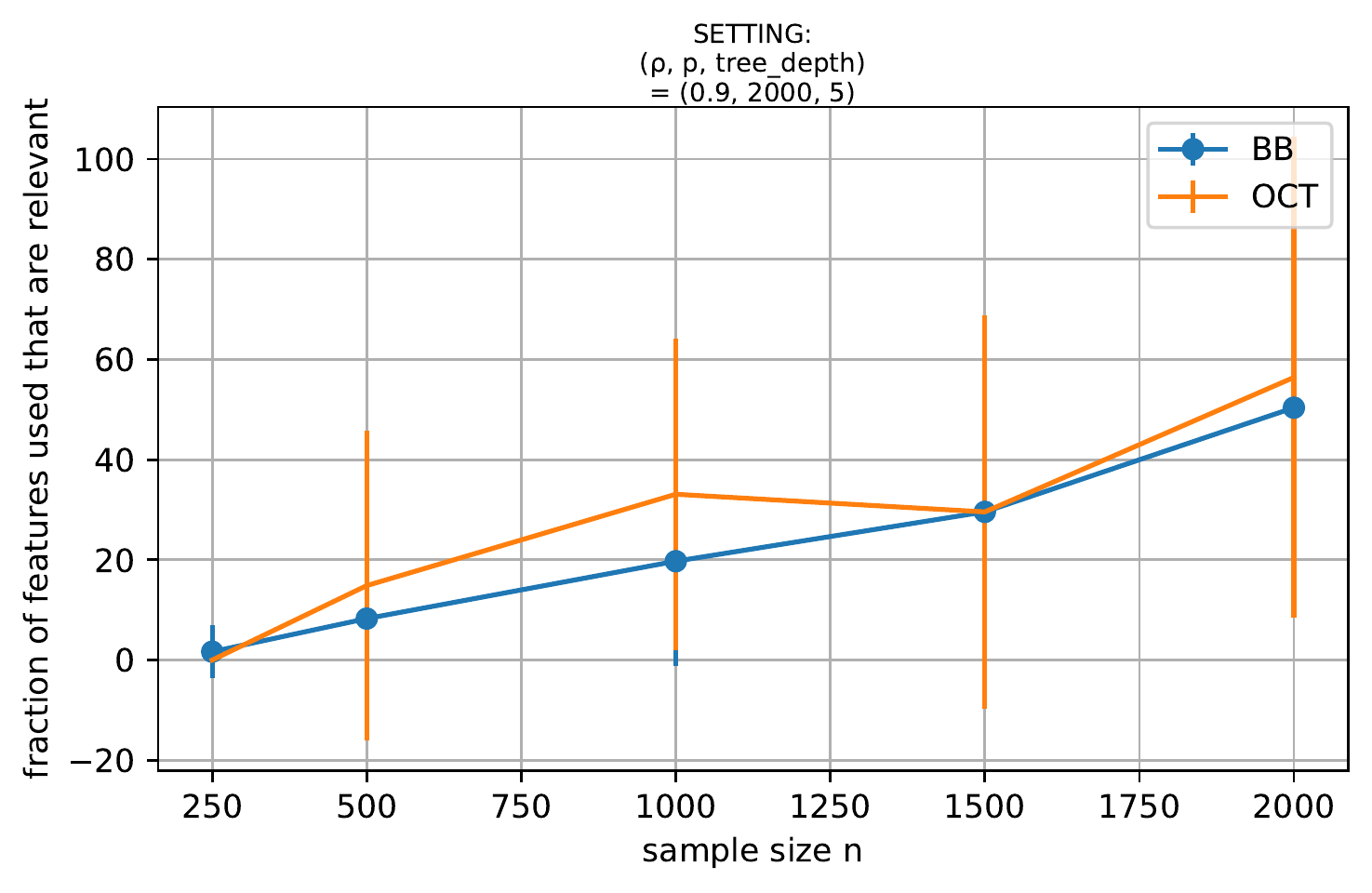}}
    \subfigure[Learned tree depth]{\includegraphics[width=0.49\textwidth]{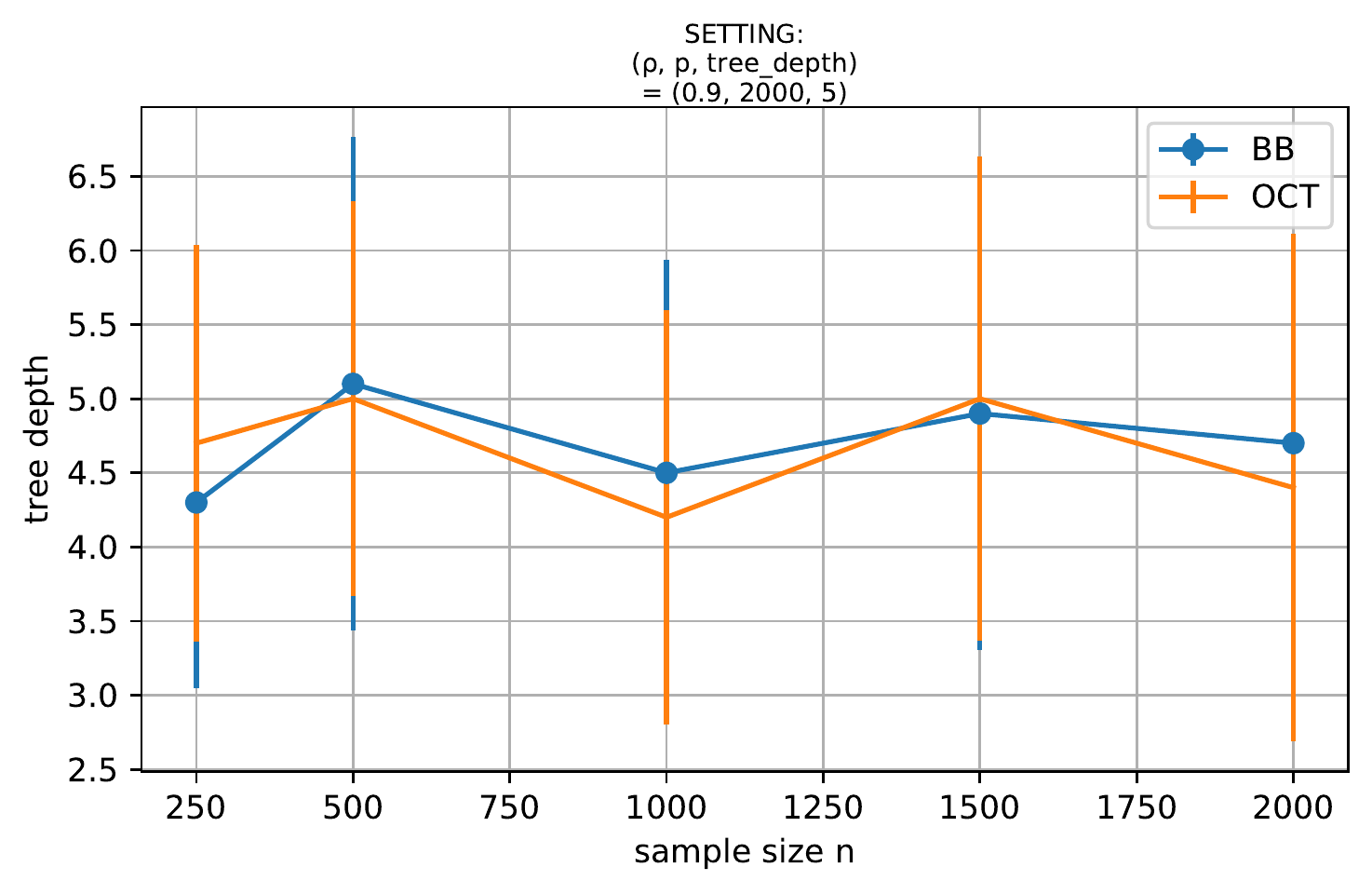}} 
    \subfigure[Out-of-sample AUC]{\includegraphics[width=0.49\textwidth]{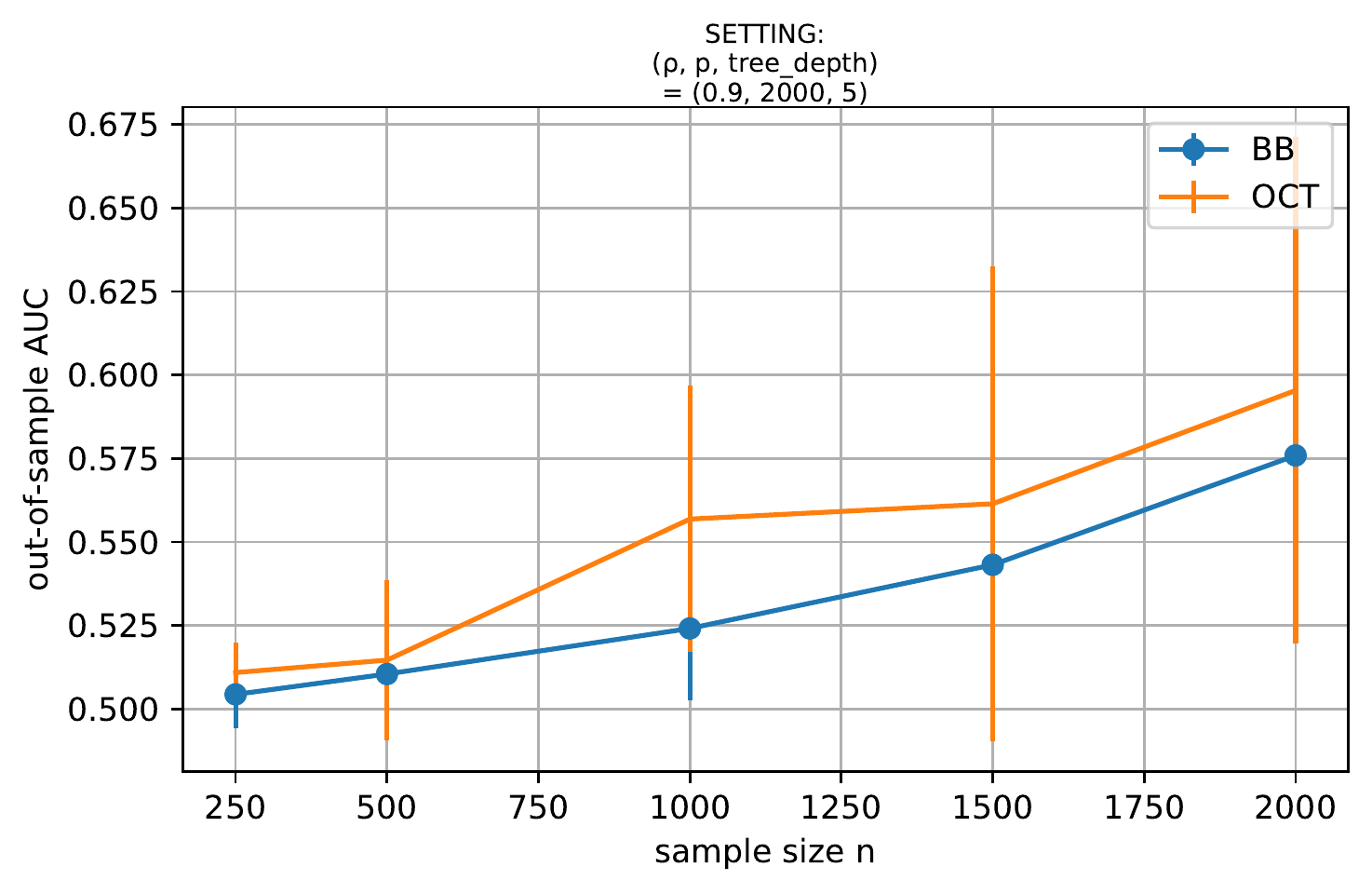}}
    \subfigure[Computational Time (sec)]{\includegraphics[width=0.49\textwidth]{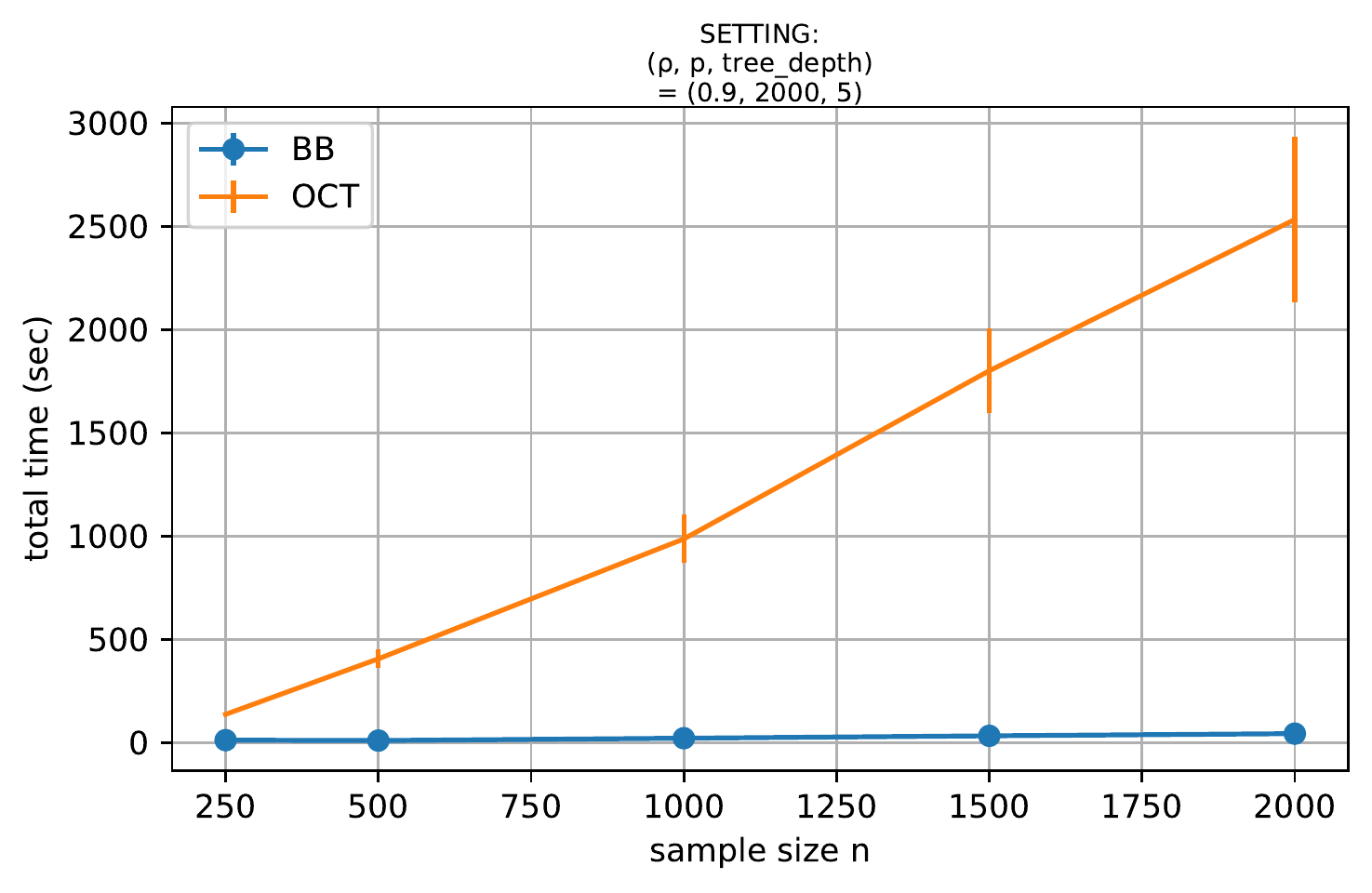}}
    \caption{Comparison with optimal classification trees applied to the entire feature set.}
    \label{fig:trees_optimal}
\end{figure}

\subsection{Sensitivity to the Backbone Method's Hyperparameters and Components} \label{subsec:results-synth-sensitivity}

In the remainder of this section, we present a detailed analysis on the sensitivity of \verb|BB| to its components and hyperparameters in the context of regression. The sensitivity analysis for the decision tree problem leads to near-identical conclusions, so we do not include it in the paper.

We consider a sparse linear regression problem with $n=5,000$ data points, $p=10^6$ features, $k=50$ relevant features, $\text{SNR}=2$, and correlation $ \rho = 0.9.$ Unless stated otherwise, we set the parameters of \verb|BB| as follows: we use the \verb|screen| function described in Algorithm \ref{alg:screen} and set $\alpha = 0.01$ so that all but $10^4$ features are eliminated; we construct subproblems using the \verb|construct_subproblems| function (Algorithm \ref{alg:construct_subproblems}) with $M=10$ and $\beta = 0.2$; we impose a maximum allowable backbone size of $B_{\max} = 1,000$ features; we solve the subproblems using \verb|SR-REL| and the reduced problem using \verb|SR| with a time limit of 5 minutes.

\paragraph{Number of Subproblems.} In this experiment, we study the support recovery accuracy in the backbone set (i.e., fraction of relevant features that are included in the backbone set) as function of the subproblem number $m \in [M] = [30]$, that is, we report the accuracy for the same run of \verb|BB|, after solving the first subproblem, after solving the second subproblem, and so forth. The results are presented in Figure \ref{fig:bb_regression_num_subprob}. We make the following remarks:
\begin{itemize}
    \item[-] The blue line corresponds to {\color{mr}the support recovery accuracy in the $m$-th subproblem (i.e., fraction of relevant features that are selected and added to the backbone set in the $m$-th subproblem)}, as function of the subproblem number $m$. Since the $m$-th subproblem's feature set consists of a fraction $\beta=0.2$ of the total number $\alpha p = 10^4$ of features, we would expect to select a fraction $\beta$ of the relevant features in the $m$-th subproblem (assuming that we solve subproblems perfectly). We observe that, on average, the fraction of relevant features selected in the $m$-th subproblem is slightly higher; this is due to the more informed sampling scheme used in the \texttt{construct\_subproblems} function (Algorithm \ref{alg:construct_subproblems}).
    
    \item[-] The orange line corresponds to {\color{mr}the support recovery accuracy in total after $m$ subproblems (i.e., fraction of relevant features that have been selected overall, across all $m$ subproblems, after solving the $m$-th subproblem)}, as function of the subproblem number $m$. With all other backbone parameters being fixed, as the number of subproblems increases, the support recovery accuracy in the backbone set increases and stabilizes after few subproblems, at the expense of an increased computational time. In this case, the backbone set achieves perfect accuracy after $M=15$ subproblems.
\end{itemize}

% 8.1.2 Number of Subproblems in BB vs Number of Trees in RF

% In this experiment, we test howBBperforms as the number of subproblems increasesin comparison with aRFmodel that consists of the same number of trees.

\begin{figure}[htbp] 
    \centering
    \subfigure[Number of Subproblems: Support recovery accuracy in each subproblem and in total]{\includegraphics[width=0.49\textwidth]{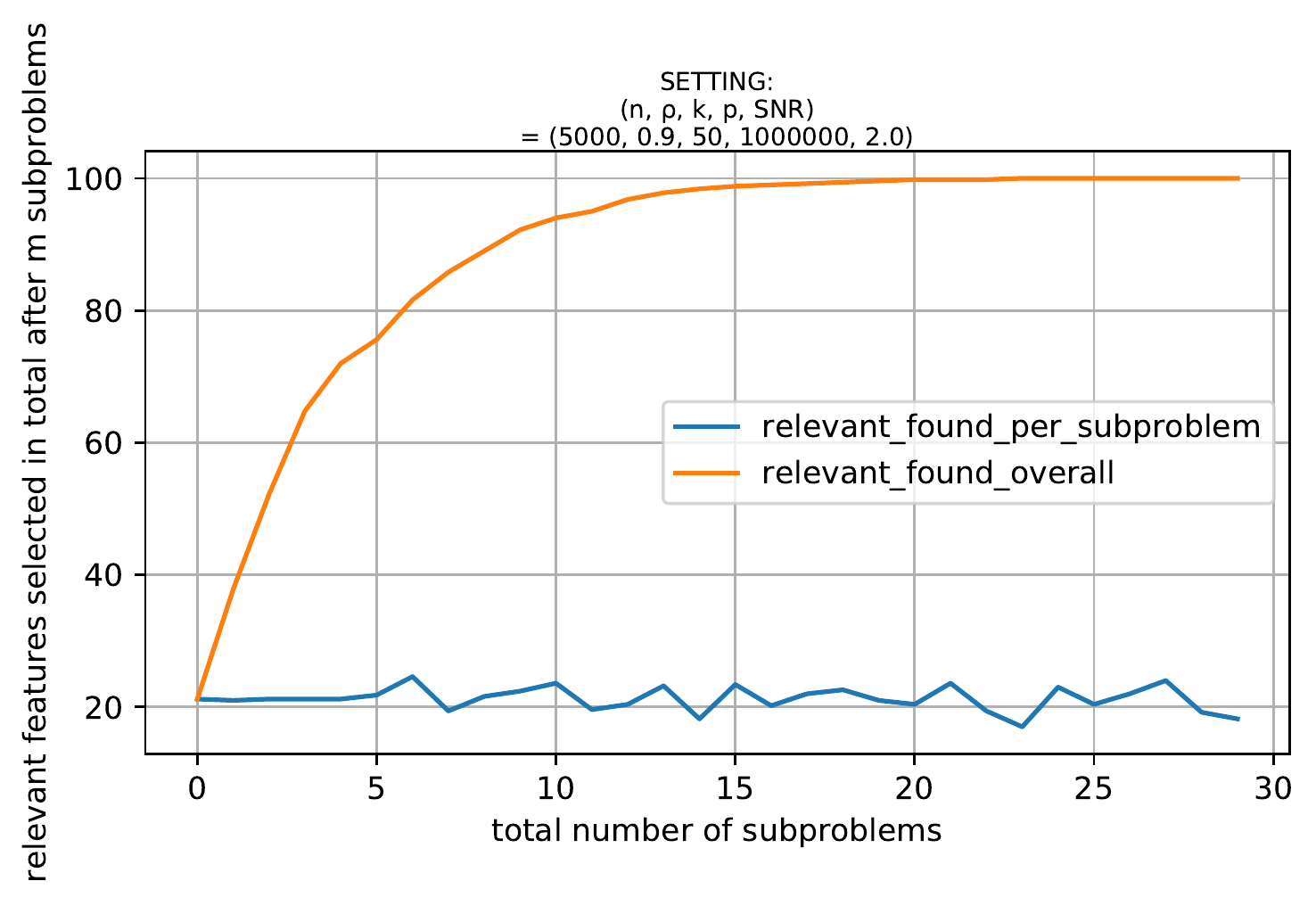} \label{fig:bb_regression_num_subprob}}
    \subfigure[Function \texttt{construct\_subproblems}: Support recovery accuracy in the features sampled in each subproblem and in total]{\includegraphics[width=0.48\textwidth]{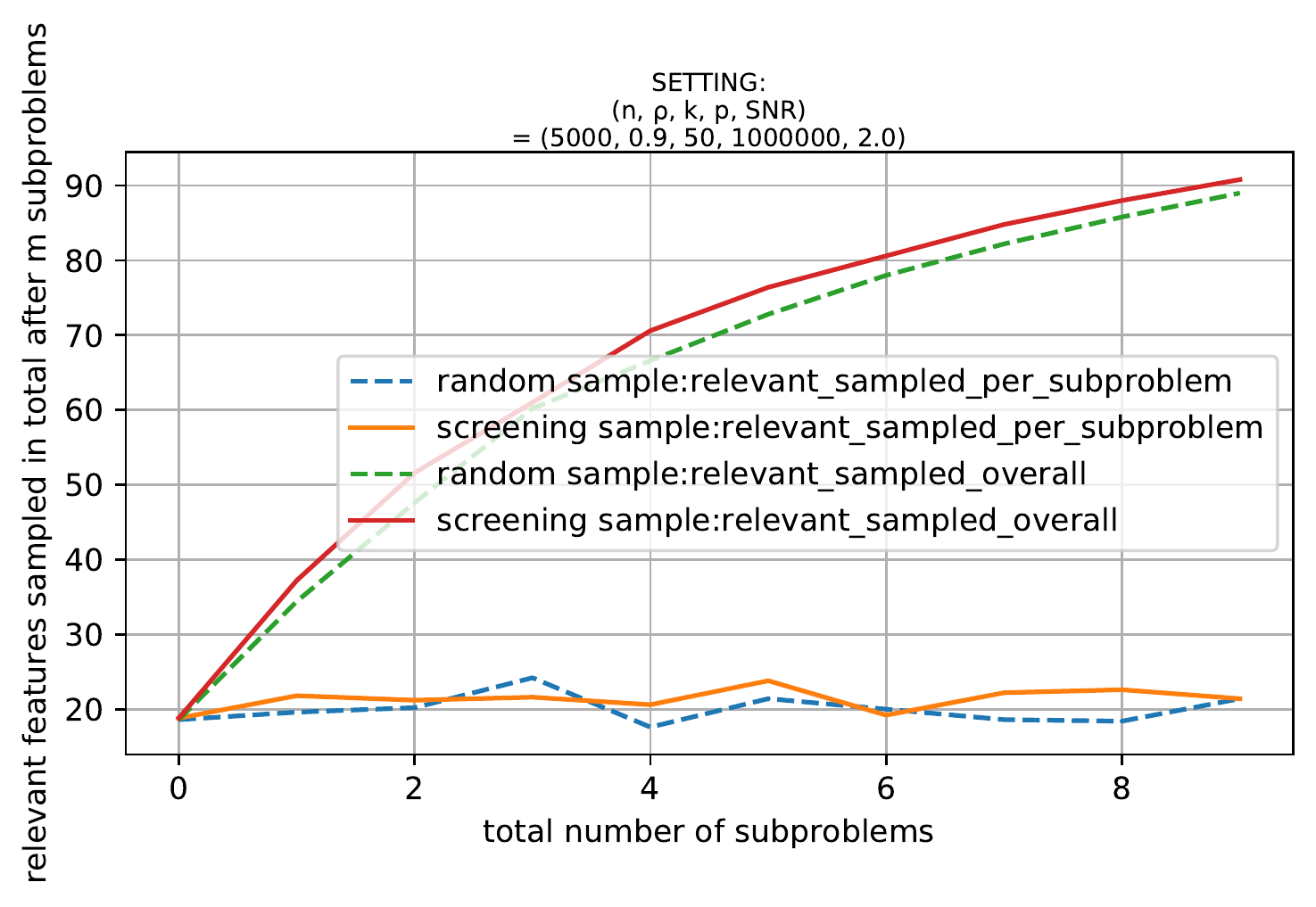} \label{fig:bb_regression_build_subprob}} 
    % \subfigure[Subproblem feature sets' accuracy]{\includegraphics[width=0.49\textwidth]{figs/regression_build_subprob_2.0_0.9_1000000_50_sampled_features_accuracy.pdf}}
    % \subfigure[Computational Time (sec)]{\includegraphics[width=0.49\textwidth]{figs/regression_build_subprob_2.0_0.9_1000000_50_t_total.pdf}}
    \hfill
    \subfigure[Function \texttt{solve\_subproblem}: Support recovery accuracy in the backbone set]{\includegraphics[width=0.48\textwidth]{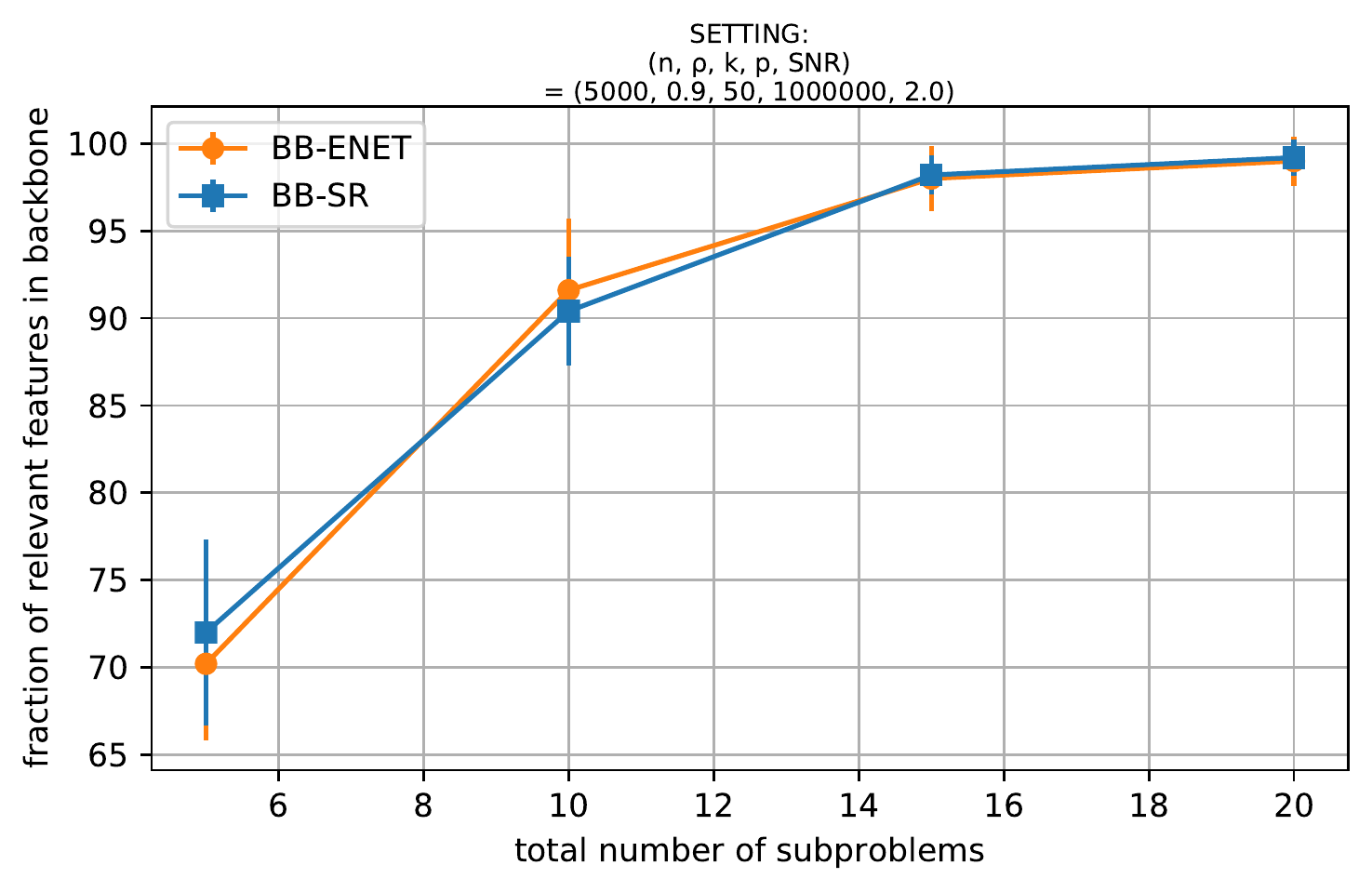} \label{fig:bb_regression_solve_subprob_a}}
    \subfigure[Function \texttt{solve\_subproblem}: Backbone size]{\includegraphics[width=0.48\textwidth]{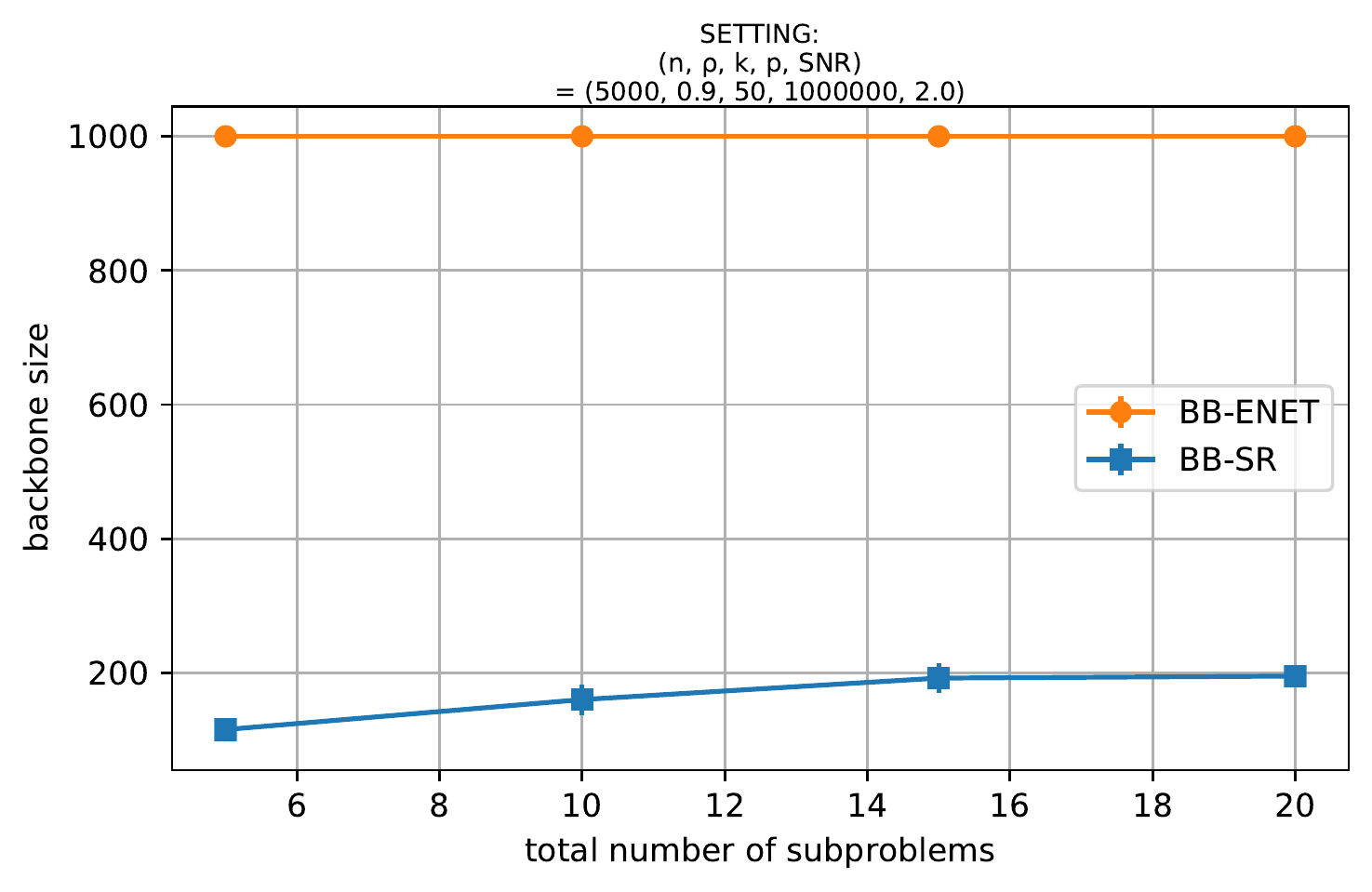} \label{fig:bb_regression_solve_subprob_b}} 
    \hfill
    % \subfigure[Function \texttt{solve\_subproblem}: \\ Computational time (sec)]{\includegraphics[width=0.48\textwidth]{figs/regression_solve_subprob_5000_0.9_50_1000000_2.0_t_total.pdf} \label{fig:bb_regression_solve_subprob_c}}
    \caption{Hyperparameter $M$ and components of the backbone method.}
    \label{fig:bb_regression_components}
\end{figure}

\paragraph{Function \texttt{construct\_subproblems}.} In this experiment, we examine the impact of the \texttt{construct\_subproblems} function on \verb|BB|. We compare two approaches for the \texttt{construct\_subproblems} function. The first one is based on constructing subproblems' feature sets by sampling features uniformly at random, as per the theoretical analysis in Section \ref{subsec:sr-theory} and in Appendix \ref{sec:appendix-proof}; we call this approach \texttt{random\_sample}. The second one is the non-uniformly random sampling approach described in Section \ref{subsec:backbone-construct}; we call this approach \texttt{screening\_sample}. The results are presented in Figure \ref{fig:bb_regression_build_subprob}. The dashed lines correspond to {\color{mr}the support recovery accuracy in the features that are sampled in the $m$-th subproblem's feature set and overall after $m$ subproblems (across all $m$ subproblems)}, as function of the subproblem number $m$, under the \texttt{random\_sample} approach; the solid lines correspond to the \texttt{screening\_sample} approach. With all other backbone parameters being fixed, we observe the following:
\begin{itemize}
    \item The \texttt{screening\_sample} approach samples a slightly larger number of relevant features in the feature set of each subproblem.
    
    \item The \texttt{screening\_sample} approach samples a notably larger number of relevant features across all subproblems. This is due to the fact that, when the \texttt{screening\_sample} approach does not sample a relevant feature in a subproblem's feature set, it is more likely to sample another relevant feature in its place; this is not the case for the \texttt{random\_sample} approach.
\end{itemize}

\begin{sloppypar}
\paragraph{Function \texttt{solve\_subproblem}.} In this experiment, we explore the impact of the \texttt{solve\_subproblem} function on \verb|BB|. We compare two approaches for the \texttt{solve\_subproblem} function: in the first approach, we use \texttt{SR-REL} to solve subproblems, whereas in the second, we use \texttt{ENET} (we set $\mu=1$ so that the pure lasso, which generally leads to sparser models compared to the elastic net, is used). With all other backbone parameters being fixed, we observe the following:
\begin{itemize}
    % \item The computational time for the \texttt{SR-REL}-based approach is higher and more sensitive to an increase in the number of subproblems.
    
    \item The two approaches perform comparably in terms of the support recovery accuracy in the backbone set (Figure \ref{fig:bb_regression_solve_subprob_a}).
    
    \item The \texttt{ENET}-based approach results in a large backbone set, which, in fact, exceeds the limit of $B_{\max}=1,000$ features. We conclude that this approach does not produce sufficiently sparse solutions to the subproblems (Figure \ref{fig:bb_regression_solve_subprob_b}).
\end{itemize}
\end{sloppypar}

{\color{mr}
\paragraph{Number of Features Selected by \texttt{screen}.} In this experiment, we test the impact on the backbone set of the hyperparameter $\alpha$, which determines what fraction of the features will be selected by the \verb|screen| function. Through this experiment, we aim to shed more light on the impact of the different phases of our approach; specifically, we investigate the separate effect of the screening phase and the backbone construction phase, as well as their interaction.

For this experiment, we reduce the total number of features in the data to $p=10^5$, as we need to solve problems with all $p$ features (and hence we would need an excessive amount of memory). We vary $\alpha$ such that $\alpha p \in \{k, 2k, 10k, 100k, 10^3 k, p\}$ and fix the maximum backbone size to $B=2k$. By doing so: 
\begin{itemize}
    \item[-] When $\alpha p\leq2k$, the backbone construction phase is omitted (because the number of features selected by screen is already within the maximum backbone size) and we simply apply sparse regression to the screened features.
    \item[-] When $\alpha p=k$, we simply select features via screen (without even applying sparse regression).
    \item[-] When $2k < \alpha p < p$, we apply the backbone method in its entirety.
    \item[-] When $\alpha p=p$, we omit the screening step.
\end{itemize}
We  set  the  remaining  backbone  hyperparameters as follows. For a fair comparison, we fix the $\beta = 0.2$. To capture $M$’s dependency on $\alpha p$, we set the number of subproblems to $M = 5+ \frac{\log(\alpha pk)}{5\log(\frac{1}{1-\beta})}$. This value is obtained from the theoretical expression for $M$. As we are interested in studying the trade-off between the screening step and the backbone construction phase, we report the computational time required for this part of the backbone method. Figures \ref{fig:bb_regression_variables_screened-a} and \ref{fig:bb_regression_variables_screened-b} suggest the following:
\begin{itemize}
    \item[-] As $\alpha$ increases, the computational time increases (since more, bigger subproblems need to be solved) and the accuracy also increases (since fewer relevant features will be missed during \verb|screen|). There is a trade-off between the two.
    \item[-] There exists some threshold $\alpha_0$ such that, for $\alpha \geq \alpha_0$, the method succeeds (i.e., recovers tha majority of relevant features). Setting $\alpha$ to the smallest possible value, i.e., $\alpha = \alpha_0$, speeds up the process, as we get rid of irrelevant features. Even without any screening, i.e., for $\alpha=1$, the method succeeds and the overhead in computational time is not prohibitive.
    \item[-] The backbone construction phase is significantly more selective than screening, i.e., it has higher support recovery accuracy and lower support recovery false alarm rate. In all cases where the backbone construction phase is applied, the size of the backbone set is $|\mathcal{B}| \leq B$ and the support recovery accuracy is substantially higher compared to selecting $B$ features via \verb|screen|.
\end{itemize}

%a smaller value of $\alpha$ (which results in fewer features surviving the screening step) is beneficial, provided that no relevant features are missed during this step. With all other backbone parameters being fixed, as $\alpha$ increases, it becomes more likely to miss a relevant feature from the backbone set and it takes more to solve the subproblems (as subproblems are both bigger and harder).

\begin{figure}[htbp] 
    \centering
    \subfigure[Number of Features Selected by \texttt{screen}: Support recovery accuracy in the backbone set]{\includegraphics[width=0.49\textwidth]{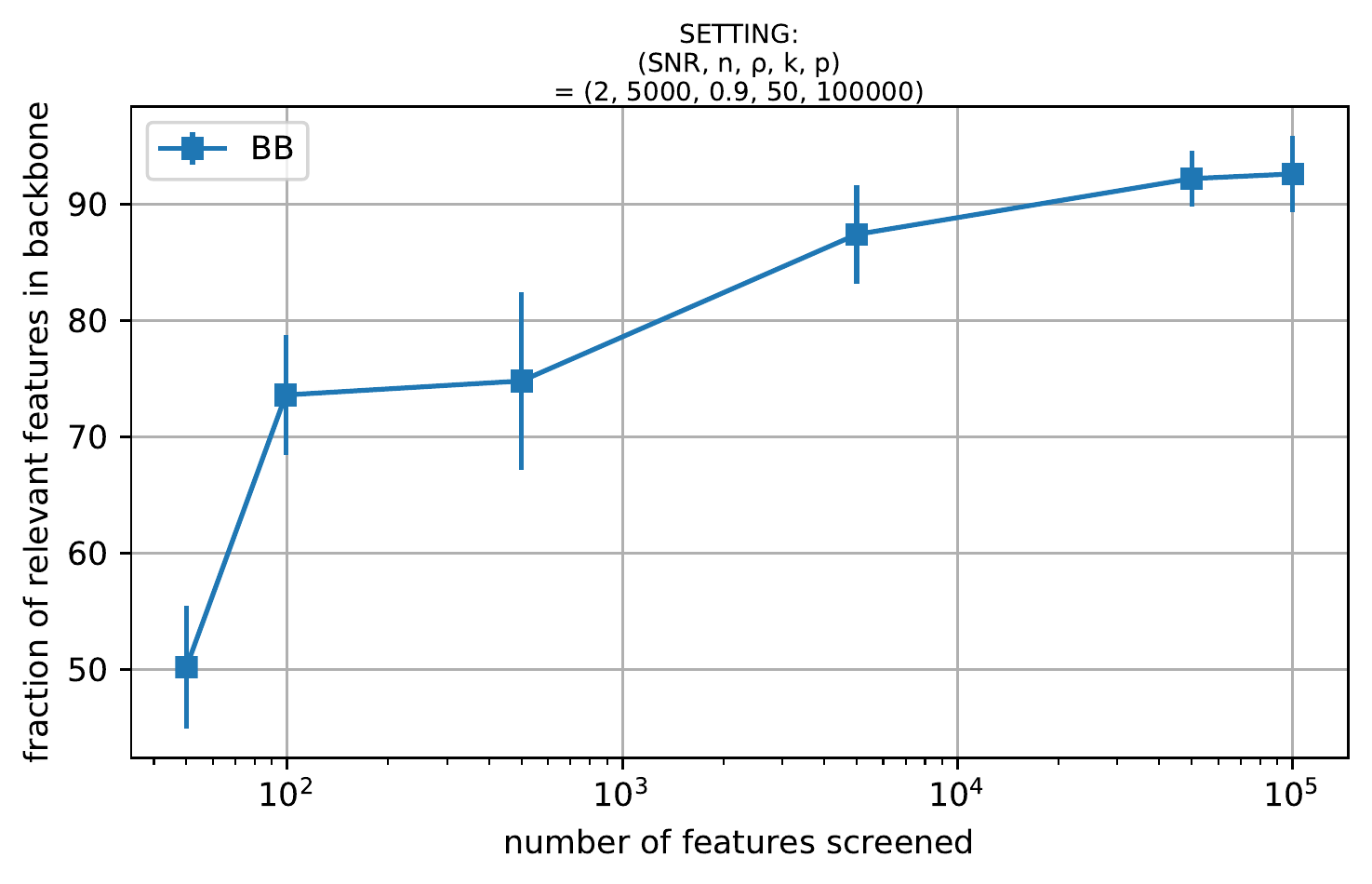}\label{fig:bb_regression_variables_screened-a}}
    \hfill
    \subfigure[Number of Features Selected by \texttt{screen}: Computational time for the screening and backbone construction phases (sec)]{\includegraphics[width=0.49\textwidth]{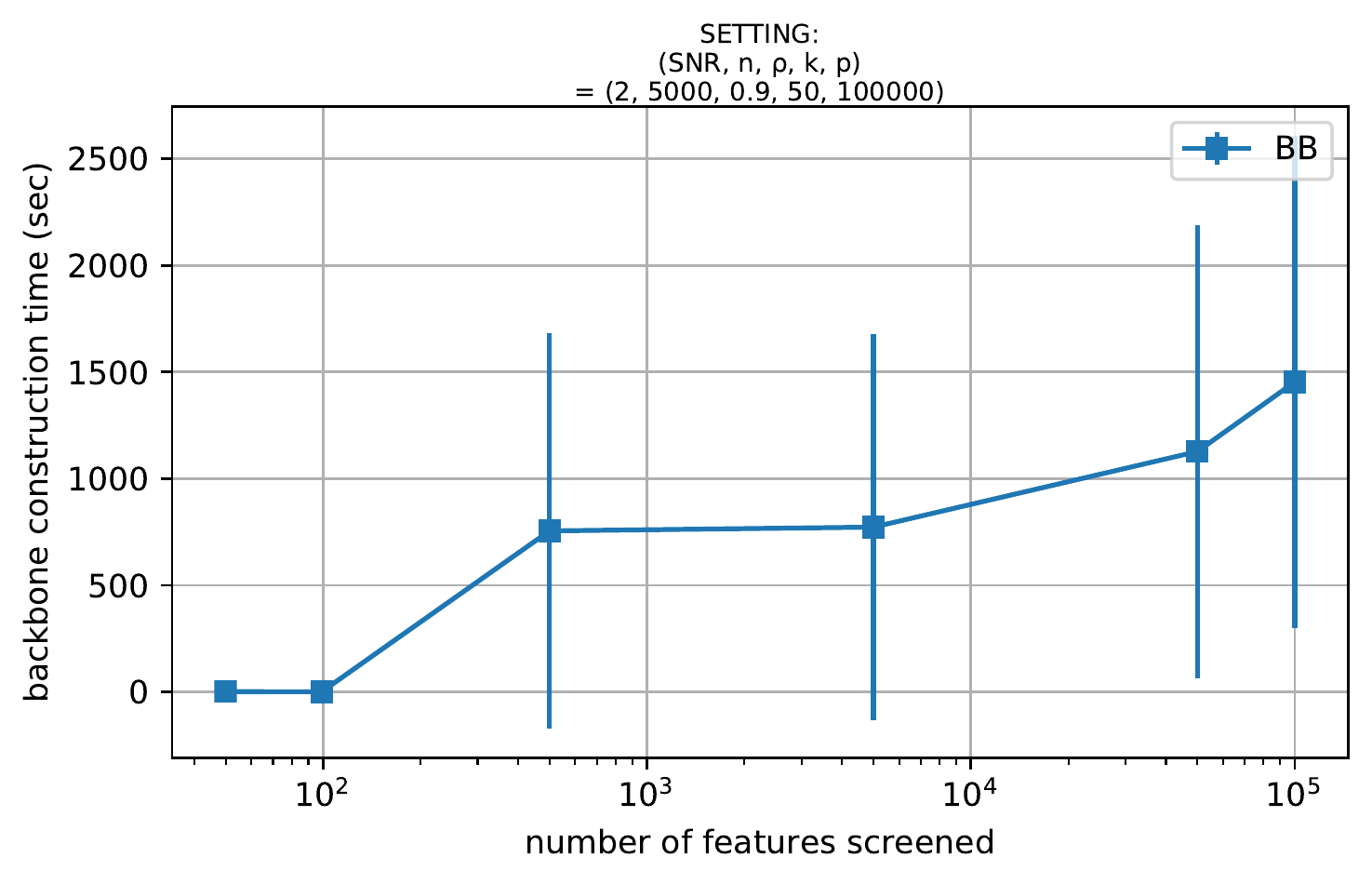}\label{fig:bb_regression_variables_screened-b}}
    \subfigure[Number of Features per Subproblem: Support recovery accuracy in the backbone set]{\includegraphics[width=0.49\textwidth]{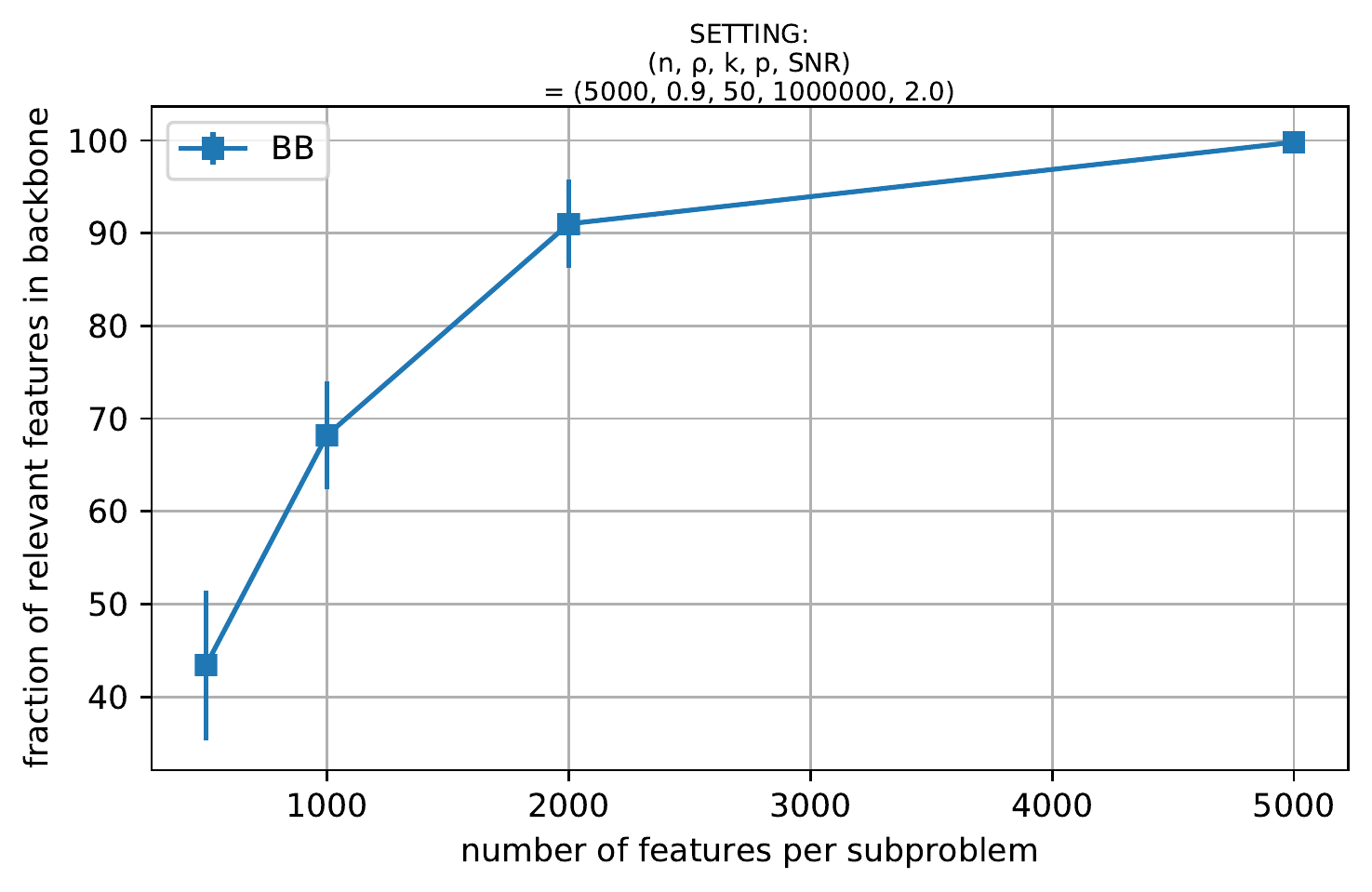}\label{fig:bb_regression_variables_subprob-a}}
    \hfill
    \subfigure[Number of Features per Subproblem: Computational time (sec)]{\includegraphics[width=0.49\textwidth]{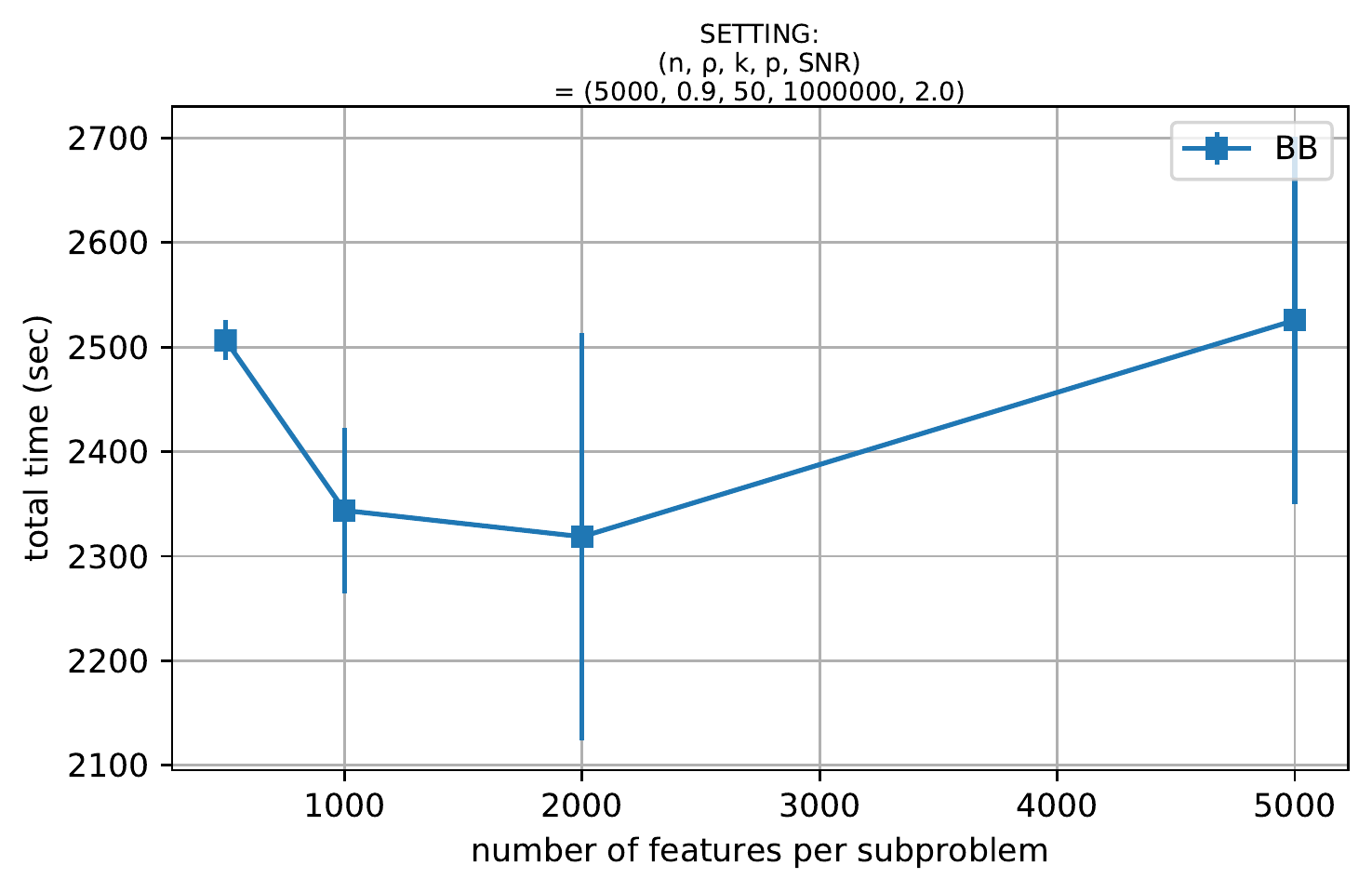}\label{fig:bb_regression_variables_subprob-b}}
    \subfigure[Maximum Backbone Size: Support recovery accuracy in the backbone set]{\includegraphics[width=0.49\textwidth]{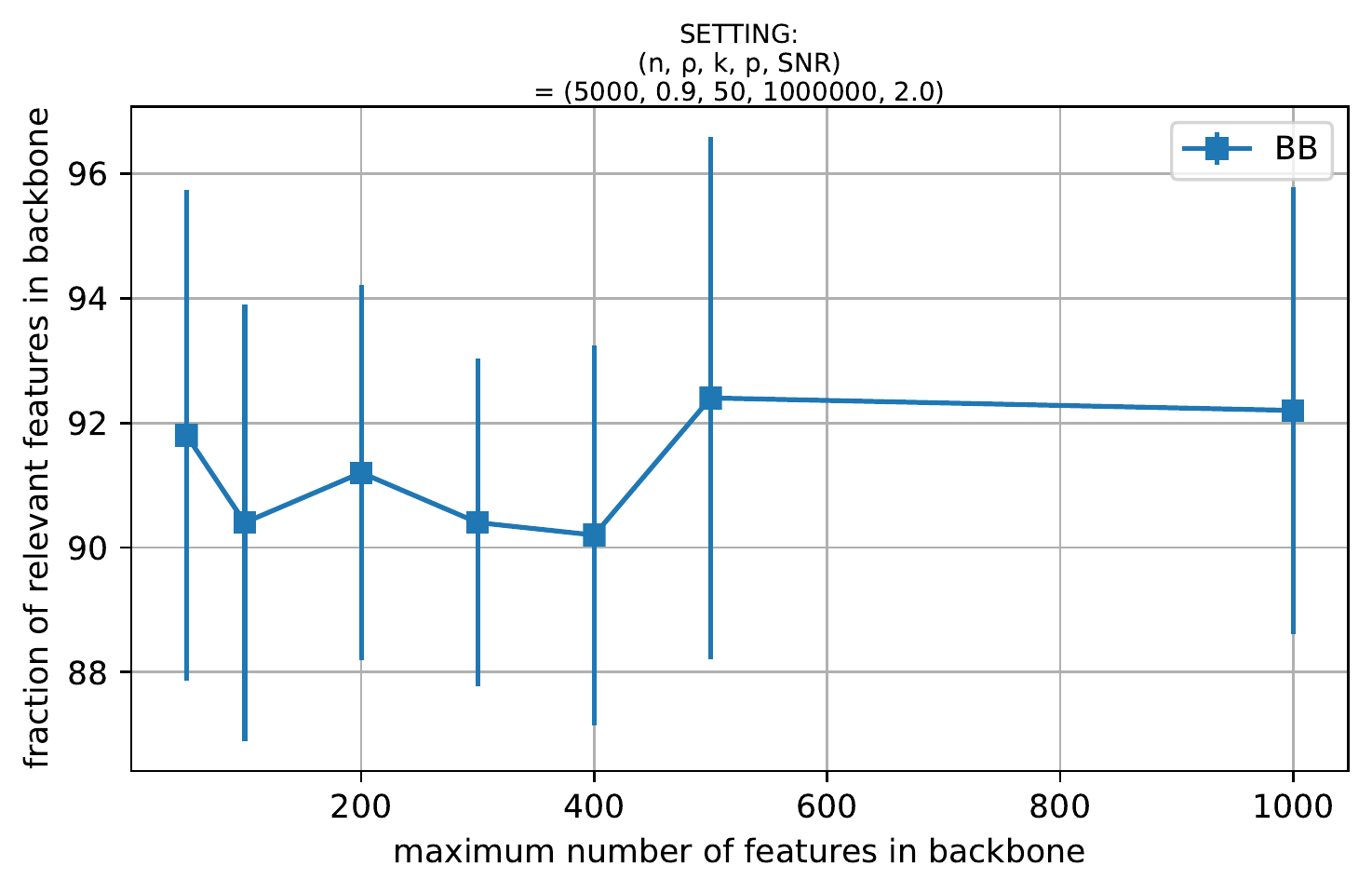}\label{fig:bb_regression_max_backbone-a}}
    % \subfigure[Maximum Backbone Size: \\ MIO solver optimality gap]{\includegraphics[width=0.49\textwidth]{figs/regression_max_backbone_size_5000_0.9_50_1000000_2.0_reduced_gap.pdf}}
    \subfigure[Maximum Backbone Size: Computational time (sec)]{\includegraphics[width=0.49\textwidth]{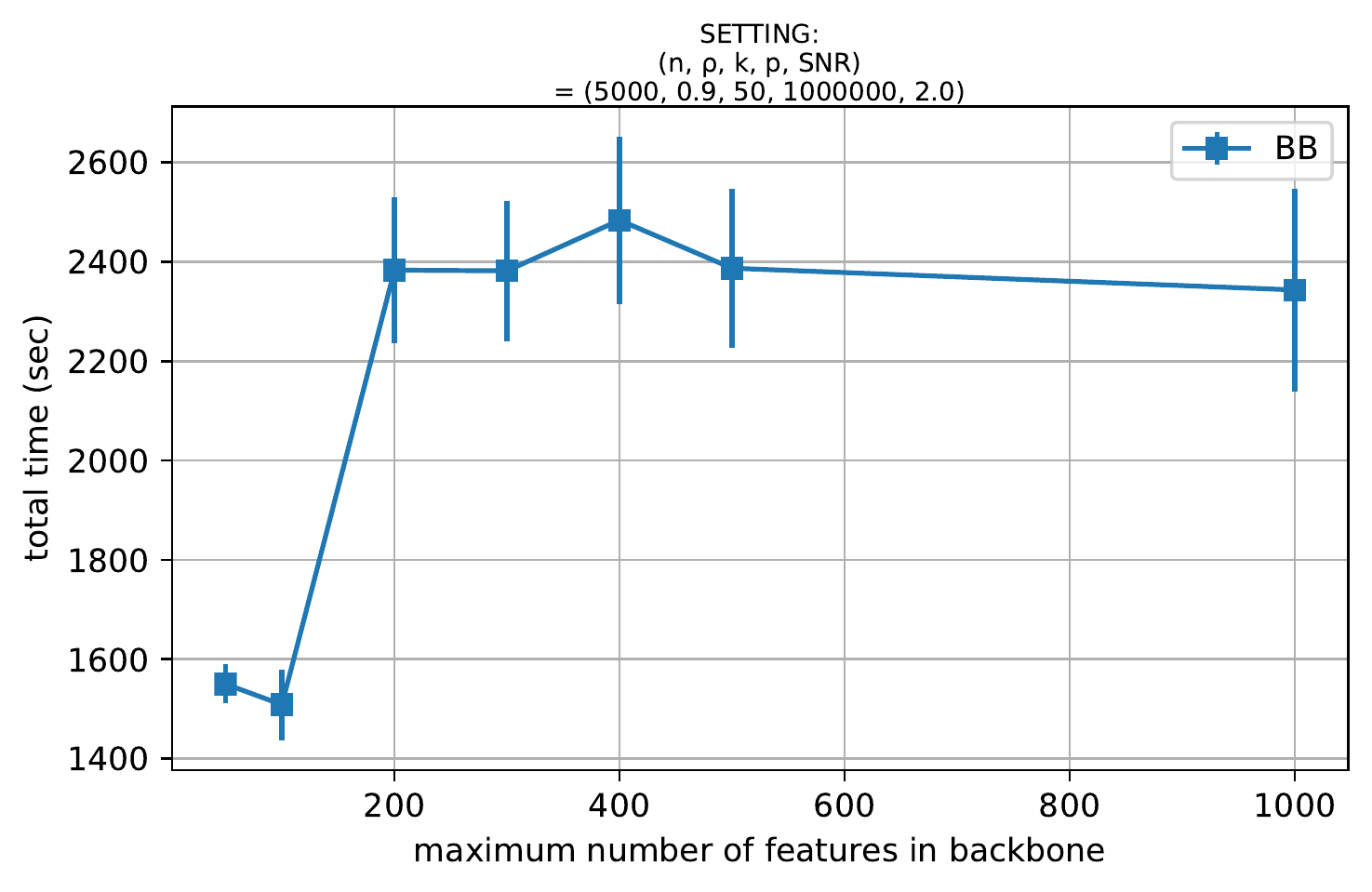}\label{fig:bb_regression_max_backbone-b}}
    \caption{Hyperparameters $\alpha$, $\beta$, $B_{\max}$.}
    \label{fig:bb_regression_variables}
\end{figure}
}

\paragraph{Number of Features per Subproblem.} In this experiment, we test the impact on the backbone set of the subproblem size, which we control through the hyperparameter $\beta$. Figures \ref{fig:bb_regression_variables_subprob-a} and \ref{fig:bb_regression_variables_subprob-b} indicate that a value $\beta \approx 0.5$ is ideal for the problems that we consider. With all other backbone parameters being fixed, as $\beta$ increases, we observe {\color{mr}the following:
\begin{itemize}
    \item[-] The support recovery accuracy in the backbone set increases.
    \item[-] The computational time increases, except when the subproblem size is too small. In the latter case, the decreased support recovery accuracy in the backbone set means that an increased number of relevant features are not included therein and, therefore, solving the reduced problem becomes harder (as it contains a smaller amount of signal).
\end{itemize}
Intuitively, we want $\beta$ to be large enough, so that enough signal is contained in the subproblems, and small enough, so that the subproblems can be solved fast.
}

\paragraph{Maximum Backbone Size.} In this experiment, we explore the impact on the backbone set of the maximum backbone size, controlled by the hyperparameter $B_{\max}$. Figures \ref{fig:bb_regression_max_backbone-a} and \ref{fig:bb_regression_max_backbone-b} indicate that, with all other backbone parameters being fixed, as $B_{\max}$ increases, the support recovery accuracy in the backbone set slightly increases; the computational time is not affected, except when the maximum backbone size is too small. When this is the case, the reduced problem can be solved very fast and therefore the overall computational time drops; this, however, comes with a higher risk of not including relevant features in the backbone set. We also remark that the maximum backbone size and number of iterations of the hierarchical backbone algorithm (Algorithm \ref{alg:backbone}) are connected, since decreasing the maximum backbone size will likely result in more iterations. 

\section{Computational Results on Real-World Data} \label{sec:results-real}

In this section, we empirically evaluate the backbone method on real-world datasets and compare its performance with various baselines and state-of-the-art alternatives. The metrics and algorithms/software that we use are the same as those outlined in Sections \ref{subsec:results-synth-metrics} and \ref{subsec:results-synth-algorithms}, respectively.

\subsection{Datasets} \label{subsec:results-real-data}

We experiment on 2 regression and 2 classification datasets from the UCI machine learning repository \citep{Dua2019uci}:
\begin{itemize}
    \item[-] \emph{Communities and Crime:}
    This is a regression problem; the goal is to predict the crime rate in various communities in the US \citep{redmond2002data}. 
    We remove all features whose values are missing in more than 10 data points; then, we remove all data points that still have any missing value. 
    The resulting dataset consists of $n=1,993$ data points and $p=100$ features.
    The original dataset is available at \url{https://archive.ics.uci.edu/ml/datasets/Communities+and+Crime}.
    
    \item[-] \emph{Housing:}
    This is a regression problem; the goal is to predict housing prices in Boston. 
    The original dataset has no missing values; we expand the dataset by adding squared for all features (except for the binary ones) and interaction terms for all pairs of features. 
    The resulting dataset consists of $n=506$ data points and $p=103$ features.
    The original dataset is available at \url{https://www.csie.ntu.edu.tw/~cjlin/libsvmtools/datasets/regression.html#housing}.
    
    \item[-] \emph{Breast Cancer:}
    This is a binary classification problem; the goal is to predict whether a breast cancer instance is benign or malignant \citep{wolberg1990multisurface, zhang1992selecting}. We perform no preprocessing to the dataset, which consists of $n=599$ data points and $p=9$ features.
    The original dataset is available at \url{https://archive.ics.uci.edu/ml/datasets/Breast+Cancer+Wisconsin+%28Original%29}.
    
    \item[-] \emph{Ionoshphere:}
    This is a binary classification problem; the goal is to classify radar returns from the ionosphere \citep{sigillito1989classification}. We perform no preprocessing to the dataset, which consists of $n=351$ data points and $p=33$ features.
    The original dataset is available at \url{https://archive.ics.uci.edu/ml/datasets/Ionosphere}.
    
\end{itemize}

{\color{mr}Along the lines of} \cite{hazimeh2020fast}, we append to the data matrix 1,000 random permutations of each (original) feature (column) for both regression and classification instances. By doing so, we have a way to assess the support recovery performance of each method by measuring the fraction of features in the solution that are original features, that is, they were not generated according to the {\color{mr}aforementioned random permutation process}. Moreover, the expanded datasets become truly ultra-high dimensional, which is the regime that we are interested in{\color{mr}: in the sparse linear regression case, the number of features in the expanded dataset is in the order of $10^5$ and, in the classification tree case, the number of features is in the order of $10^4$.}

We conduct our experiments as follows. We randomly split, 5 times independently, each original dataset (which we have not expanded yet) into training and testing, at a ratio of $\frac{n_{\text{train}}}{n_{\text{test}}} = 4$. For each split, we expand, 5 times independently, the resulting training and testing sets, using the process outlined {\color{mr}in the previous paragraph} {\color{mr}(for each training-testing set split, we apply each independent expansion separately to the training and the testing set)}. Therefore, for each split, we obtain 5 different realizations of the noisy features. In total, we conduct 25 experiments per problem. We report both the mean and the standard deviation of each metric.

\subsection{Sparse Linear Regression} \label{subsec:results-real-sr}

In this section, we present the results for the regression datasets, on which we apply sparse linear regression methods. 

We tune \verb|BB| as follows. We select the \texttt{screen} function's parameter $\alpha$ such that all but $10,000$ features are eliminated. We set $\beta=0.5$ and solve $M=10$ subproblems. We set $B_{\max}=500.$ We solve the subproblems using \texttt{SR-REL}; in the $m$-th subproblem, we cross-validate 3 values for $k_m \in \{ 10,20,30 \}$ and set $\gamma_m = \frac{1}{\sqrt{n_m}}$. We solve the reduced problem using \texttt{SR} with a time limit of 5 minutes; we cross-validate 5 values for $k \in \{ 10,20,30,40,50 \}$ and 5 values for the hyperparameter $\gamma$. 

We benchmark \verb|BB| against some of the methods outlined in Section \ref{subsec:results-synth-algorithms}, tuned as explained below. For \verb|SIS-ENET|, we cross-validate the number of features that are eliminated; then, we apply \verb|ENET| to the selected features, tuned as described in Section \ref{subsec:sr-implementation-backbone}; more specifically, we cross-validate 5 values for the hyperparameter $\mu \in [0,1]$ (the pure lasso model and the ridge regression model are included in the cross-validation procedure); we discard from the final model any feature whose corresponding regressor has magnitude $\leq 10^{-6}$. For \verb|RFE|, we follow the process outlined in Section \ref{subsec:results-synth-algorithms} to eliminate features; then, we apply \verb|SR| on the selected features, tuned in the exact same way that we tune \verb|SR| for the reduced problem in \verb|BB|. For \verb|DECO|, we follow the process outlined in Section \ref{subsec:results-synth-algorithms}; we tune \verb|ENET| applied to the selected features exactly as in \verb|SIS-ENET|. Finally, we also compare against an oracle model, whereby we apply \verb|SR| on the original features (i.e., we exclude all noisy features from the model); therefore, this approach can serve as an upper bound for the performance of the remaining methods.

In Table \ref{tab:communities}, we present the results for the communities case study, whereas Table \ref{tab:housing} reports the results for the housing case study. We make the following observations:
\begin{itemize}
    \item Out-of-sample R$^2$: In both case studies, \verb|BB| achieves an increased out-of-sample R$^2$ compared to \verb|RFE| and \verb|DECO|. In the communities case study, \verb|BB| notably outperforms \verb|SIS-ENET| and approaches the performance of the oracle model. In the housing case study, \verb|SIS-ENET| is particularly effective, in that the screening step manages to eliminate all noisy features, and hence almost matches the performance of \verb|SR-ORACLE|.
    
    \item Support recovery accuracy and sparsity: In both case studies, \verb|BB| uses a fraction of 15-20$\%$ of original features and results in models with sparsity close to that of \verb|SR-ORACLE|. In terms of the fraction of features used that are noisy, \verb|BB| achieves the lowest rate in the communities case study, but is outperformed by \verb|SIS-ENET| and \verb|RFE| in the housing case study.
    
    \item Optimality gap: In both case studies, \verb|BB| achieves an optimality gap that is comparable to that of \verb|SR-ORACLE| and hence returns a solution with optimality guarantees, albeit for the reduced problem.
    
    \item Computational time: In both case studies, the solution obtained via \verb|BB| is computed in less than 2 hours, and in time comparable to that of \verb|SR-ORACLE|. \verb|RFE| is substantially slower, taking several hours to run, whereas \verb|SIS-ENET| and \verb|DECO|, which do not involve applying \verb|SR|, are computed in less than 2 minutes. 
\end{itemize}

% \begin{figure}[htbp] 
%     \centering
%     \subfigure[Out-of-sample R$^2$]{\includegraphics[width=0.24\textwidth]{figs/regression_real_c_o_m_m_u_n_i_t_i_e_s_score.pdf}} 
%     \subfigure[Support recovery accuracy]{\includegraphics[width=0.24\textwidth]{figs/regression_real_c_o_m_m_u_n_i_t_i_e_s_acc.pdf}}
%     \subfigure[Sparsity: support size]{\includegraphics[width=0.24\textwidth]{figs/regression_real_c_o_m_m_u_n_i_t_i_e_s_sparsity.pdf}}
%     \subfigure[MIO solver optimality gap]{\includegraphics[width=0.24\textwidth]{figs/regression_real_c_o_m_m_u_n_i_t_i_e_s_reduced_gap.pdf}}
%     % \subfigure[Computational time (sec)]{\includegraphics[width=0.19\textwidth]{figs/regression_real_c_o_m_m_u_n_i_t_i_e_s_t_total.pdf}}
%     \caption{Results for the communities case study.}
%     \label{fig:bb_regression_real_communities}
% \end{figure}

% \begin{figure}[htbp] 
%     \centering
%     \subfigure[Out-of-sample R$^2$]{\includegraphics[width=0.24\textwidth]{figs/regression_real_h_o_u_s_i_n_g_score.pdf}} 
%     \subfigure[Support recovery accuracy]{\includegraphics[width=0.24\textwidth]{figs/regression_real_h_o_u_s_i_n_g_acc.pdf}}
%     \subfigure[Sparsity: support size]{\includegraphics[width=0.24\textwidth]{figs/regression_real_h_o_u_s_i_n_g_sparsity.pdf}}
%     \subfigure[MIO solver optimality gap]{\includegraphics[width=0.24\textwidth]{figs/regression_real_h_o_u_s_i_n_g_reduced_gap.pdf}}
%     \caption{Results for the housing case study.}
%     \label{fig:bb_regression_real_housing}
% \end{figure}

\begin{table}[htbp]
\resizebox{\textwidth}{!}{
\begin{tabular}{|c|c|c|c|c|c|c|}
\hline
\rowcolor[HTML]{B0B3B2} 
\textbf{} & \textbf{R$^2$} & \textbf{SR-ACC} & \textbf{SR-FA} & \textbf{Sparsity} & \textbf{OG} & \textbf{time (sec)} \\ \hline
\cellcolor[HTML]{D4D4D4}\texttt{SR-ORACLE} & 0.649 (0.017) & 32.4 (13.626) & 0.0 (0.0) & 32.4 (13.626) & 0.34 (0.375) & 3565.03 (387.337) \\ \hline
\cellcolor[HTML]{D4D4D4}\texttt{SIS-ENET} & 0.556 (0.035) & 43.84 (10.907) & 76.773 (8.591) & 299.28 (350.675) & - & 31.061 (9.529) \\ \hline
\cellcolor[HTML]{D4D4D4}\texttt{RFE} & 0.57 (0.018) & 11.4 (2.082) & 71.474 (5.188) & 39.96 (0.2) & - & 54388.664 (11261.57) \\ \hline
\cellcolor[HTML]{D4D4D4}\texttt{DECO} & 0.573 (0.031) & 29.12 (3.528) & 67.178 (12.332) & 103.32 (46.992) & - & 108.263 (29.504) \\ \hline
\cellcolor[HTML]{D4D4D4}\texttt{BB} & 0.615 (0.024) & 18.36 (4.734) & 53.647 (11.718) & 40.36 (7.353) & 0.147 (0.305) & 2505.225 (910.012) \\ \hline
\end{tabular}
}
\caption{Results for the communities case study.}
\label{tab:communities}
\end{table}

\begin{table}[htbp]
\resizebox{\textwidth}{!}{
\begin{tabular}{|c|c|c|c|c|c|c|}
\hline
\rowcolor[HTML]{B0B3B2} 
\textbf{} & \textbf{R$^2$} & \textbf{SR-ACC} & \textbf{SR-FA} & \textbf{Sparsity} & \textbf{OG} & \textbf{T} \\ \hline
\cellcolor[HTML]{D4D4D4}\texttt{SR-ORACLE} & 0.864 (0.05) & 41.553 (10.305) & 0.0 (0.0) & 42.8 (10.614) & 0.475 (0.425) & 3928.42 (272.247) \\ \hline
\cellcolor[HTML]{D4D4D4}\texttt{SIS-ENET} & 0.863 (0.038) & 50.563 (11.062) & 0.0 (0.0) & 52.08 (11.394) & - & 6.144 (2.066) \\ \hline
\cellcolor[HTML]{D4D4D4}\texttt{RFE} & 0.712 (0.064) & 15.961 (3.761) & 58.453 (10.988) & 40.4 (4.546) & - & 13462.06 (2958.84) \\ \hline
\cellcolor[HTML]{D4D4D4}\texttt{DECO} & 0.757 (0.055) & 27.379 (2.831) & 73.791 (14.516) & 140.48 (71.93) & - & 28.402 (8.599) \\ \hline
\cellcolor[HTML]{D4D4D4}\texttt{BB} & 0.765 (0.046) & 16.66 (3.715) & 64.464 (9.278) & 48.76 (3.431) & 0.531 (0.445) & 4597.401 (1318.081) \\ \hline
\end{tabular}
}
\caption{Results for the housing case study.}
\label{tab:housing}
\end{table}

\subsection{Classification Trees} \label{subsec:results-real-trees}

In this section, we present the results for the classification datasets, on which we apply classification tree methods. 

We tune \verb|BB| as follows. We select the \texttt{screen} function's parameter $\alpha$ such that all but $1,000$ features are eliminated. We set $\beta=0.5$ and solve $M=15$ subproblems. We set $B_{\max}=100.$ We solve the subproblems using \texttt{CART}; in the $m$-th subproblem, we cross-validate $D_m \in \{ 2,3,4,5 \}$ (we also cross-validate the minbucket and complexity parameter of \verb|CART|). We solve the reduced problem using \texttt{OCT}; we cross-validate $D \in \{ 3,4,5,6,7,8 \}$ (we also cross-validate the minbucket and complexity parameter of \verb|OCT|).

We benchmark \verb|BB| against some of the methods outlined in Section \ref{subsec:results-synth-algorithms}, tuned as explained below. The first two baselines are \verb|CART|, tuned in the exact same way as \verb|OCT| in solving the reduced problem for \verb|BB|, and \verb|RF|, in which we use 100 trees, we impose no depth limit for the trees (which is common in practice), and we cross-validate the number of features that are considered in each split in each tree between 5 values. We also benchmark against \verb|RFE|, which is tuned as explained in Section \ref{subsec:results-synth-algorithms}. Finally, we also compare against an oracle model, whereby we apply \verb|OCT| to the original features (i.e., we exclude all noisy features from the model); therefore, this approach can serve as an upper bound for the performance of the remaining methods.

In Table \ref{tab:breast}, we present the results for the breast cancer case study, whereas Table \ref{tab:ionosphere} reports the results for the ionosphere case study. We make the following observations:
\begin{itemize}
    \item Out-of-sample AUC: In the breast cancer case study, \verb|BB| outperforms \verb|CART| and \verb|RFE|, matches the performance of \verb|OCT-ORACLE|, and achieves a 0.03 lower AUC compared to \verb|RF|, while still outputting a single, interpretable tree. In the ionosphere case study, the feature selection based methods, namely, \verb|BB| and \verb|RFE|, outperform \verb|OCT-ORACLE| and \verb|CART|, and compete with \verb|RF|, having a 0.02 lower AUC.
    
    \item Support recovery accuracy and sparsity: In both case studies, \verb|BB| uses more original features compared to \verb|CART| and \verb|RFE|, and results in models that are sparser than those obtained via \verb|OCT-ORACLE|. In terms of the fraction of features used that are noisy, \verb|BB| outperforms \verb|CART| and is outperformed by \verb|RFE|, which generally results in very shallow trees that use the smallest number of features.
    
    \item Computational time: In both case studies, \verb|BB|'s computational time is comparable with that of \verb|CART| and \verb|OCT-ORACLE|. Compared to \verb|RF| and \verb|RFE|, \verb|BB| is approximately 5 times faster in the breast cancer case study and more than 30 times faster in the ionosphere case study.
\end{itemize}

}

\begin{table}[htbp]
\resizebox{\textwidth}{!}{
\begin{tabular}{|c|c|c|c|c|c|}
\hline
\rowcolor[HTML]{B0B3B2} 
\textbf{} & \textbf{AUC} & \textbf{SR-ACC} & \textbf{SR-FA} & \textbf{Sparsity} & \textbf{T} \\ \hline
\cellcolor[HTML]{D4D4D4}\texttt{OCT-ORACLE} & 0.943 (0.023) & 73.778 (11.055) & 0.0 (0.0) & 6.64 (0.995) & 1.364 (0.091) \\ \hline
\cellcolor[HTML]{D4D4D4}\texttt{CART} & 0.917 (0.031) & 28.444 (5.629) & 55.353 (18.613) & 6.4 (2.102) & 6.167 (0.478) \\ \hline
\cellcolor[HTML]{D4D4D4}\texttt{RF} & 0.972 (0.014) & - & - & - & 68.593 (6.154) \\ \hline
\cellcolor[HTML]{D4D4D4}\texttt{RFE} & 0.933 (0.014) & 28.444 (5.629) & 20.711 (26.468) & 3.84 (2.055) & 42.121 (5.391) \\ \hline
\cellcolor[HTML]{D4D4D4}\texttt{BB} & 0.941 (0.022) & 34.222 (9.58) & 46.252 (13.927) & 5.92 (1.498) & 9.815 (0.489) \\ \hline
\end{tabular}
}
\caption{Results for the breast cancer case study.}
\label{tab:breast}
\end{table}

\begin{table}[htbp]
\resizebox{\textwidth}{!}{
\begin{tabular}{|c|c|c|c|c|c|}
\hline
\rowcolor[HTML]{B0B3B2} 
\textbf{} & \textbf{AUC} & \textbf{SR-ACC} & \textbf{SR-FA} & \textbf{Sparsity} & \textbf{T} \\ \hline
\cellcolor[HTML]{D4D4D4}\texttt{OCT-ORACLE} & 0.864 (0.033) & 25.939 (6.724) & 0.0 (0.0) & 8.56 (2.219) & 2.639 (0.161) \\ \hline
\cellcolor[HTML]{D4D4D4}\texttt{CART} & 0.844 (0.021) & 6.061 (0.0) & 61.825 (11.906) & 5.8 (1.893) & 31.731 (3.009) \\ \hline
\cellcolor[HTML]{D4D4D4}\texttt{RF} & 0.891 (0.023) & - & - & - & 314.683 (42.199) \\ \hline
\cellcolor[HTML]{D4D4D4}\texttt{RFE} & 0.877 (0.017) & 6.303 (0.839) & 11.2 (20.478) & 2.56 (1.044) & 801.467 (152.225) \\ \hline
\cellcolor[HTML]{D4D4D4}\texttt{BB} & 0.871 (0.042) & 12.606 (1.134) & 14.667 (10.887) & 4.92 (0.493) & 11.399 (0.25) \\ \hline
\end{tabular}
}
\caption{Results for the ionosphere case study.}
\label{tab:ionosphere}
\end{table}

\section{Concluding Remarks} \label{sec:concl}
In this paper, we developed the backbone method, a novel framework that can be used to train a variety of sparse machine learning models. As we showed, the backbone method can accurately and effectively sparsify the set of possible solutions and, as a result, the MIO formulation that exactly models the learning problem can be solved fast for ultra-high dimensional problems. We gave concrete examples of problems where the backbone method can be applied and discussed in detail the implementation details for the sparse regression problem and the decision tree problem. {\color{changecolor} For the sparse regression problem, we showed that, under certain assumptions and with high probability, the backbone set consists of the {\color{mr}truly relevant} features.}

{\color{changecolor} As far as the sparse regression problem is concerned, our computational study illustrated that the backbone method outperforms or competes with state-of-the-art methods for ultra-high dimensional problems, accurately scales to problems with $p\sim10^7$ features in minutes and $p \sim 10^8$ features in hours, and drastically reduces the problem size in problems with $p\sim10^5$ features hence making the work of exact methods much easier. Regarding the decision tree problem, the backbone method scales to problems with $p\sim10^5$ features in minutes and, assuming that the underlying problem is indeed sparse (in that only few features are involved in splits in the decision tree), outperforms CART, and competes with random forest, while still outputting a single, interpretable tree. In problems with $p\sim10^3$, the backbone method can accurately filter the feature set and compute decision trees that match those obtained by applying the state-of-the-art optimal trees framework to the entire problem.
}

Finally, as we discussed throughout the paper, the backbone method is generic and can be directly applied to any sparse supervised learning model; examples include sparse support vector machines (SVMs) and sparse principal component analysis. In addition, our proposed framework can be extended to non-supervised sparse machine learning problems, such as the clustering problem. Furthermore, the backbone construction phase can naturally be implemented in a parallel/distributed fashion.

%%%%%%%%%%%%%%%%%%%%%%%%%%%%%%%%%%%%%%%%%%%%%%%%%%%%%%%%%%%%%%%%%%%%%%%%%%%%%%%%%%%%%%%%%%%%%%%%%%%%%%%%%%%%%%
%%%%%%%%%%%%%%%%%%%%%%%%%%%%%%%%%%%%%%%%%%%%%%%%%%%%%%%%%%%%%%%%%%%%%%%%%%%%%%%%%%%%%%%%%%%%%%%%%%%%%%%%%%%%%%
%%%%%%%%%%%%%%%%%%%%%%%%%%%%%%%%%%%%%%%%%%%%%%%%%%%%%%%%%%%%%%%%%%%%%%%%%%%%%%%%%%%%%%%%%%%%%%%%%%%%%%%%%%%%%%

\appendix

{\color{changecolor}

\section{Proof of Theorem \ref{theo:backbone}} \label{sec:appendix-proof}

{\color{mr}

In this section, we provide the proof of Theorem \ref{theo:backbone}. First, we restate our model and assumptions, introduce some additional notation, and give a simplified version of the backbone method (Algorithm \ref{alg:backbone}), which we analyze. Then, we proceed with the proof, which we split into three main parts.

\subsection{Model \& Assumptions}  \label{subsec:appendix-model}

We begin by repeating our model and assumptions, introducing some additional notation, and giving a simplified version of Algorithm \ref{alg:backbone}.

\paragraph{Assumptions. \quad} To facilitate the reader, we restate the conditions that our model satisfies (Assumption \ref{asmn:model}). Let $p > n \geq k > 0$ be integers. Further, let
\begin{itemize}
    \item[-] $\bm X \in \mathbb{R}^{n\times p}$ be a random design matrix such that each row $\bm x_i, i \in [n],$ is an iid copy of the random vector $X = (X_1,...,X_{p}) \sim \mathcal{N}(0,\bm I_{p})$.
    \item[-] $\bm \beta \in \{-1,0,1\}^{p}$ be fixed but unknown regressors that satisfy $\|\bm \beta \|_0 = k$ and let $\mathcal{S}^{\text{true}}$ denote the set of indices that correspond to the support of the true regressor $\bm \beta$.
    \item[-] $\bm y = \bm X \bm \beta + \bm \varepsilon$ be a response vector where the noise term $\bm \varepsilon$ consists of iid entries $\varepsilon_i \sim \mathcal{N}(0,\sigma^2), i \in [n].$
\end{itemize}
Moreover, we assume that $\log p = O(n^\xi)$ for some $0<\xi<1.$ When we make asymptotic arguments, we take $p \rightarrow \infty.$

\paragraph*{Notation. \quad} Given $\mathcal{S} \subseteq [p]$, we denote by $X_{\mathcal{S}}$ the random vector constructed by selecting from $X$ the entries that are in $\mathcal{S}$, i.e., $X_{\mathcal{S}} = (X_i)_{i \in \mathcal{S}}$. We let $\mathcal{S}_k^{p} = \{ \bm s \in \{0,1\}^{p}: \sum_{j \in [p]} s_j = k \}.$

\paragraph*{Simplified Algorithm. \quad} We analyze a simplified version of the backbone method, which we outline below:
\begin{enumerate}
    \item Screening step: select top $\lceil \alpha p \rceil$ features based on  their empirical marginal correlation with the response $s_j = \bm X_j^{\top} \bm y$. Parameter $\alpha$ satisfies $0<\alpha\leq1$. 
    
    \item Construct $M$ subproblems. In each subproblem, sample uniformly at random $\lceil \beta \alpha p \rceil$ features among those that survived in Step 1. %{\color{mr}In each subproblem, sample each feature among those that survived in Step 1 with probability $\beta$.} % 
    Parameter $\beta$ satisfies $0<\beta\leq1$. Note that, within each subproblem, features are sampled without replacement, i.e., each feature can appear at most once in a subproblem's feature set.
    
    \item For $m=1,\dots,M,$ approximately solve sparse regression via its boolean relaxation on the $m$-th selected subset of features. Let $\mathcal{S}^m$ be the solution to the $m$-th subproblem.
    
    \item Define the backbone set $\mathcal{B} = \cup_{m=1}^M S^m$.
\end{enumerate}
The simplifications we perform are as follows. Contrary to Algorithm \ref{alg:backbone}, we now assume that the backbone method operates in a single iteration. Further, in each subproblem, we sample features uniformly at random (instead of using Algorithm \ref{alg:construct_subproblems}). Our goal is to show that, with high probability, $\mathcal{S}^{\text{true}} \subseteq \mathcal{B}.$

\begin{rem}
    Similar to \cite{fan2008sure}, to avoid the selection bias in the screening step, we can split the sample in two halves and use the first for Step 1 and the second for Step 3 of the simplified algorithm.
\end{rem}

\subsection{Proof}

We next provide the proof of Theorem \ref{theo:backbone}.

\begin{proof}

We organize the proof as follows. First, we define the events which our analysis is based on and develop our high-level approach. Then, we prove the desired statement by analyzing separately three events.

\paragraph*{Definition of Events \& High-Level Approach. \quad}

We consider the following events:
\begin{itemize}
    \item $\mathcal{E}_{\text{backbone}}:$ the simplified algorithm fails, namely, $\exists j \in [p]$ such that $j \in \mathcal{S}^{\text{true}}$ and $j \not\in \mathcal{B}$.
    
    \item $\mathcal{E}_{\text{screen}}:$ any relevant feature is missed in Step 1.
    
    \item $\mathcal{E}^m_{\text{subproblem}}:$ sparse regression's boolean relaxation fails to recover the relevant features in subproblem $m \in [M]$. Let $p_{\text{subproblem}}$ be an upper bound on the probability of this event across all subproblems, i.e., $\mathbb{P}(\mathcal{E}^m_{\text{subproblem}}) \leq p_{\text{subproblem}}, \ \forall m \in [M].$
    
    \item $\mathcal{E}^j_{\text{miss}}:$ relevant feature $j \in \mathcal{S}^{\text{true}}$ is selected in Step 1 but is not selected after Steps 2 and 3 of the simplified algorithm.
\end{itemize}
By the union bound, the probability of failure of the simplified algorithm is at most
\begin{equation} \label{eq:union-bound}
    \begin{split}
        p_{\text{backbone}} := \mathbb{P}(\mathcal{E}_{\text{backbone}}) 
        & = \mathbb{P} \left[ \mathcal{E}_{\text{screen}} \cup \left( \cup_{j \in S^{\text{true}}} \mathcal{E}^j_{\text{miss}} \right) \right] \\
        & \leq \mathbb{P} \left( \mathcal{E}_{\text{screen}} \right) + \sum_{j \in S^{\text{true}}} \mathbb{P} \left( \mathcal{E}^j_{\text{miss}} \right).%\\
        % & \leq \mathbb{P} \left( \mathcal{E}_{\text{screen}} \right) + \sum_{j \in S^{\text{true}}} \sum_{m=0}^M \left(p_{\text{subproblem}}\right)^m \mathbb{P} \left( Z_j = m \right)\\
        % & = \mathbb{P} \left( \mathcal{E}_{\text{screen}} \right) + k \sum_{m=0}^M \left(p_{\text{subproblem}}\right)^m \mathbb{P} \left( Z = m \right),
    \end{split}
\end{equation}
We aim to show that both terms in the RHS of Equation \eqref{eq:union-bound} converge to zero at a rate of at least $O(\frac{1}{\alpha p}).$
% where $Z \sim \text{Bino}(M,\beta).$

\paragraph*{Analysis of Event $\mathcal{E}_{\text{screen}}$. \quad}

To show that $\mathbb{P}(\mathcal{E}_{\text{screen}})$ converges to zero, we use Theorem \ref{theo:sis} (below) from \cite{fan2008sure} (given as Theorem 1 in their original paper). Informally, Theorem \ref{theo:sis} asserts that, under five conditions and provided that $\alpha$ is sufficiently large, Step 1 succeeds with high probability. %The model described in Assumption \ref{asmn:model} trivially satisfies four out of these five conditions (e.g., on the eigenvalues of the data covariance matrix, on the magnitude of the nonzero regressors, etc.), whereas the fifth condition requires that $p>n,$ $\log p = O(n^\xi)$ for some $0<\xi<1$. 
The conditions are the following:
\begin{enumerate}
    \item $\bm X$ satisfies the concentration property, i.e., there exist some $c,c_1>1$ and $C_1>0$ such that the deviation inequality
    $$\mathbb{P}\left[\lambda_{\max}(\Tilde{p}^{-1}\Tilde{X}\Tilde{X}^\top) > c_1 \text{ or } \lambda_{\min}(\Tilde{p}^{-1}\Tilde{X}\Tilde{X}^\top) < \frac{1}{c_1}\right] \leq \exp(-C_1n),$$ 
    where $\lambda_{\max}(\cdot),\lambda_{\min}(\cdot)$ denote the largest and smallest eigenvalue of a matrix, holds for any $n \times \Tilde{p}$ submatrix $\Tilde{\bm X}$ of $\bm X$ with $cn<\Tilde{p}\leq p.$\\
    This is known to be true when $X$ (i.e., the distribution of the iid rows of $\bm X$) has a $p$-variate Gaussian distribution.
    
    \item $X$ has a spherically symmetric distribution and $\varepsilon \sim \mathcal{N}(0,\sigma^2)$ for some $\sigma>0.$\\
    These are both satisfied by the model described in Assumption \ref{asmn:model}.
    
    \item $\text{var}(Y) = O(1)$ and, for some $\kappa \geq 0$ and $c_2,c_3>0,$ 
    $$\min_{j \in \mathcal{S}^{\text{true}}} | \beta_j | \geq \frac{c_2}{n^\kappa} \text{ and } \min_{j \in \mathcal{S}^{\text{true}}} | \text{cov}(\beta_j^{-1}Y,X_j) | \geq c_3.$$
    Concerning the first inequality, $\kappa$ controls the rate of probability error in recovering the true sparse model. In our setting, since $\beta_j \in \{-1,1\}$ for all $j \in \mathcal{S}^{\text{true}}$, we can set $\kappa=0$ and hence $0<c_2\leq1$.\\
    Concerning the second inequality, it rules out the situation in which a relevant feature is marginally uncorrelated with $Y$, but jointly correlated with $Y$. In our setting, it holds that $\text{cov}(\beta_j^{-1}Y,X_j) = \mathbb{E}[X_j^2]$ for all $j \in \mathcal{S}^{\text{true}}$ and hence $0<c_3\leq1$.
    
    \item There exist some $\tau \geq 0$ and $c_4>0$ such that $\lambda_{\max}(\bm \Sigma) \leq c_4 n^\tau,$ where $\bm \Sigma$ is the data covariance matrix.\\
    This condition rules out the case of strong collinearity. In our setting, this trivially holds with $\tau=0$ and $c_4\geq 1$ since $\bm \Sigma = \bm I_{p}.$
    
    \item $p>n$ and $\log p = O(n^\xi)$, for some $0<\xi<1-2\kappa=1$, since we have set $\kappa=0$

\end{enumerate}
Then, formally, \cite{fan2008sure} prove the following theorem (adjusted to our notation):
\begin{theo} \label{theo:sis}
    Consider the model described in Assumption \ref{asmn:model} and assume $\log p = O(n^\xi)$, for some $0<\xi<1$. Then, $\exists \phi<1$ such that, when $\alpha = \Theta\left(\frac{n^{1-\phi}}{p}\right)$, we have $ \mathbb{P}(\mathcal{E}_{\text{screen}}) = O\left[ \exp \left( -\frac{n}{\log n} \right) \right]. $ 
\end{theo}
Notice that, since $\log p = O(n^\xi)$, $n \rightarrow \infty$ as $p \rightarrow \infty.$
Since we pick $\alpha = \Theta\left(\frac{n^{1-\phi}}{p}\right)$, we have that $\frac{1}{\alpha p} = \Theta(n^{\phi -1}) = \Theta(\exp(-(1-\phi)\log n)).$
Therefore, by Theorem \ref{theo:sis}, $\mathbb{P}(\mathcal{E}_{\text{screen}})$ decays as $O\left[ \exp \left( -\frac{n}{\log n} \right) \right]$, which is obviously faster than $O(\frac{1}{\alpha p}).$

% {\color{red} Check (and maybe explicitly state) Fan conditions}

% {\color{red}
% Write this as function of p and elaborate how fast this goes to 0.

% Because of logp = O(n xi), n goes to infty. 

% We pick alpha = theta ( whatever )

% 1/ alpha p = Theta($n^{\phi -1}$) = Theta(exp(-(1-phi)logn)) 

% Therefore, \mathbb{P}(E1) decays faster than 1/alpha p
% }

\paragraph*{Analysis of Event $\mathcal{E}^m_{\text{subproblem}}$. \quad}

We next show that $\mathbb{P}(\mathcal{E}^m_{\text{subproblem}})$ converges to zero for any subproblem $m$; despite the fact that $\mathbb{P}(\mathcal{E}^m_{\text{subproblem}})$ does not appear in the RHS of Equation \eqref{eq:union-bound}, we need it to show convergence of $\sum_{j \in S^{\text{true}}} \mathbb{P} \left( \mathcal{E}^j_{\text{miss}} \right).$
To show that $\mathbb{P}(\mathcal{E}^m_{\text{subproblem}})$ converges to zero, we momentarily turn to the original sparse regression problem. Let $\mathcal{S}$ denote the set of features that sparse regression's boolean relaxation selects when applied to the entire problem (where all $p$ features are considered). \cite{bertsimas2020sparse} prove the following theorem:
\begin{theo} \label{theo:sr}
    Consider the model described in Assumption \ref{asmn:model} and assume $\gamma=\frac{1}{n}$, $p-k>k$. Then, for all $\theta \geq 1$, for samples $n \geq \theta (\sigma^2+2k) \log(p-k),$
    we have $\mathbb{P}(\mathcal{S} \not= \mathcal{S}^{\text{true}}) = O \left( e^{-\theta} \right)$.
\end{theo}

In our setting, we cannot directly apply Theorem \ref{theo:sr} since, within each subproblem, we do not necessarily sample all relevant features.
We consider an arbitrary subproblem $m \in [M]$ and denote by $\mathcal{P}^m$ the set of features sampled in the $m$-th subproblem's feature set.
When we solve sparse regression's boolean relaxation for the $m$-th subproblem, we only observe the relevant features in $\mathcal{S}^{\text{true}} \cap \mathcal{P}^m$.
Any relevant feature $j$ such that $j \in \mathcal{S}^{\text{true}}$ and $j \not \in \mathcal{P}^m$ is viewed as noise in the $m$-th subproblem.
Therefore, we consider the new noise term 
$$\varepsilon' = \left(X_{\mathcal{S}^{\text{true}} \cap (\mathcal{P}^m)^c}\right)^{\top} \bm \beta_{\mathcal{S}^{\text{true}} \cap (\mathcal{P}^m)^c}+ \varepsilon,$$
which is the sum of at most $k-k_0$ (with $k_0 = |\mathcal{S}^{\text{true}} \cap \mathcal{P}^m|$) independent $\mathcal{N}(0,1)$ random variables and one $\mathcal{N}(0,\sigma^2)$ random variable.
Thus, $ \varepsilon' \sim \mathcal{N}(0,k-k_0+\sigma^2).$

% To simplify our analysis, we take $k_0=1$, namely, only one relevant feature is sampled in subproblem $m$. As we explain next, this gives a valid upper bound of $\mathbb{P}(\mathcal{E}^m_{\text{subproblem}})$. 

Let $\mathcal{S}^m$ denote the set of features that sparse regression's boolean relaxation selects when applied to subproblem $m$. Then, by directly applying the result stated in Theorem \ref{theo:sr}, we obtain the following lemma:
\begin{lem} \label{lem:sr-sub}
    Consider the model described in Assumption \ref{asmn:model} and assume $\gamma=\frac{1}{n}$, $\beta \alpha p>2k_0$.
    Then, for all $\theta \geq 1$, for samples $n \geq \theta (\sigma^2+2k) \log(\beta \alpha p) \geq \theta (\sigma^2+k+k_0) \log(\beta \alpha p-k_0),$
    we have 
    $$ \mathbb{P}(\mathcal{E}^m_{\text{subproblem}}) = \mathbb{P}( \mathcal{S}^{\text{true}} \cap \mathcal{P}^m \setminus \mathcal{S}^m \not= \emptyset) \leq \mathbb{P}(\mathcal{S}^m \not= \mathcal{S}^{\text{true}} \cap \mathcal{P}^m) = O \left( e^{-\theta} \right).$$
\end{lem}
We make the following remarks on Lemma \ref{lem:sr-sub}:
\begin{rem}
    Lemma \ref{lem:sr-sub} asserts that the probability of not exactly recovering the true support -which is, in fact, more restrictive than our requirement to not miss any relevant feature- decreases exponentially with the parameter $\theta$; as the sample size $n$ increases, we are able to select larger $\theta$ and hence obtain tighter guarantees. 
\end{rem}
\begin{rem}
    In the case $k_0=0$, we are dealing with a problem with no relevant features, so the upper bound on the probability of error $\mathbb{P}(\mathcal{E}^m_{\text{subproblem}})$ is trivially satisfied. 
\end{rem}
Since all subproblems are constructed in the same way, the analysis is the same for all $m \in [M]$.
This gives the upper bound $\mathbb{P}(\mathcal{E}^m_{\text{subproblem}}) \leq p_{\text{subproblem}} = O \left( e^{-\theta} \right).$ 
To guarantee a $O(\frac{1}{\alpha p})$ rate of convergence, we set $\theta = \log(\alpha p)$
and, therefore, require that the sample size satisfies 
\begin{equation} \label{eqn:sr-sub-samples}
    n \geq (\sigma^2+2k) \log(\alpha p)  \log(\beta \alpha p).
\end{equation}
Put together, we have that, for all $m \in [M]$, for sufficiently many samples, $\mathbb{P}(\mathcal{E}^m_{\text{subproblem}}) \leq p_{\text{subproblem}} = O(\frac{1}{\alpha p}).$ 
\begin{rem}
    The RHS in Equation \eqref{eqn:sr-sub-samples} is $\Theta(\log^2(n)),$ so, for sufficiently large $n$, the requirement in Equation \eqref{eqn:sr-sub-samples} is satisfied.
\end{rem}

%All other parameters being fixed, the required sample size is maximized when $k_0=0$; in this case, we are dealing with a problem with no relevant features, so the upper bound on the probability of error $\mathbb{P}(\mathcal{E}^m_{\text{subproblem}})$ is trivially satisfied. 

\paragraph*{Analysis of Event $\mathcal{E}^j_{\text{miss}}$. \quad}

To show that $\sum_{j \in S^{\text{true}}} \mathbb{P} \left( \mathcal{E}^j_{\text{miss}} \right)$ converges to zero, we fix an arbitrary feature $j \in \mathcal{S}^{\text{true}}$. We make the following observations:
\begin{itemize}
    \item[-] To miss feature $j$ in subproblem $m$ we have to either not select it in the feature set of subproblem $m$ or fail to select it as relevant after solving subproblem $m$ (notice that the latter scenario is contained in the event $\mathcal{E}^m_{\text{subproblem}}$). To miss feature $j$ across all $M$ subproblems, in which case event $\mathcal{E}^j_{\text{miss}}$ is realized, the aforementioned has to hold for all subproblems $m \in [M]$.
    \item[-] The uniform sampling and the subproblems' solution method are independent across subproblems.
    \item[-] The number of subproblems in which feature $j$ is sampled follows a binomial distribution with parameters $M$ (number of subproblems) and $\beta$ (fraction of features that are included in any subproblem); we denote $Z_j \sim \text{Bino}(M,\beta).$ \footnote{Since, in each subproblem, we sample $\lceil \beta \alpha p\rceil$ features uniformly at random from a total of $\lceil \alpha p \rceil$ features, a given feature $j$ is selected with probability $\beta$. The joint distribution of sampled features within a subproblem is multinomial, but the union bound we used in Equation \eqref{eq:union-bound} allows us to look at the marginal distributions of sampled (relevant) features, which are binomial.}
\end{itemize}
Combining the above and denoting by $\mathcal{P}^m$ the set of features sampled in the $m$-th subproblem's feature set, the probability that we miss feature $j$ can be upper-bounded as
\begin{equation} \label{eqn:miss-feature}
\begin{split}
    \mathbb{P} \left( \mathcal{E}^j_{\text{miss}} \right) 
    & \leq \prod_{m=1}^M \mathbb{P} \left( j \not\in \mathcal{P}^m \bigcup j \in \mathcal{P}^m \bigcap \mathcal{E}^m_{\text{subproblem}} \right) \\
    & \leq \sum_{m=0}^M \mathbb{P} \left( Z_j = m \right) \left(p_{\text{subproblem}}\right)^m.
\end{split}
\end{equation}
Plugging Equation \eqref{eqn:miss-feature} into the second term in the RHS of Equation \eqref{eq:union-bound}, letting $Z \sim \text{Bino}(M,\beta)$, and using the binomial theorem, we obtain
\begin{equation} \label{eqn:miss-any}
    \begin{split}
        \sum_{j \in S^{\text{true}}} \mathbb{P} \left( \mathcal{E}^j_{\text{miss}} \right)
        & \leq \sum_{j \in S^{\text{true}}} \sum_{m=0}^M \mathbb{P} \left( Z_j = m \right) \left(p_{\text{subproblem}}\right)^m\\
        & = k \sum_{m=0}^M \mathbb{P} \left( Z = m \right) \left(p_{\text{subproblem}}\right)^m\\
        & = k \sum_{m=0}^M {M\choose m} \beta^m (1-\beta)^{M-m} \left(p_{\text{subproblem}}\right)^m \\
        & = k (1-\beta)^{M} \sum_{m=0}^M {M\choose m} \left(\frac{\beta p_{\text{subproblem}} }{1-\beta}\right)^m  \\
        & = k (1-\beta)^{M} \left(1+\frac{ \beta p_{\text{subproblem}} }{1-\beta}\right)^M \\
        & = k \left(1 - \beta + \beta p_{\text{subproblem}} \right)^M.
    \end{split}
\end{equation}

Let us briefly review the quantities that appear in equation \eqref{eqn:miss-any}: $k$ is the number of relevant features; $p_{\text{subproblem}}$ is an upper bound on the probability of failure for any subproblem and converges to 0 at rate given by Lemma \ref{lem:sr-sub}; $M$ denotes the number of subproblems that we solve; $\beta$ denotes the fraction of the $\alpha p$ screened features that we sample in each subproblem. $M$ and $\beta$ are parameters of the algorithm; among the two, we believe it is realistic to only control $M$, since $\beta$ depends on the available computational resources (e.g., memory). Therefore, we next determine how $M$ has to be selected as function of $k,p_{\text{subproblem}}, \beta,$ so as to guarantee that $k \left(1-\beta+ \beta p_{\text{subproblem}} \right)^M \rightarrow 0.$ Specifically, to guarantee a $\frac{1}{\alpha p}$ rate of convergence, we pick
\begin{equation}
    k \left(1-\beta+ \beta p_{\text{subproblem}} \right)^M = O \left(\frac{1}{\alpha p}\right)
\end{equation}
and, therefore, %{\color{mr}since $p_{\text{subproblem}} = O(\frac{1}{\alpha p})$,}
\begin{equation}
    M 
    = \Omega \left( \frac{\log \left(\alpha p k\right)}{\log\left( \frac{1}{1-\beta+\beta p_{\text{subproblem}}} \right)} \right).
\end{equation}

\paragraph*{Final Result. \quad}

Plugging everything back into Equation \eqref{eq:union-bound}, we get that
\begin{equation} \label{eq:final-result}
    \begin{split}
        p_{\text{backbone}} & \leq \mathbb{P} \left( \mathcal{E}_{\text{screen}} \right) + k \left(1-\beta+ \beta p_{\text{subproblem}} \right)^M \\
        & = o\left( \frac{1}{\alpha p} \right) + O\left( \frac{1}{\alpha p} \right).
    \end{split}
\end{equation}
This completes the proof of Theorem \ref{theo:backbone}.

\end{proof}
}
}

%%%%%%%%%%%%%%%%%%%%%%%%%%%%%%%%%%%%%%%%%%%%%%%%%%%%%%%%%%%%%%%%%%%%%%%%%%%%%%%%%%%%%%%%%%%%%%%%%%%%%%%%%%%%%%
%%%%%%%%%%%%%%%%%%%%%%%%%%%%%%%%%%%%%%%%%%%%%%%%%%%%%%%%%%%%%%%%%%%%%%%%%%%%%%%%%%%%%%%%%%%%%%%%%%%%%%%%%%%%%%
%%%%%%%%%%%%%%%%%%%%%%%%%%%%%%%%%%%%%%%%%%%%%%%%%%%%%%%%%%%%%%%%%%%%%%%%%%%%%%%%%%%%%%%%%%%%%%%%%%%%%%%%%%%%%%

\begin{acknowledgements}
We would like to thank Ryan Cory-Wright, Korina Digalaki, Michael Li, Theodore Papalexopoulos, and Ilias Zadik for fruitful discussions. We are grateful to the referees for their constructive comments.

\end{acknowledgements}

% \section*{Conflict of interest}
%
% The authors declare that they have no conflict of interest.

% BibTeX users please use one of
\bibliographystyle{spbasic}      % basic style, author-year citations
\bibliography{references}   % name your BibTeX data base

\end{document}